\documentclass{article}
\usepackage{jfrExamplee}
\usepackage{graphicx}
\usepackage{apalike}
\usepackage{setspace}

\let\bibhang\relax

\usepackage{natbib}
\usepackage{caption}
\usepackage{subcaption}
\usepackage{floatrow}
\usepackage{multirow}

\usepackage[binary-units = true]{siunitx}
\DeclareSIUnit\px{px}
\DeclareSIUnit\ha{ha}
\usepackage{amsmath}

\usepackage[toc,page]{appendix}
\usepackage[hyphens]{url}

\usepackage{graphicx} 
\usepackage{graphbox} 

\usepackage[super]{nth}

\usepackage{enumitem}
\usepackage{pifont}%
\usepackage{tikz}
\usepackage{color}

\usepackage{ctable}
\newcolumntype{?}{!{\vrule width 1pt}}

\makeatletter
\newcommand\footnoteref[1]{\protected@xdef\@thefnmark{\ref{#1}}\@footnotemark}
\makeatother

\providecommand{\keywords}[1]
{
  \small	
  \textbf{\textit{Keywords--}} #1
}

\def\secref#1{Sec.~\ref{#1}}
\def\figref#1{Fig.~\ref{#1}}
\def\tabref#1{Tab.~\ref{#1}}
\def\eqref#1{Eq.~(\ref{#1})}

\title{GrowliFlower: An image time series dataset for GROWth analysis of cauLIFLOWER}

\author{Jana~Kierdorf\thanks{Corresponding author}\\
Institute of Geodesy and Geoinformation\\
Remote Sensing Group\\
University of Bonn, Germany\\
Niebuhrstraße 1A, 53113 Bonn \\
\texttt{jkierdorf@uni-bonn.de} \\
\And
Laura Verena~Junker-Frohn \\
Institute of Bio- and Geosciences\\
IBG-2: Plant Sciences\\
Forschungszentrum Jülich GmbH, Germany \\
\texttt{l.junker-frohn@fz-juelich.de} \\
\And
Mike Delaney \\
JB Hyperspectral Devices GmbH, Germany \\
\texttt{mike@jb-hyperspectral.com}\\
\And
Mariele~Donoso Olave\\
Institute of Geodesy and Geoinformation\\
University of Bonn, Germany\\
\texttt{s7memuel@uni-bonn.de} \\
\And
Andreas~Burkart \\
JB Hyperspectral Devices GmbH, Germany \\
\texttt{andreas@jb-hyperspectral.com} \\
\And
Hannah~Jaenicke\\
Horticulture Competence Centre\\
University of Bonn, Germany \\
\texttt{hjaenick@uni-bonn.de} \\
\And
Onno~Muller \\
Institute of Bio- and Geosciences\\
IBG-2: Plant Sciences\\
Forschungszentrum Jülich GmbH, Germany \\
\texttt{o.muller@fz-juelich.de} \\
\And
Uwe~Rascher \\
Institute of Bio- and Geosciences\\
IBG-2: Plant Sciences\\
Forschungszentrum Jülich GmbH, Germany \\
\texttt{u.rascher@fz-juelich.de} \\
\And
Ribana~Roscher \\
Institute of Geodesy and Geoinformation\\
Remote Sensing Group\\
University of Bonn, Germany\\
\texttt{ribana.roscher@uni-bonn.de} \\
}

\begin{document}

\maketitle

\begin{abstract}

This article presents GrowliFlower, a georeferenced, image-based UAV time series dataset of two monitored cauliflower fields of size $0.39$ and \SI{0.60}{\ha} acquired in 2020 and 2021.
The dataset contains RGB and multispectral orthophotos from which about 14,000 individual plant coordinates are derived and provided.
The coordinates enable the dataset users the extraction of complete and incomplete time series of image patches showing individual plants. The dataset contains collected phenotypic traits of 740 plants, including the developmental stage as well as plant and cauliflower size. As the harvestable product is completely covered by leaves, plant IDs and coordinates are provided to extract image pairs of plants pre and post defoliation, to facilitate estimations of cauliflower head size. Moreover, the dataset contains pixel-accurate leaf and plant instance segmentations, as well as stem annotations to address tasks like classification, detection, segmentation, instance segmentation, and similar computer vision tasks. 
The dataset aims to foster the development and evaluation of machine learning approaches. It specifically focuses on the analysis of growth and development of cauliflower and the derivation of phenotypic traits to foster the development of automation in agriculture.
Two baseline results of instance segmentation at plant and leaf level based on the labeled instance segmentation data are presented. The entire data set is publicly available.

\end{abstract}


\section{Introduction}
\label{sec:introduction}
Field-grown crops are strongly affected by prevailing environmental conditions.
As a consequence, crop production requires careful plant management and complex decisions to minimize yield losses due to abiotic or biotic stresses. 
Farmers support plant growth and development through irrigation, fertilization, weeding, and pesticides applications, which are, however, costly and labor-intensive. 
To optimize plant management and support decision-making, farmers rely on frequent crop monitoring, but to date, this remains time-consuming and requires expert knowledge. 
Typically, farmers and agricultural advisors monitor fields regularly through spot checks of individual plants.
Here, remote sensing and analysis methods can help farmers to monitor whole fields more comprehensively \citep{chi2016big,weiss2020remote}.

Remote sensing data can be collected at all scales without damaging or impacting crops. 
Large-scale observations from satellites or aircraft and medium-scale observations from Unmanned Aerial Vehicles (UAVs) provide an overview of larger agricultural areas \citep{lillesand2015remote}. 
Large-area crop monitoring with such sensors makes it possible to detect heterogeneity in the field and support the farmer's decision-making regarding field management. 
With such area-wide yet detailed information on biotic and abiotic stress, these factors can be counteracted more selectively to support environmentally friendly plant management. 
Medium-scale and close-range observations acquired from UAVs and ground robots are beneficial for collecting detailed information and can be used especially well for phenotyping individual plants.
\cite{nock2016functional}, for example, use optical remote sensing data to define traits like structural and phenotypical characteristics on all levels from individual plants to whole areas.
Other applications using remote sensing data are yield estimation \citep{chaparro2018band}, yield forecasting \citep{mosleh2015application}, and monitoring of rapid land surface changes \citep{verger2014near}.

To process and interpret large amounts of remote sensing data, machine learning (ML) methods become increasingly important \citep{lary2016machine}. 
ML is concerned with learning a predictive function that relates observations to the desired output. The learned models can be flexibly designed with respect to the type of observations \citep{debolini2015changes, reichstein2019deep}, and can, for example, identify plant traits from remote sensing data \citep{ali2015review,verrelst2019quantifying}.
A main area of application is plant phenotyping, which can be made more objective and automated by using advanced ML methods such as deep neural networks.
\cite{romera2016recurrent}, \cite{ren2017end}, and \cite{scharr2016leaf}, for example, learn ML models to infer phenotypic traits such as the number of leaves per plant.
Similar traits can also be derived with a combination of object and leaf keypoint detection, allowing observation of plant growth as done by \cite{weyler2021joint}.
\cite{sa2016deepfruits} use deep convolutional neural networks for the detection of single fruits, which serves e.g.~as a precursor for later autonomous harvesting \citep{arad2020development}.
\cite{drees2021temporal} use image time series of cauliflower and broccoli to predict the growth in the field using conditional generative adversarial networks \citep{isola2017image}. In detail, they generate an image of a plant at a later time point and use Mask R-CNN \citep{he2017mask} to calculate the projected leaf area.
Another typical agricultural application is weed control in the field, where weeds, crops, and soil need to be distinguished. With the use of neural networks, promising results have already been achieved, where the task can be approached using classification \citep{lottes2017effective}, detection \citep{lottes2018fully} or semantic segmentation \citep{milioto2018real,ahmadi2021virtual}.

To foster the development of ML methods for plant-specific tasks using remote sensing data, benchmark datasets with annotations and in-situ measurements are beneficial.
Although various benchmark datasets already exist, many of these are domain-specific with objects such as buildings \citep{roscher2020semcity}, animals \citep{deng2009imagenet} and other semantics such as land cover \citep{cordts2015cityscapes}.
Therefore, they are not suitable for plant applications in general. The link between ML and plant sciences is becoming increasingly important \citep{lary2016machine}, as can be seen from the growing number of related publications in recent years \citep{chebrolu2017agricultural,forster2019hyperspectral,kierdorf2019detection,zabawa2019detection,halstead2020fruit,ahmadi2021virtual}. Despite increased demand, there are only a few publicly available plant-specific datasets that can be used for ML purposes. 

Among the few datasets that are publicly available or covered in the literature, many have been acquired in a greenhouse \citep{scharr2014annotated,minervini2016finely,murecsan2017fruit,halstead2020fruit} or are based on synthetically generated data \citep{ward2018synthetic,kierdorf2022behind}, making them difficult to apply to real-world scenarios.
In particular, the greenhouse-grown plant Arabidopsis thaliana rosettes is a frequently used research plant in the ML community due to its simple rosette morphology \citep{scharr2014annotated}.
However, agricultural crop plants are more diverse in their morphology and their development is affected by changing environmental conditions and abiotic as well as biotic stresses. Because of this, the need for agricultural datasets that represent field conditions and cover challenges such as occlusions and variable shapes, poses, and colors of plants and plant parts is high.

An active research area is modeling the temporal development of plant growth and plant traits, which requires datasets that monitor plants over time. However, publicly available time series datasets of plants are rare. 
One of these is the cauliflower and broccoli (Brassica oleracea) dataset from \cite{bender2020high}. The dataset was acquired with a camera-equipped robot that took close-range images at several time points. However, the dataset is limited to a few plants and lacks semantic information and accurate georeferencing of single plants.

Cauliflower is a suitable crop plant to develop ML algorithms, as its cultivation, morphology and economic value give rise to many potential applications in the context of the digitization of agriculture.
It is a high-value crop that needs to fulfill high quality criteria. Precise timing of plant management procedures is needed to avoid yield losses by abiotic or biotic stress and produce marketable cauliflowers. Cauliflower harvesting is a labor-intensive process, as each cauliflower must be harvested within about one week, when heads have a sufficient size but are not yet overripe. Due to within-field variability in plant development, cauliflower must be harvested by hand. As the head is covered by leaves, each individual cauliflower head must be touched to assess, if it fulfills the size criteria. After cutting and removing the surrounding leaves, product quality is visually assessed, to dismiss heads with discolorations, misshapes, or stress symptoms. Cauliflower growth is highly climate-dependent, making it hard to predict the time for harvest. Depending on the prevailing temperature, irradiance, and soil water availability, plants may develop rather heterogeneous, so that harvesting of simultaneously established fields can take weeks. Under favorable conditions, plants in sequentially established fields develop may need to be harvested at the same time, which requires more workers and lowers the price per cauliflower. The early prediction of harvestable plants and the time of harvest would allow for better planning of sales and bring economic advantages for farmers.

This article presents an agricultural dataset suitable for the development of ML approaches.
The provided dataset is meant to specifically address the analysis of growth and development of crop plants and the derivation of phenotypic traits relevant for agricultural applications to foster the development of automation in agriculture. 
The dataset comprises:
\begin{itemize}
    \item RGB and multispectral orthophotos of two different cauliflower fields acquired over the whole growing period, from planting to harvest;
    \item plant IDs and coordinates enabling the dataset user the extraction of complete and incomplete time series of image patches showing individual plants accompanied with in-situ reference data captured manually on the field;
    \item plant IDs and coordinates enabling the dataset user the extraction of image pairs of plants pre and post defoliation accompanied with a time series of the respective plant to allow an analysis of the correlation between the external appearance and the internal head of the cauliflower plant;
    \item pixel-accurate labeled data useful for classification, detection, segmentation, instance segmentation, and similar computer vision tasks on plant and leaf level.
\end{itemize}
We further present two baselines showing application examples of plant and leaf instance segmentation using our data with the application of Mask R-CNN \citep{he2017mask}.

\section{Field design}\label{sec:fieldDesign}
\begin{figure}[t]
	\centering
    \includegraphics[trim= 50 0 0 0, clip, width=\textwidth]{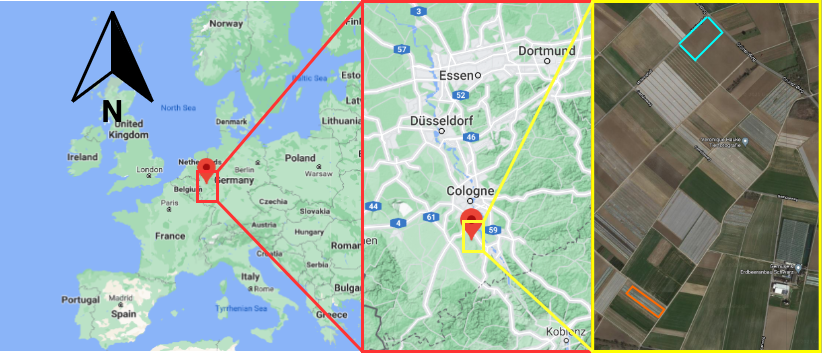}
	\caption{Field locations. The fields are located close to Cologne, a city in west of Germany. Blue: field 1 (2020). Orange: field 2 (2021). (Map source: Google Maps).}
	\label{fig:map}
\end{figure}

Cauliflower fields were located on a farm in Western Germany (50°46'6.742" N, 6°58'20.271"O), close to the city of Bornheim, \SI{20}{\kilo\meter} south of Cologne (see \figref{fig:map}).
The mean annual temperature in Bornheim is 14°C and the mean annual precipitation is \SI{383}{\milli\meter} per year. It is dry 142 days a year with an average humidity of $81\%$. On the farm, fertile soil is available.

We acquired data for two fields: 1) the field shown in \figref{fig:map} in blue, further referred to as field 1, in the year 2020 and 2) the field shown in orange, further referred to as field 2, in the year 2021. The cauliflower plants for both fields are planted in rows with an orientation from northwest to southeast. 
Fields are designed for sprayers with a working width of \SI{18}{\meter}. Before planting, fields were plowed to prepare the soil. Tractors with \SI{1.8}{\meter} track width were used to plant five rows of nursery-grown young cauliflower plants at a time, with 3 rows between tractor tracks. The distance between rows was \SI{0.6}{\meter} and the distance between plants within a row was \SI{0.5}{\meter}, resulting in a planting density of 33000 plants/hectare. Every \SI{18}{\meter}, there is a \SI{2}{\meter} wide lane for spraying and irrigation. Fields were subject to conventional farming practices including hoeing of cauliflower plants before canopy closure to reduce weeds as well as application of pesticides (including herbicides, insecticides, and fungicides). In addition, fields were irrigated when needed using sprinklers. In both fields, abiotic and biotic stresses were consequently rather low, and plants developed rather uniform.

\subsection{Field 1}
\begin{figure}[h]
	\centering
	 \subfloat[]{
	 \includegraphics[trim=0 0 39 0, clip, width=0.19\textwidth]{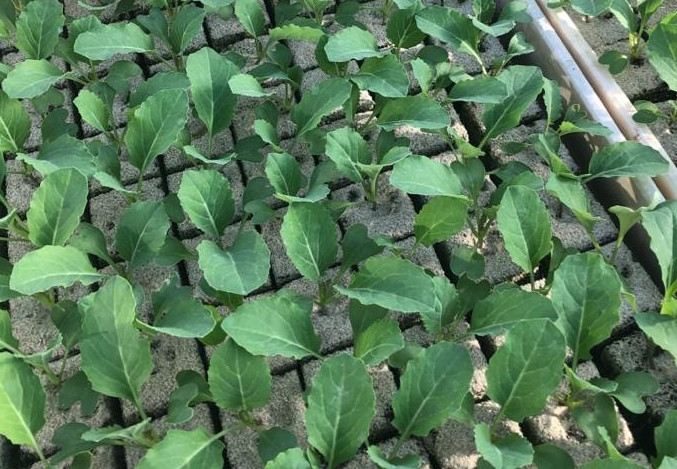}} 
     \subfloat[]{
     \includegraphics[trim=0 24 0 20, clip, width=0.19\textwidth]{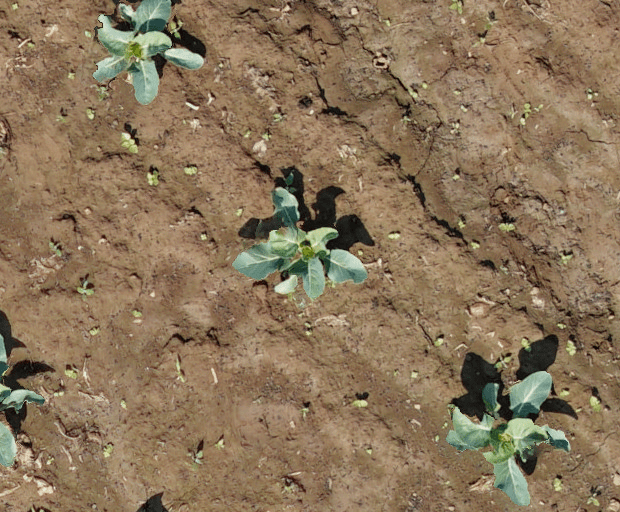}} 
     \subfloat[]{
     \includegraphics[width=0.19\textwidth]{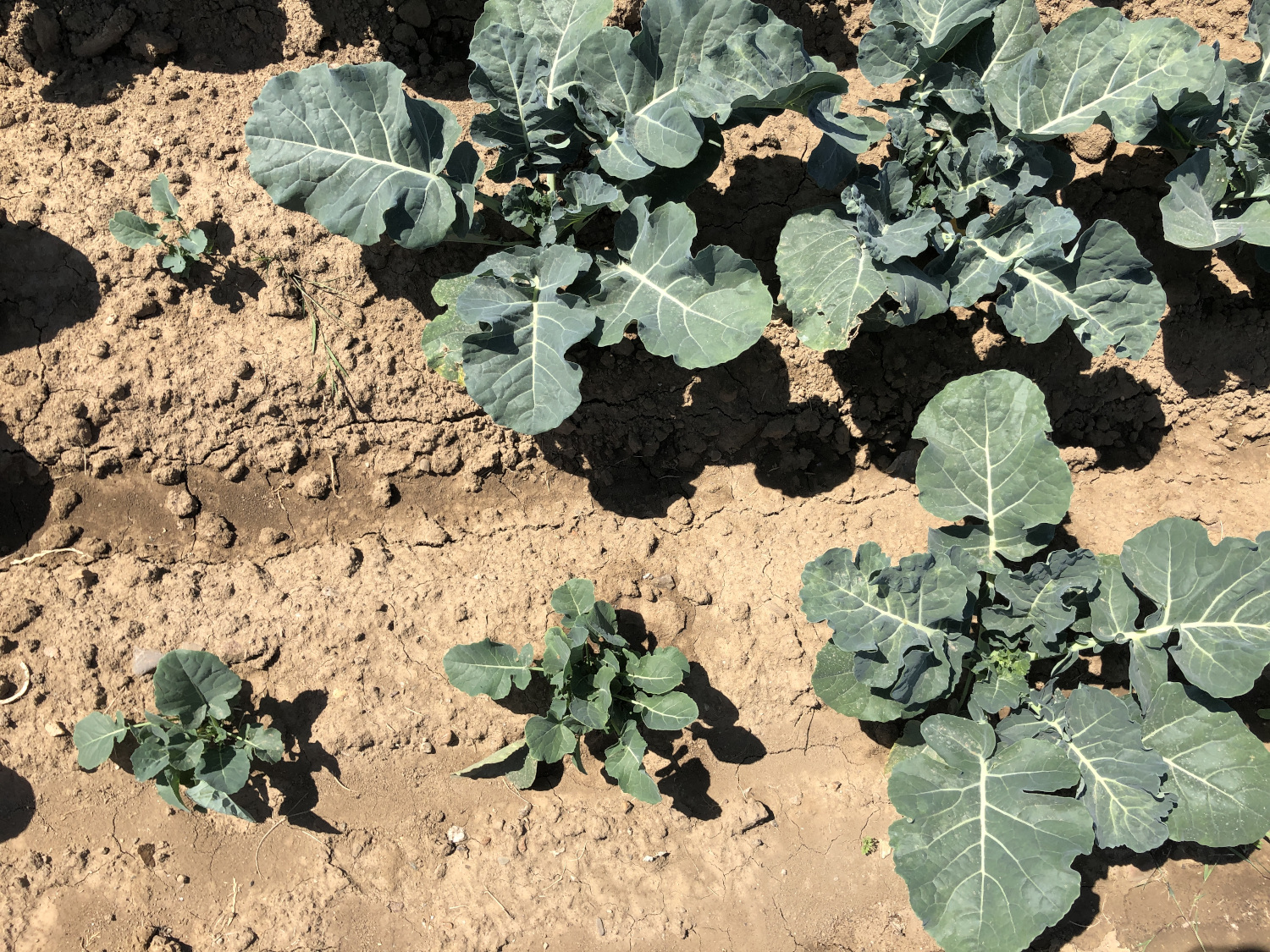}} 
     \subfloat[]{
     \includegraphics[width=0.19\textwidth]{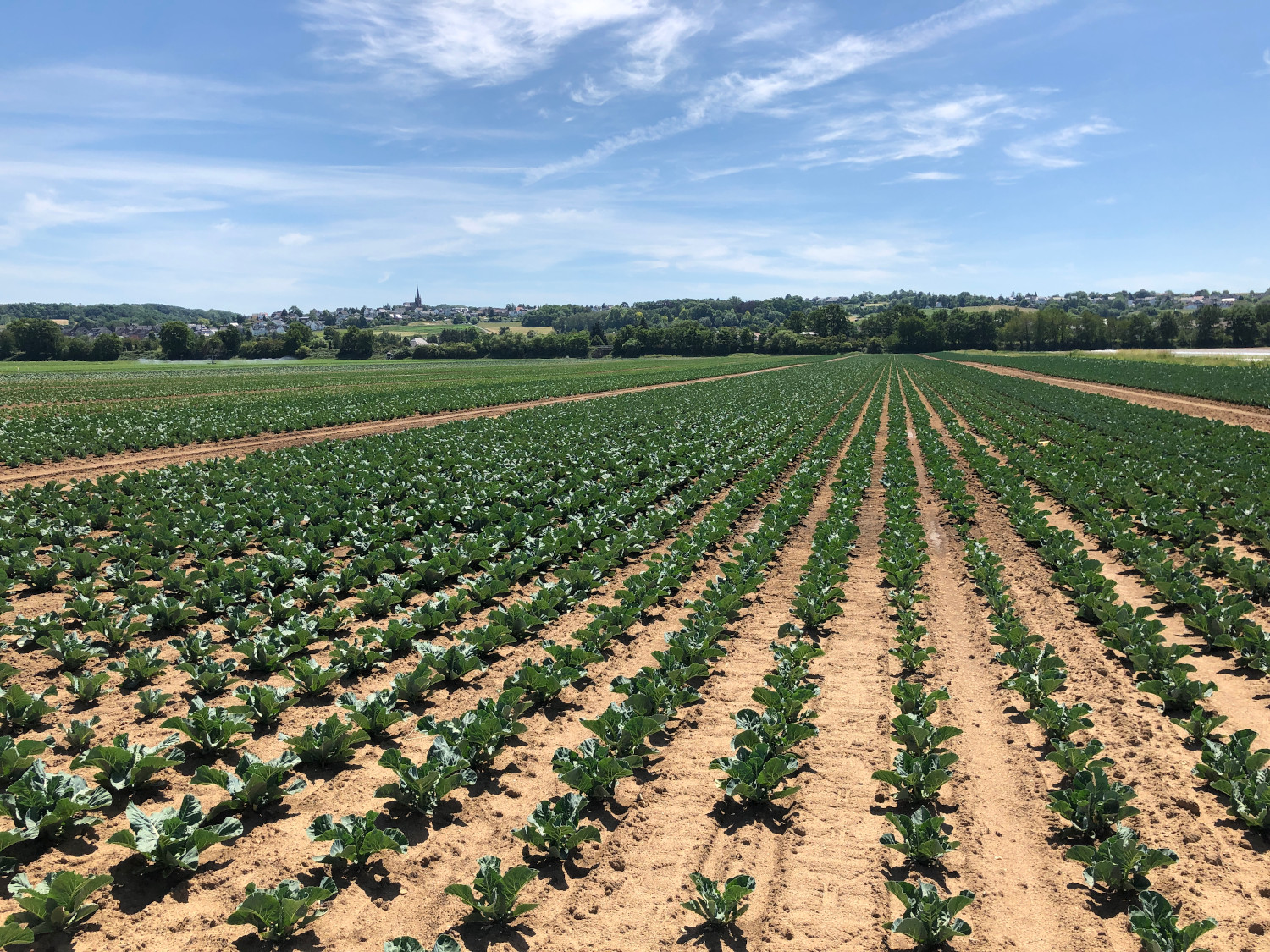}} 
     \subfloat[]{
     \includegraphics[width=0.19\textwidth]{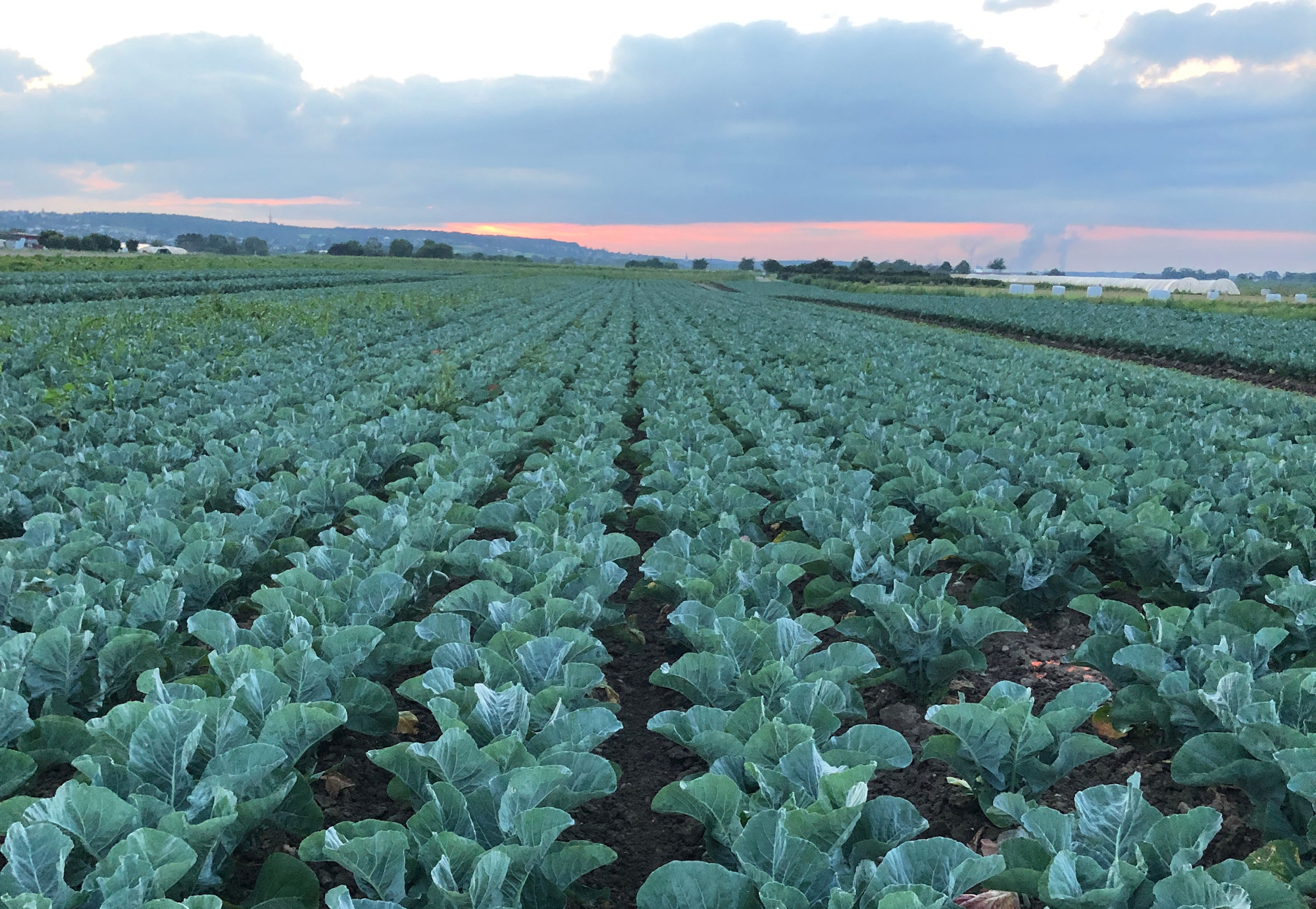}} 
	\caption{Example field and plant images. Image (a) shows seedling trays before planting. Image (b) shows plants two weeks after planting. Image (c-d) were taken four weeks after planting and illustrates how different plants develop over time and to get a feeling of how a field looks like. Image (e) depicts plants shortly before head formation.}
	\label{fig:field_images}
\end{figure}

This field has a width of about \SI{100}{\meter} and a length of \SI{240}{\meter}. Thus, the area is about
\SI{2.4}{\ha}. The field was planted with the cultivar Korlanu (Syngenta, Maintal, Germany). Three quarters of the field were planted with plants from seedling trays (\figref{fig:field_images}, left) on July \nth{28}, 2020 from the southwest direction. The remaining north-eastern part of the field was planted on July \nth{29}, 2020. It is worth mentioning that in this field, almost no weeds have grown.

\subsection{Field 2}
This field has a width of about \SI{55}{\meter} and a length of \SI{210}{\meter}. Thus, the area is approximately \SI{1.32}{\ha}. The field was planted with the cultivar Guideline (Syngenta, Maintal, Germany). The plants were transplanted from seedling trays on June \nth{15}, 2021. Field 2 shows more weeds than field 1, especially along the southwestern edge of the field due to earlier rhubarb cultivation.

\section{Data collection}\label{sec:dataCollection}
Three types of data are collected, namely:
\begin{enumerate}
    \item RGB and multispectral UAV image data with a high spatial resolution which is an indirect measurement of the phenotypic development of the plants;
    \item georeferenced ground control points (GCPs) to locate the data in space, spatially arranged according to the field size to ensure accurate and robust processing of orthophotos \citep{persia2020archival};
    \item in-situ measurements of phenotypic traits characterizing the development state and stress factors that serve as reference observations.
\end{enumerate}

\begin{figure}[t]
	\centering
    \subfloat[Field 1]{
    \includegraphics[trim= 0 184 0 0, clip, width=\textwidth]{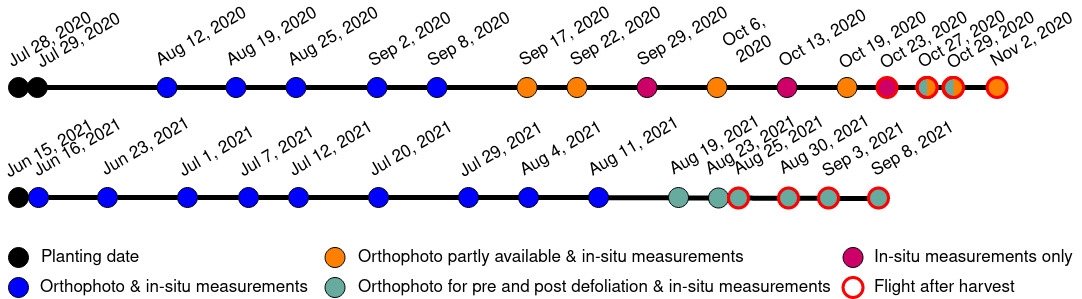} 
    \label{fig:timeline_field1}}
	
    \subfloat[Field 2]{
    \includegraphics[trim= 0 0 0 108, clip, width=\textwidth]{Zeitstrahle2_insitu}
    \label{fig:timeline_field2}}
	\caption{Timelines of acquired data for (a) field 1 and (b) field 2. The illustrated colors represent the data availability for images and in-situ measurements.}
	\label{fig:timeline}
\end{figure}

The different types of data are collected on the same day to synchronize them. However, to ensure that workers are not visible in the image data, the data acquisitions were not conducted at the same time.
The acquisition was carried out once a week during the entire growth period. During the harvest period, data was collected once between two different harvest days and once after the last harvest. Drone flights were only performed on sunny or overcast days to ensure stable illumination for the generation of orthophotos without shading effects due to moving clouds.
Due to this, the time intervals between successive overflights vary slightly. \figref{fig:timeline} illustrates the dates of data collection for both monitored fields. 
As seen in the top timeline, seven orthophotos are only partly available, which will be discussed in \secref{sec:orthophotos}. The data collection took a few hours per day, with the in-situ measurements being the most time-intensive. Data collection was adjusted to both field conditions, resulting in adaptations to camera settings, number of GCPs, and flight altitude.

Thus, the following subsections describe the procedure separately for field 1 and field 2.

\subsection{RGB and multispectral imaging using UAVs}
\begin{figure}[t]
	\centering
    \includegraphics[width=0.32\textwidth]{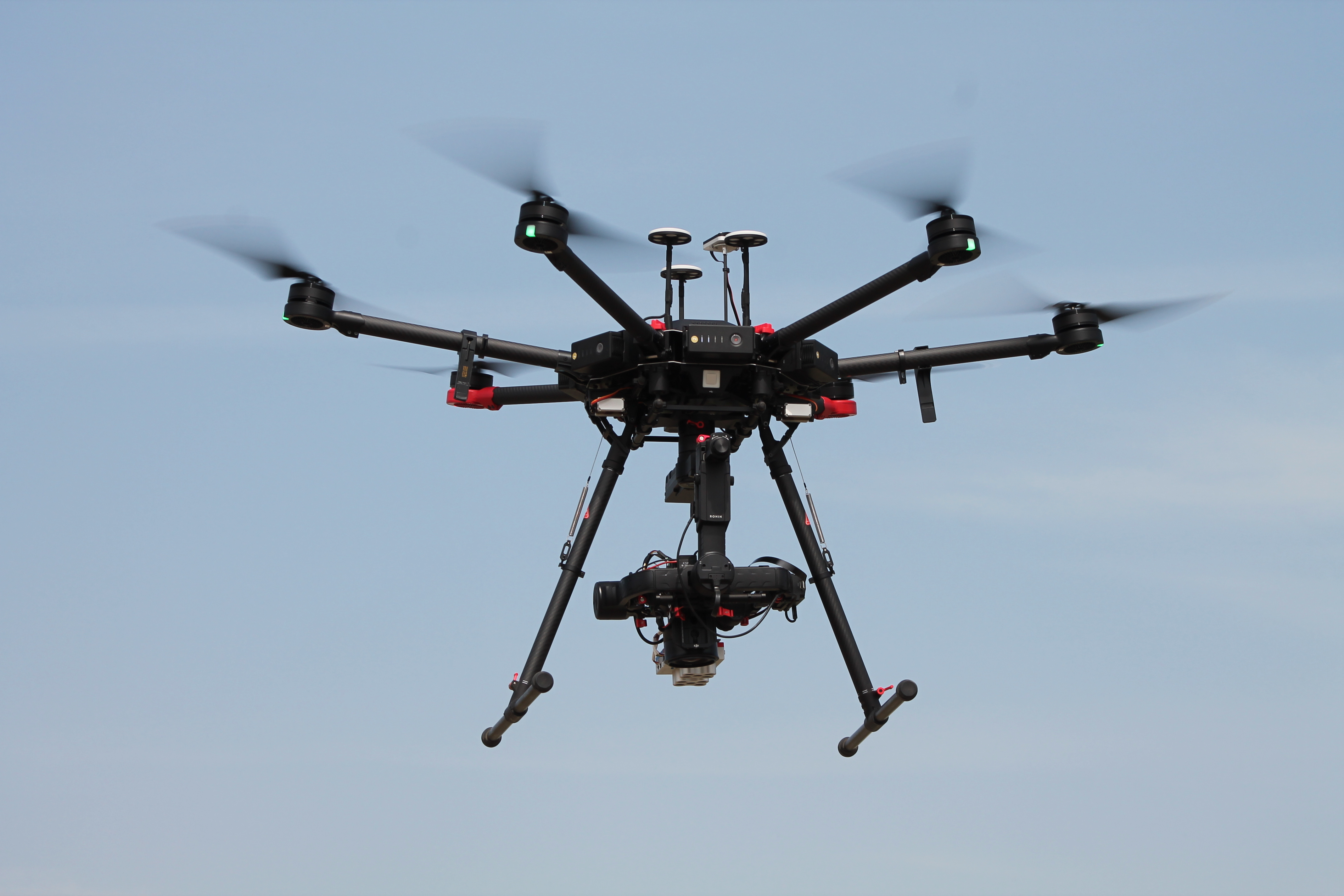}
	\hspace{12pt}
    \includegraphics[width=0.32\textwidth]{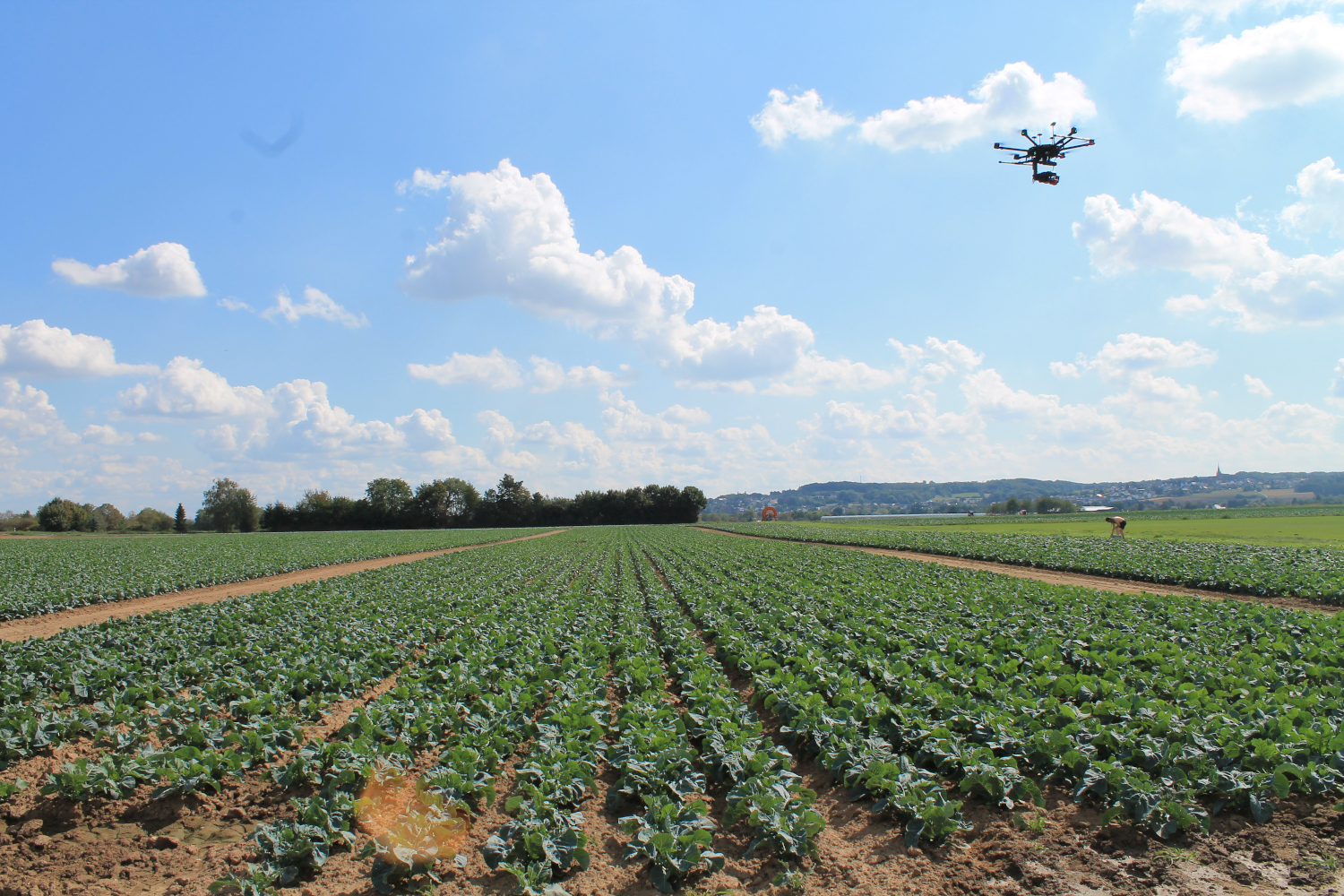}
	\caption{Used DJI Matrice 600 hexacopter for UAV image-based measurements.}
	\label{fig:drone}
\end{figure}

UAV images were taken with a DJI Matrice 600 hexacopter and two mounted cameras (\figref{fig:drone}).
The first camera is a Sony A7 rIII RGB camera. It contains a Zeiss/Batis 2.0 lens with a resolution of 47.4 MP. The focal length is \SI{25}{\milli\meter} with a field of view of \SI{71.5}{\degree}. A shutter speed of \emph{1/\nth{1250}} and aperture of \emph{floating} with a largest value of 2.0 was chosen. The ISO value was set to \emph{automated} for field 1 and changed to 50 for field 2 in order to align our approach with the image-capture settings recommended by Agisoft. 
The second camera is a MicaSense 5CH for multispectral image data. It contains five built-in lenses with a resolution of 1.2 MP per band. The wavelengths of the five acquired bands are \SI{475}{\nano\meter}, \SI{560}{\nano\meter}, \SI{668}{\nano\meter}, \SI{717}{\nano\meter}, and \SI{840}{\nano\meter}. The focal length of the camera is \SI{5.4}{\milli\meter}. 
For field 1, an altitude of around \SI{10}{\meter} and image overlap of 60/80 was used, whereas, for field 2, an altitude of around \SI{16}{\meter} and image overlap of 80/80 was used to optimize data acquisition and resulting image data processing.
We ensured the following for each flight: No irrigation in or too close to the flight area, flying with a temperature and wind speed within the drone's safe operating range, and no rain during the whole flight.

\subsubsection{Time series flights}
For each acquisition date, a specified field area was flown over once which remains the same for the whole growing period. For field 1, this area had a width of \SI{91}{\meter} and length of \SI{62}{\meter}, resulting in approximately \SI{0.60}{\ha}. For field 2, the area had a width of \SI{30}{\meter} and length of \SI{131}{\meter},
resulting in approximately \SI{0,39}{\ha}. 

Because the plant does not necessarily grow straight, the center of the plant in later growing stages does not match the position of the seedling traits exactly \citep{grenzdorffer2019automatic}. A shift of up to $\pm \SI{10}{\centi\meter}$ between the center position of the head and the stem position of the early growing stages was observed.

\subsubsection{Defoliation flights}
\begin{figure}
    \centering
     \begin{minipage}{0.6\textwidth}
        \subfloat[Defoliation plots for field 2.]{
        \includegraphics[trim = 87 25 87 0, clip, width=\textwidth]{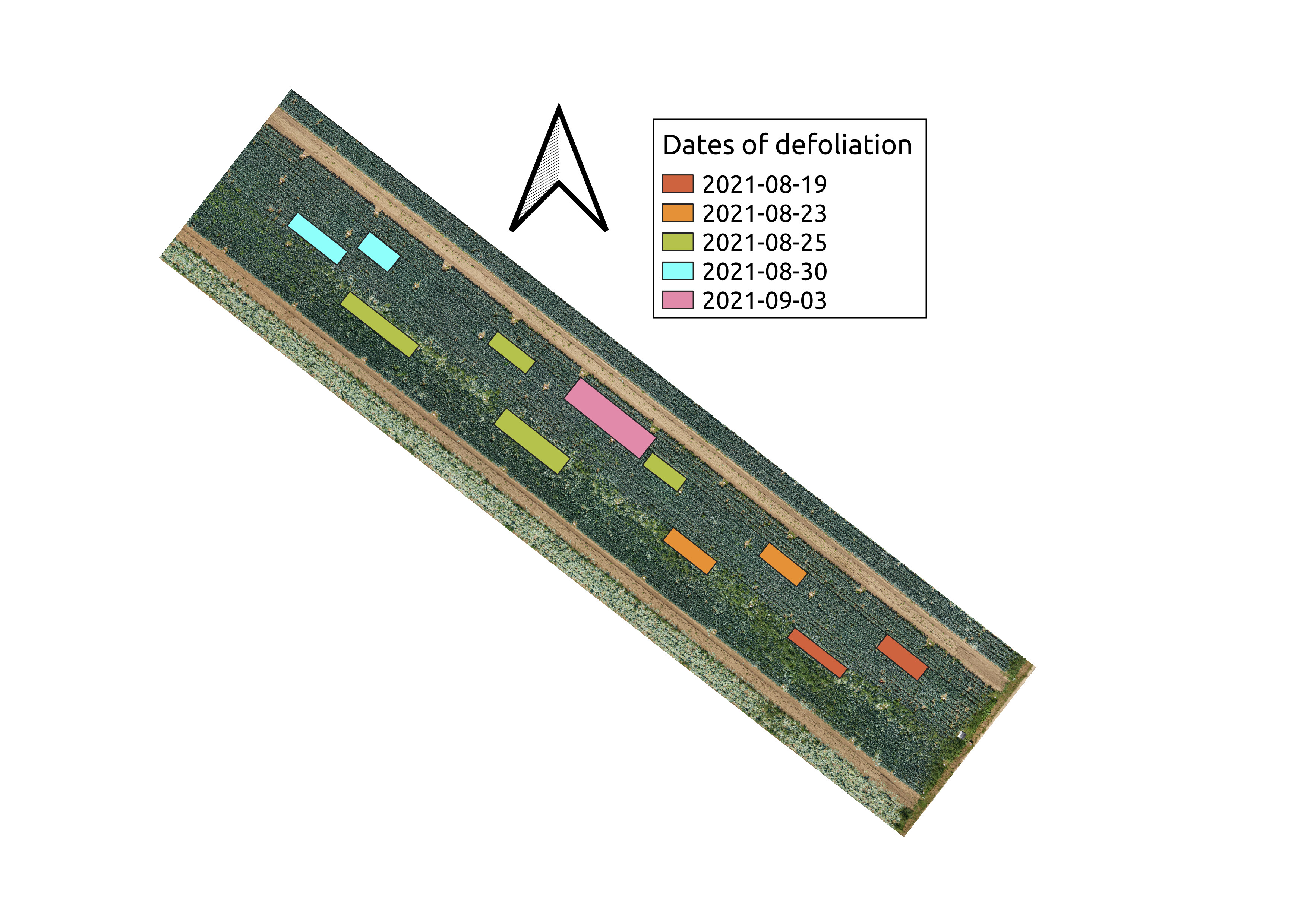}
        \label{fig:defoliation_map}}
    \end{minipage}
    \centering
    \begin{minipage}{0.38\textwidth}
     	\subfloat[Pre defoliation plant.]{
        \includegraphics[width=0.55\textwidth]{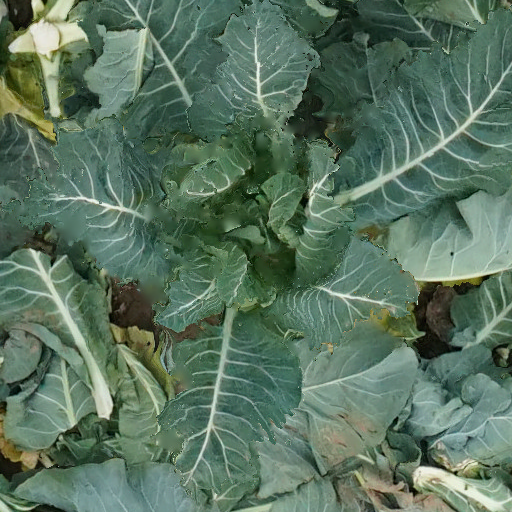}
        \label{fig:example_defoliation_pre}}
        
        \subfloat[Post defoliation plant.]{
         \includegraphics[width=0.55\textwidth]{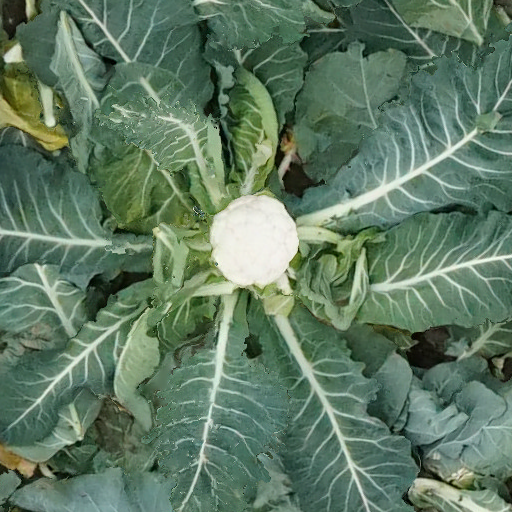}
         \label{fig:example_defoliation_post}}
     \end{minipage}
    \caption{Visual overview of our defoliated plant locations for the first five weeks of defoliation in field 2. Figures (b) and (c) show images of a plant pre and post defoliation. The locations of random distributed defoliated plants of week six are not shown.}
    \label{fig:example_defoliation}
\end{figure}

In addition to time series flights, so-called defoliation flights were conducted. For those images, the upper leaf layers covering the cauliflower head were manually removed for individual plants after the time series flight. This step is further referred to as defoliation. Care has been taken to ensure that the defoliated leaves do not affect neighboring plants. The defoliated plants give information about the development of the head in relation to the outer appearance of the plant. By performing another UAV flight after defoliation, a dataset of plants is given for which the time series of the outer appearance (\figref{fig:example_defoliation_pre}) of the plant in addition to the inner head (\figref{fig:example_defoliation_post}) on the day of defoliation is recorded.

For field 1, the defoliation of plants was performed on two days, October \nth{27} and \nth{29}, after harvesting took place. Thus, the defoliated plants represented plants, whose head size did not fulfill the quality criteria for harvest, which majorly meant that the head size was too small.
For field 2, starting on August, 19, when most of the cauliflower heads started developing, weekly between 70 and 200 plants were defoliated. All plants with developed heads were defoliated in rectangular plot regions to minimize the impact of defoliation on the biological growth of neighboring plants.
Care was taken not to defoliate the reference plants described in section in-situ measurements (\secref{sec:manualMeasurements}).
A distribution of plots for the first 5 defoliation time points is shown in \figref{fig:defoliation_map}. 
For the last overflight (after the last harvest), most of the remaining plants that had not yet been harvested were defoliated, which resulted in a random distribution and is therefore not shown in \figref{fig:defoliation_map}.

\subsection{Georeferenced Ground Control Points (GCPs)}
\begin{figure}[t]
	\centering
    \includegraphics[align=c,height=1in, width=0.15\textwidth]{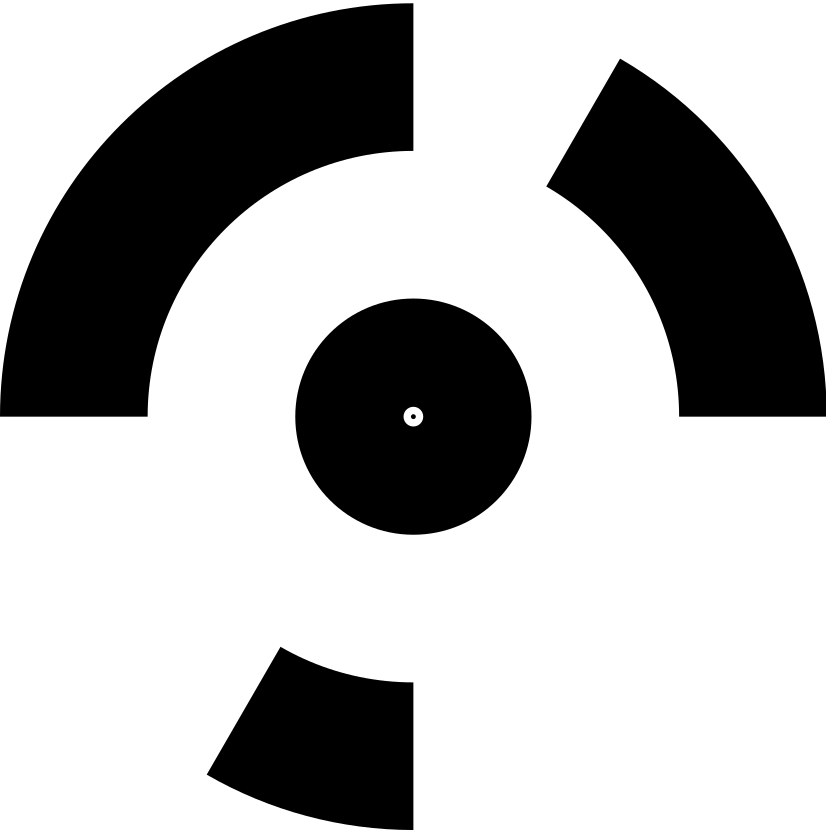}
    \hspace{10pt}
    \includegraphics[align=c,height=1in, width=0.15\textwidth]{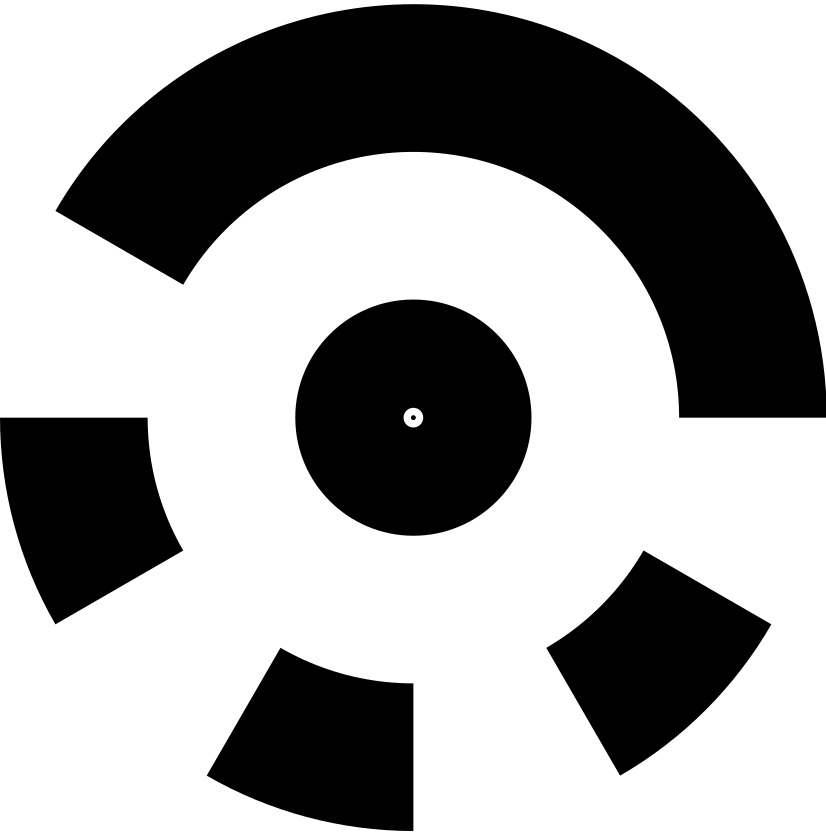}
	\caption{Two ground control point (GCP) patterns used for acquisition.}
    \label{fig:gcp}
\end{figure}

To localize the image data globally in space, the data were georeferenced with the help of circular 12-bit GCPs with a diameter of around \SI{20}{\cm}, as shown in \figref{fig:gcp}. The GCPs were fixed in the ground using plastic pegs. 
GCPs were evenly distributed across the field (see appendix \figref{fig:gcp_location}) and positioned on tractor tracks or between plants, to avoid displacement by external influences such as plowing. To ensure their visibility on image data, surrounding plants were removed where necessary.
We used 21 GCPs in field 1 (35 GCPs/ha) and 44 GCPs in field 2 (113 GCPs/ha) (see \figref{fig:gcp_location} in the appendix), with each GCP showing a different pattern. The greater number of GCPs in field 2 is due to the fact that they facilitate subsequent image alignment by ensuring at least 3 GCPs in each captured UAV image, especially for growth stages with a high degree of plant overlap and dense canopy. 

As measuring device, the Trimble R4-Model 3 Base station with a horizontal standard deviation of $\pm$ \SI{5}{\milli\meter} + 0.5 ppm RMS and vertical standard deviation of $\pm$ \SI{5}{\milli\meter} + 1 ppm RMS was used for both fields. In addition, the Trimble Juno slate controller was used. The measured coordinates are given in the coordinate system WGS84~/~UTM 32N.
To control that the markers for GCPs were not displaced due to external influences, the GCPs were measured at the beginning and end of the campaign to dismiss GCPs that were displaced. For field 2, a third measurement was added in the middle of the growing period.

\subsection{In-situ measurements of plant development}\label{sec:manualMeasurements}

\begin{figure}[t]
	\centering
    \subfloat[]{
    \includegraphics[trim = 87 12 87 0, clip, width=0.55\textwidth]{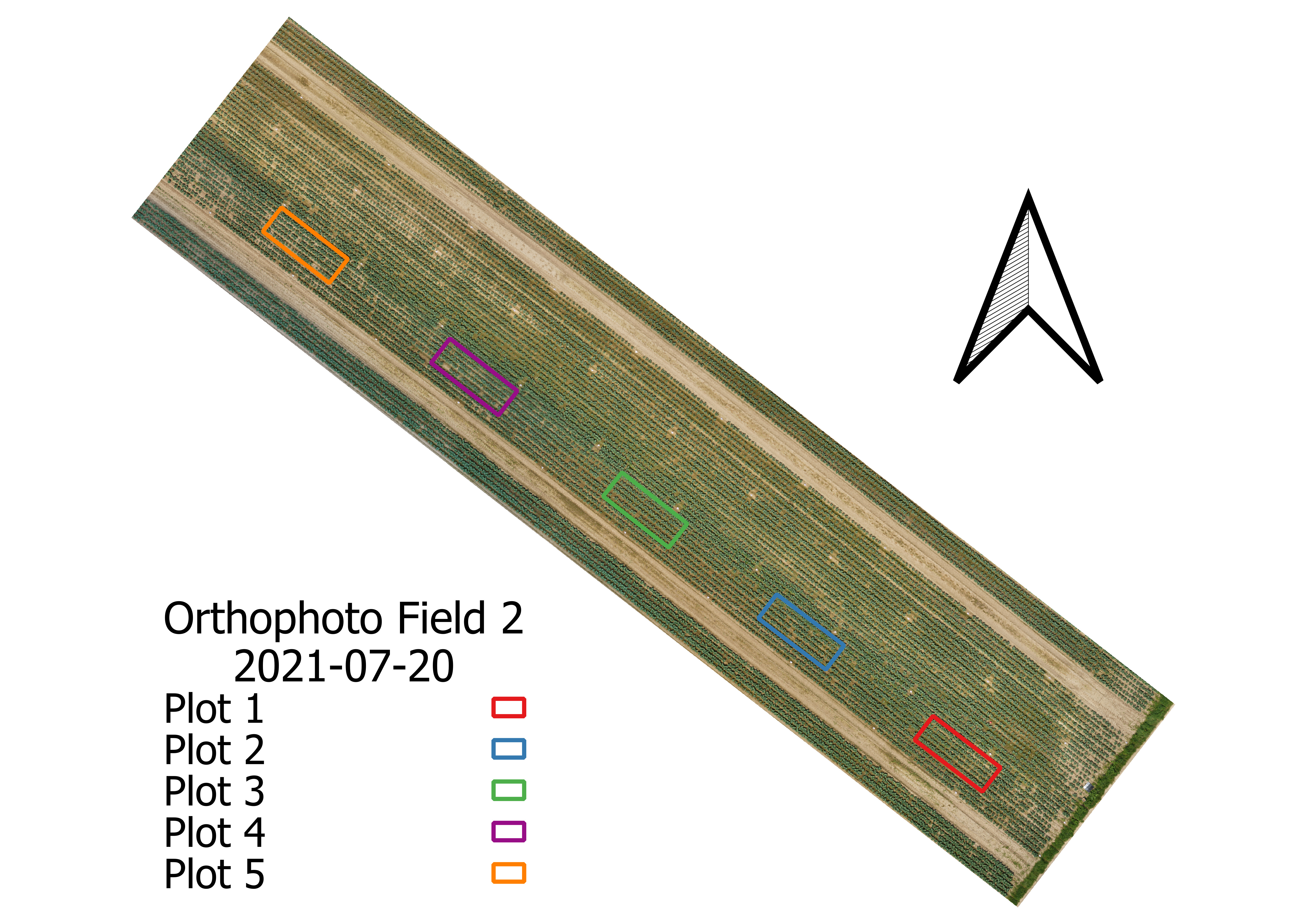}
    \label{fig:refPlots_field2_overview}}
    \hspace{10pt}
    \subfloat[]{
    \includegraphics[width=0.3\textwidth]{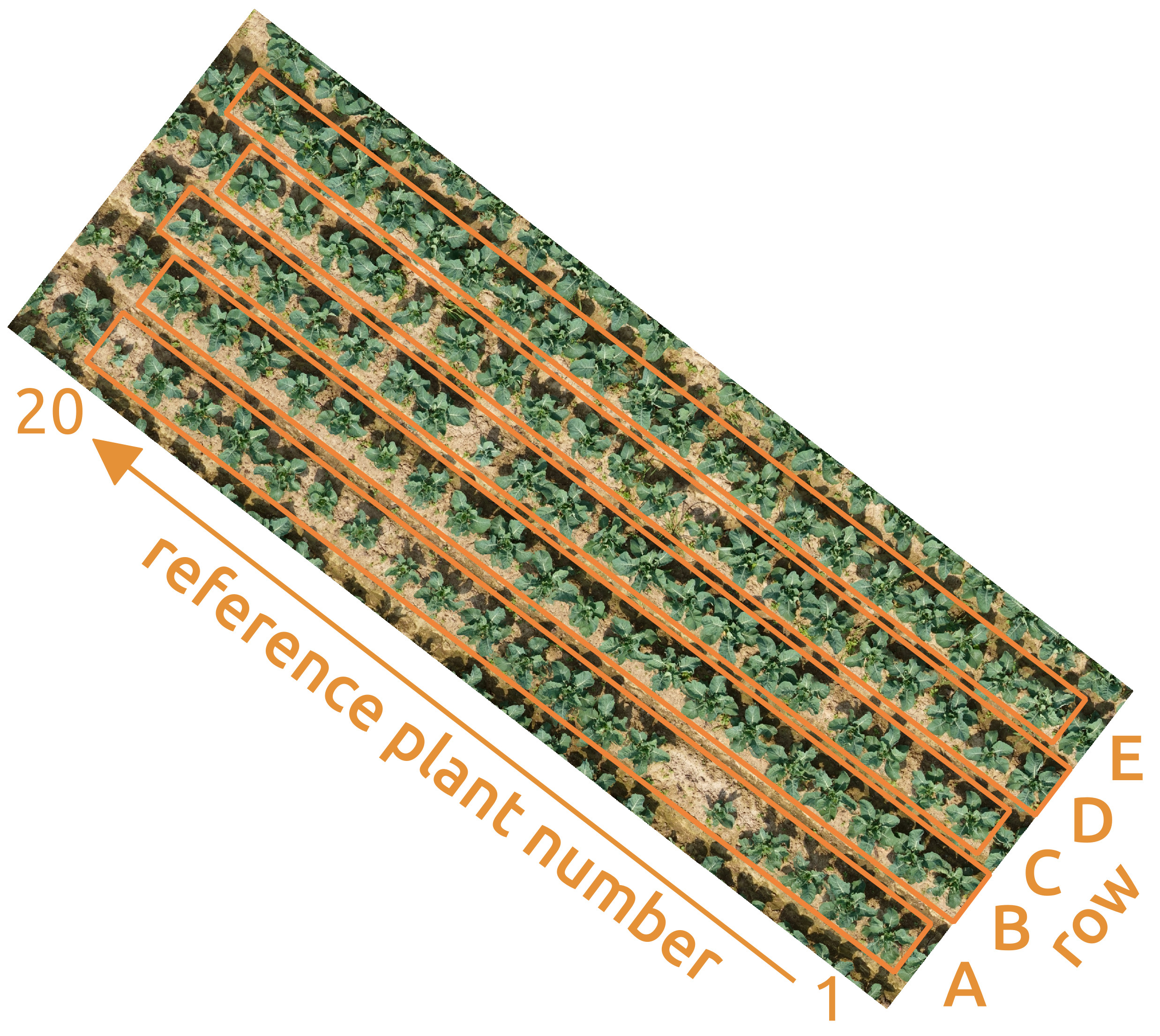}
     \label{fig:refPlot_field2}}
     
\caption{Visual overview of (a) reference plots for in-situ measurements within field 2
and (b) the respective design of reference plot 5 including reference plants and ordering of reference plant numbers. The plot design is valid for all reference plots of field 2.}
	\label{fig:refPlots_field2}
\end{figure}

In each field, so-called reference plots were selected where information about the plants, denoted as reference plants, located in these areas was manually captured.
For field 1, there are four reference plots (see appendix \figref{fig:refPlots_field1_overview}). Each plot consists of 3 rows with 20 plants each (see appendix \figref{fig:refPlot_field1}). Thus, there are 60 plants per plot, 240 plants in total. The plots are distributed in the north-western half along the long side of the field.
For field 2, there are five reference plots (\figref{fig:refPlots_field2_overview}). Each plot consists of 5 rows with 20 plants each (\figref{fig:refPlot_field2}). Thus, there are 100 plants per plot, 500 plants in total. The plots are evenly distributed in the south-western half along the long side of the field.
Thus, reference data are collected along the entire field. Each reference plant is assigned its specific plant ID, which consists of the row (Field 1: A-C; Field 2: A-E) and plant number
(Field 1: 1-10, 90-99; Field 2: 1-20).

For all reference plants of field 1, the following measurements were taken:
\begin{enumerate}
\item phenological development after BBCH/\cite{feller1995phenological}, 
\item height,
\item maximum diameter,
\item other remarks such as stress infestation (listed in the attachment in \tabref{table:stresses}),
\item head diameter, and
\item harvesting status.
\end{enumerate}
As the farmer pursued a rigorous plant protection schedule and hardly any stresses were detected in 2020, information about stresses was no longer explicitly recorded in 2021. Due to the observed homogeneous development, the focus was on measurements of BBCH and height of five representative plants per plot. Head diameter and harvest status were recorded for individual plants.

\section{Dataset}\label{sec:dataset}
\begin{figure}[t]
	\centering
    \includegraphics[width=\textwidth]{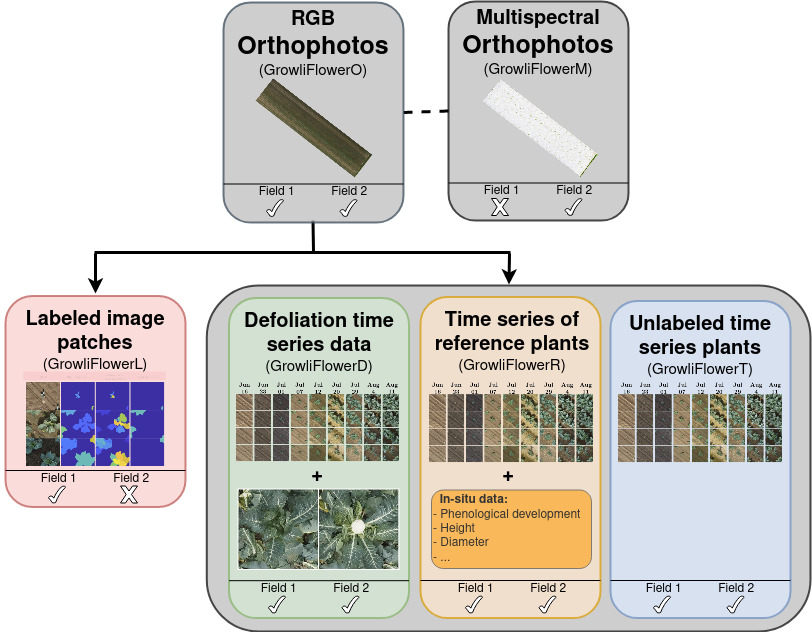}
    \caption{Overview over provided data in the GrowliFlower dataset.}
    \label{fig:data_overview}
\end{figure}

The basis of the dataset (\figref{fig:data_overview}) are RGB and multispectral orthophotos derived from captured UAV images. Single plants are identifiable via their coordinates and plant IDs in the orthophotos. 
The dataset contains four subsets intended for different machine learning tasks.
The instance segmentation subset GrowliFlowerL contains patches that are extracted and processed from the RGB orthophotos.  
The other three subsets contain time series of individual plants. 
The subset GrowliFlowerT comprises randomly selected time series representing a large variety of cauliflower developments.
The subset GrowliFlowerD contains additional image pairs of the plants before and after defoliation besides the time series. 
GrowliFlowerR contains in-situ measurements in addition to the time series.
For each field, a txt-file including measured coordinates of GCPs at the beginning and the end of the field monitoring is provided. For field two, coordinates measured during the growing period are also given.

\subsection{Orthophotos (GrowliFlowerO and GrowliFlowerM)}\label{sec:orthophotos}

The acquired RGB and multispectral UAV images are aligned to orthophotos using Agisoft Metashape Professional software to obtain a large-scale overview of the monitored fields. In combination with measured GCP coordinates, the orthophotos are georeferenced. 
The individual orthophotos are exported in the coordinate system WGS84 / UTM 32. 

The ground resolution for the RGB orthophotos of field 1 is \SI{1.65}{\milli\meter\per\px}, respectively for pixel width and height, with a minimum and maximum file storage size of 1.64 GB and 6.7 GB. The ground resolution for field 2 is \SI{3.10}{\milli\meter\per\px} with a minimum and maximum file storage size of 1.3 GB and 5.0 GB.
12 orthophotos are available for field 1, where 5 are processed entirely and 7 contain data gaps for small areas where the UAV images' quality was not high enough.
For field 2, 15 orthophotos are available, as shown in \figref{fig:timeline_field2}. 
The set of orthophotos denoted as GrowliFlowerO is provided.
Additionally, the dataset contains multispectral orthophotos for field 2 with a ground resolution of \SI{2.5}{\centi\meter} per pixel width and length, denoted as GrowliFlowerM.

\subsection{RGB image patches}
The data described in this section is extracted from the RGB orthophotos. The ground resolution of the resulting image patches is the same as for the respective orthophotos. 

Each of the following described datasets (excluding the labeled dataset in \secref{sec:labeledData}) contains a txt-file with global information for each field, containing the image ID, including the plant ID, and the corresponding georeferenced UTM coordinate for the plants. The coordinates refer to the center of the plants observed on August, \nth{19} for field 1 and July, \nth{7} for field 2. Additionally, information about the planting day and a proposed assignment to training, validation, or test subset is given as the basis for the comparison of machine learning methods. 
The proposed training, validation, and testing subsets are spatially disjoint to minimize spatial correlation between sets. Yet, certain systematic factors from a biological point of view are not excluded. The use of these sets is expected to promote the development of machine learning methods with high generalization ability.
For reference data, presented in \secref{sec:GrowliFlowerR}, the harvesting time is specified and for defoliation data, presented in \secref{sec:GrowliFlowerD}, the defoliation date of the plants. 
Furthermore, txt-files with local information for each acquisition date are provided, which contain the image ID to connect the local information with the global information, and the corresponding local pixel coordinate with respect to the respective orthophoto for each data acquisition day. Moreover, information about the day after planting (dap) is added. 

For the use of image patches showing single plants, patches have to be extracted out of the orthophotos using plant IDs and coordinates. An image side length and width of at least \SI{490}{\px} for field 1 and of at least \SI{256}{\px} for field 2 is recommended to ensure that regardless of plant developmental stage, the whole plant is fully captured on the image patch.

\subsubsection{Labeled image patches (GrowliFlowerL)}\label{sec:labeledData}
This subset consists of pixel-wise, manually annotated images and is well suited for tasks like classification, semantic segmentation, detection, instance segmentation, or stem detection.
For this set, image patches of four acquisition dates of field 1 are extracted by a sliding window approach. 
The image patches have a size of $\SI{368}{\px} \times \SI{448}{\px}$. The size of the patches varies from the proposed sizes, as only plants from the earlier stages of development are included. Furthermore, in this dataset, the focus is not on individual plants but on variability between images, so that the plants are not located in the center of the patch either.

For each RGB image patch, four annotated masks are provided which contain segmentations of: (1) plant instances, (2) leaf instances and void instances, (3) void segmentations, and (4) stem positions.  

\begin{enumerate}[label=(\arabic*)]
    \item The plant instance mask segment the image in soil and plant pixels with instance information for plants.
    \item The leaf instance mask segment the plants in their single leaves. Plants at image borders for which no stem or only a quarter of the plant is visible are annotated as void and no leaf annotation is done.
    \item The mask, including void segmentations, is a binary mask where only plants are segmented as void located at image borders where no stem is visible, or only a small amount of leaves is visible in the RGB image.
    \item The stem annotation mask represents the position of the stems of non-void plants.
\end{enumerate}

\begin{figure}[t]
	\centering
	 \begin{minipage}{0.24\textwidth}
	 \centering
	 \textbf{RGB}
	 \end{minipage}
	 \begin{minipage}{0.24\textwidth}
	 	 \centering
	 \textbf{Plant instances}
	 \end{minipage}
	 \begin{minipage}{0.24\textwidth}
	 	 \centering
	 \textbf{Leaf instances \\ + void instances}
	 \end{minipage}
	  \begin{minipage}{0.24\textwidth}
	 	 \centering
	 \textbf{Void instances}
	 \end{minipage}

    \vspace{6pt}
    \includegraphics[width=0.24\textwidth]{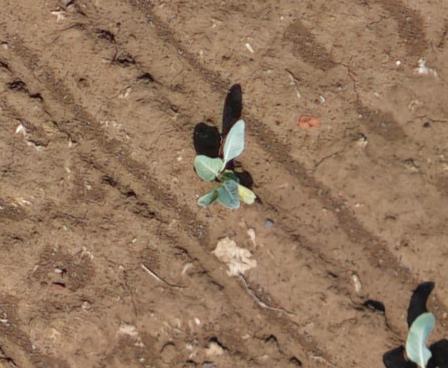} 
    \includegraphics[width=0.24\textwidth]{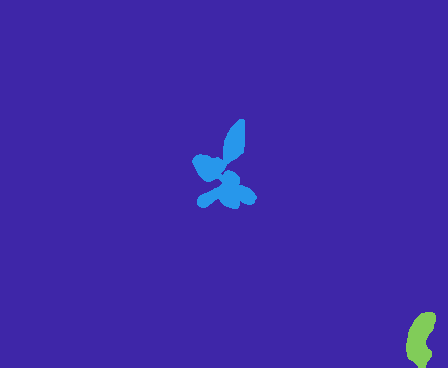}
    \includegraphics[width=0.24\textwidth]{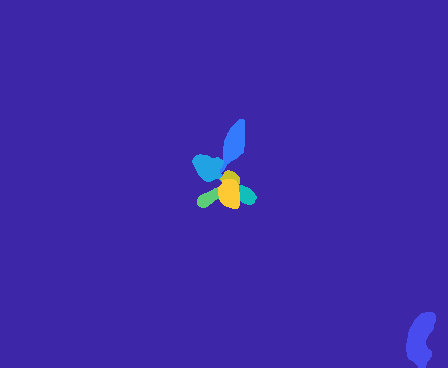} 
     \includegraphics[width=0.24\textwidth]{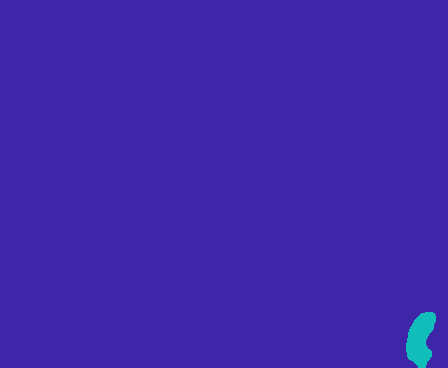}

    \includegraphics[width=0.24\textwidth]{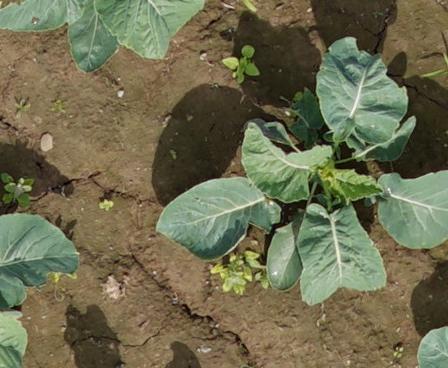} 
    \includegraphics[width=0.24\textwidth]{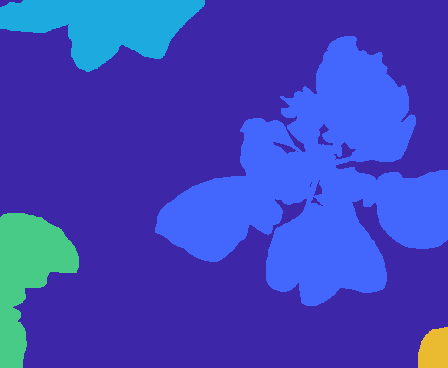}
    \includegraphics[width=0.24\textwidth]{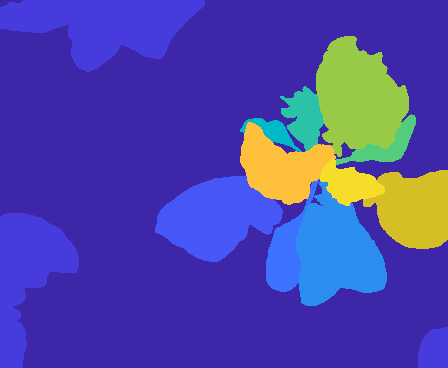} 
     \includegraphics[width=0.24\textwidth]{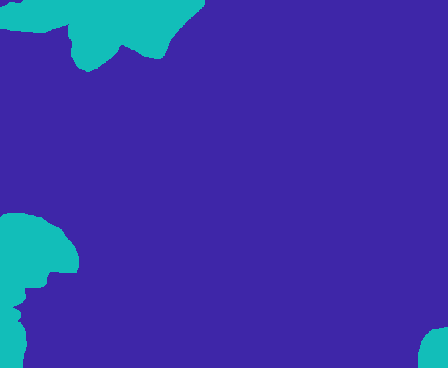}

    \includegraphics[width=0.24\textwidth]{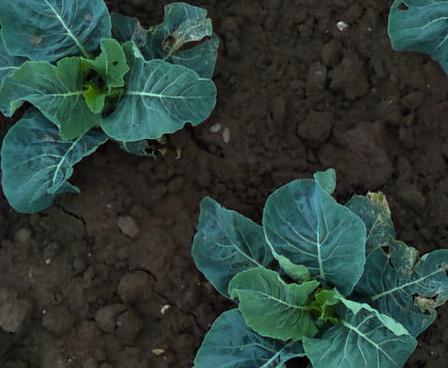} 
    \includegraphics[width=0.24\textwidth]{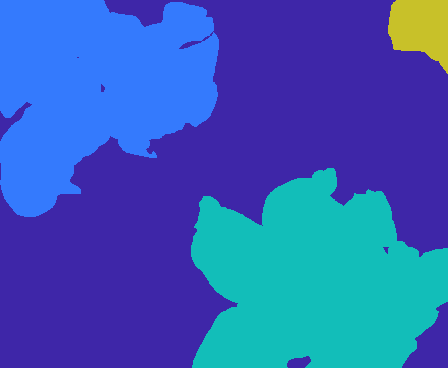}
    \includegraphics[width=0.24\textwidth]{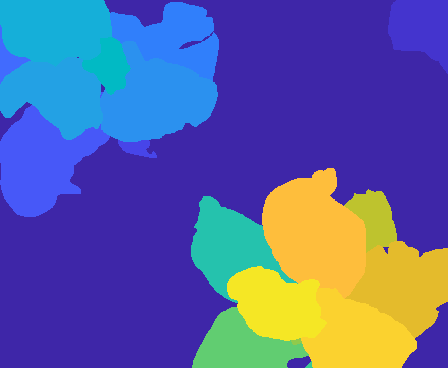} 
     \includegraphics[width=0.24\textwidth]{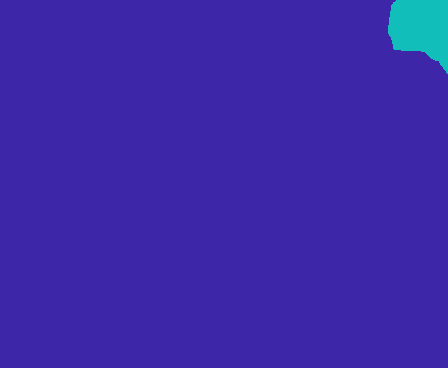}

 	\caption{Illustration of labeled image examples for different time points. Column 1 shows the RGB base for in columns 2-4 illustrated corresponding labeled plant instance masks, leaf instance masks in addition to void labels and only void instance masks. The different rows represent different points in time. Dark blue indicates the background class while other colors illustrate different (leaf) instances.}
	\label{fig:rgb_label_image}
\end{figure} 

Examples of masks (1), (2) and (3) are shown in \figref{fig:rgb_label_image}.
Since the stem positions are only represented by individual pixels and are therefore not easy to recognize visually, the masks including stem information are not shown.
The annotations are provided with a defined name dependent on the RGB image patch name. Each patch contains a maximum of four plants and there is also a number of patches in the dataset containing no plants, see \tabref{table:OPTIKOlabeled}. The dataset is divided into a training, validation, and test set. The complete labeled dataset is introduced as \text{GrowliFlowerL}.

\begin{table}[t]
    \caption{Overview of the distribution of the labeled images to the different acquisition dates.}\label{table:OPTIKOlabeled}
    \begin{center}
    \begin{tabular}{|c|c|c|c|}
        \hline
         \multirow{2}{*}{\textbf{Definition}} &  \textbf{All images} & \textbf{Images with plants} &  \textbf{Images without plants}\\
        &  &  [\textit{Train/Val/Test}]  & [\textit{Train/Val/Test}]\\
        \hline
        2020\_08\_12 & 844 & 745 \textit{[522/111/112]} & 99 \textit{[70/14/15]}\\
        2020\_08\_19 & 892 & 781 \textit{[547/117/117]} & 111 \textit{[78/16/17]}\\
        2020\_08\_25 & 383 & 367 \textit{[257/55/55]} & 16 \textit{[12/2/2]}\\
        2020\_09\_08 & 79 & 79 \textit{[56/11/12]} & 0 \textit{[0/0/0]}\\
        \hline
    \end{tabular}
    \end{center}
\end{table}

\subsubsection{Time series for plant data (GrowliFlowerT)}\label{sec:GrowliFlowerT}

For each field, plant coordinates are determined to enable the dataset user to extract plant image time series.
This data is denoted with GrowliFlowerT. Time series of field 1 comprise the early plant developmental stages as well as the harvest dates, but lack dates, when the canopy was closed. Time series of field 2 comprise all growth stages.

For field 1, coordinates for about a third of the plants in the field are determined, which are 3804 plants in total. The distribution of the location of the extracted data is visually seen in \figref{fig:random_TrainValTest_field1} in the appendix. The chosen plants are distributed along the southeastern edge of the field due to the availability of data for most time points and the possibility to determine the harvest window of individual plants. October, \nth{19}, one week before harvest, is the only day no image data for any plant is available for GrowliFlowerT.
The dataset is divided into a training, validation, and test set as shown in \figref{fig:random_TrainValTest_field1} in the appendix. In addition, it is ensured that cauliflower planted on July, \nth{28} or July, \nth{29} is included in all three sets.
Since the orthophotos do not overlap completely, image data is not available for all plants at all times. This leads to temporal incomplete time series. 
For field 2, 8736 coordinates of plants were extracted, evenly distributed over the field. The dataset is divided into training, validation, and test set as shown in \figref{fig:random_TrainValTest_field2} in the appendix. All plant coordinates are provided as georeferenced UTM coordinates.

\begin{figure}[t]
	\centering
	 \begin{minipage}{0.082\textwidth}
	 \centering
	 \textbf{Aug 12}
	 \end{minipage}
	 \begin{minipage}{0.082\textwidth}
	 	 \centering
	 \textbf{Aug 19}
	 \end{minipage}
	 \begin{minipage}{0.082\textwidth}
	 	 \centering
	 \textbf{Aug 25}
	 \end{minipage}
	 \begin{minipage}{0.082\textwidth}
	 	 \centering
	 \textbf{Sept 2}
	 \end{minipage}
	 \begin{minipage}{0.082\textwidth}
	 	 \centering
	 \textbf{Sept 8}
	 \end{minipage}
	 \begin{minipage}{0.082\textwidth}
	 	 \centering
	 \textbf{Sept 17}
	 \end{minipage}
	 \begin{minipage}{0.082\textwidth}
	 	 \centering
	 \textbf{Sept 22}
	 \end{minipage}
	 \begin{minipage}{0.082\textwidth}
	 	 \centering
	 \textbf{Oct 6}
	 \end{minipage}
	 \begin{minipage}{0.082\textwidth}
	 	 \centering
	 \textbf{Oct 27}
	 \end{minipage}
	 \begin{minipage}{0.082\textwidth}
	 	 \centering
	 \textbf{Oct 29}
	 \end{minipage}
	 \begin{minipage}{0.082\textwidth}
	 \centering
	 \textbf{Nov 2}
	 \end{minipage}

    \includegraphics[width=0.082\textwidth]{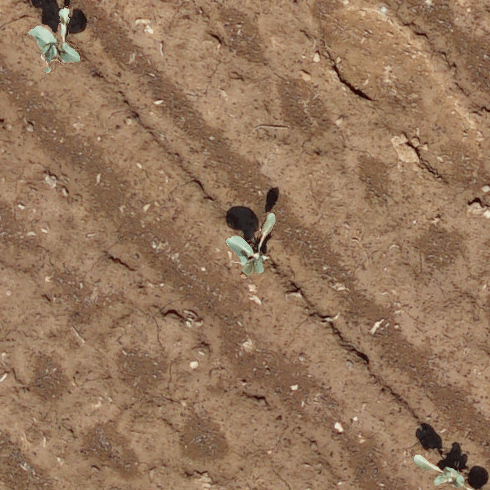}
    \includegraphics[width=0.082\textwidth]{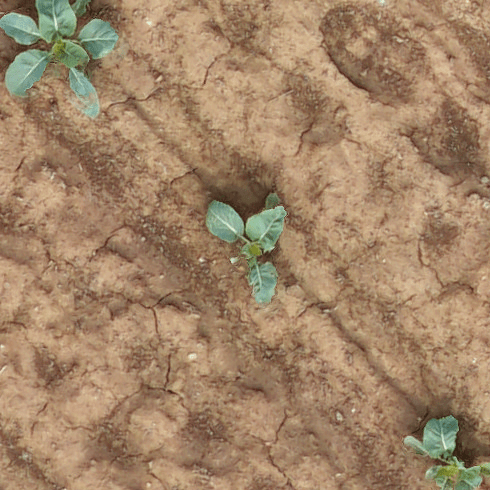}
    \includegraphics[width=0.082\textwidth]{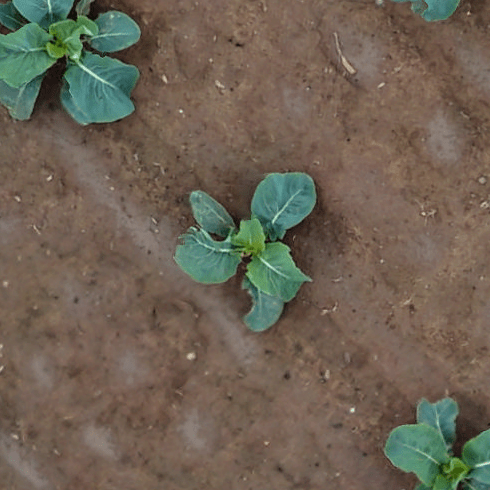}
    \includegraphics[width=0.082\textwidth]{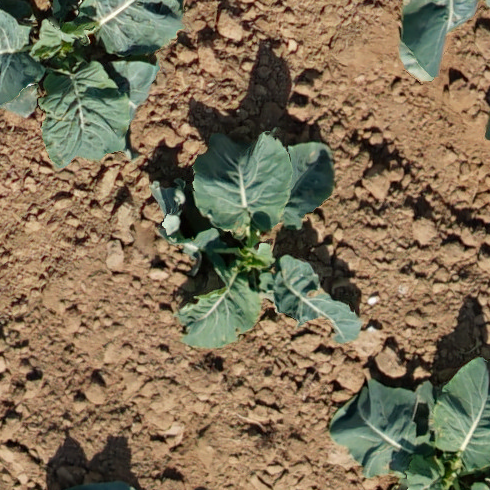}
    \includegraphics[width=0.082\textwidth]{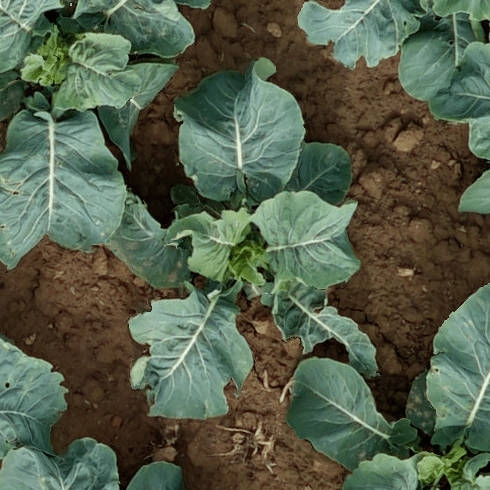}
    \includegraphics[width=0.082\textwidth]{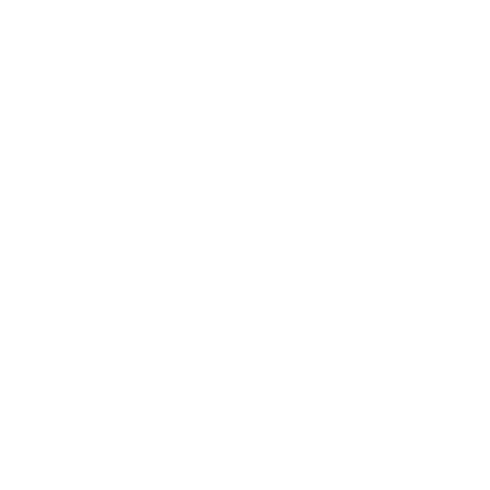}
    \includegraphics[width=0.082\textwidth]{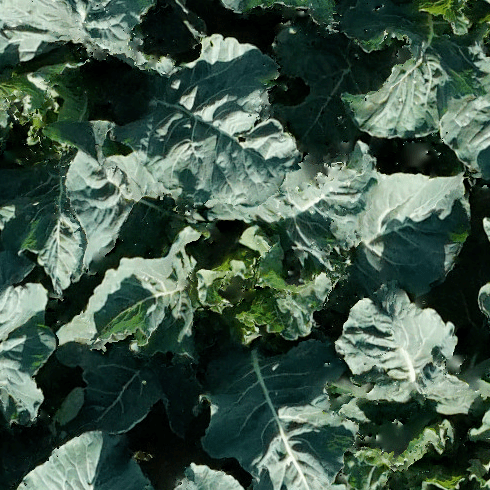}
    \includegraphics[width=0.082\textwidth]{ts_placeholder.png}
    \includegraphics[width=0.082\textwidth]{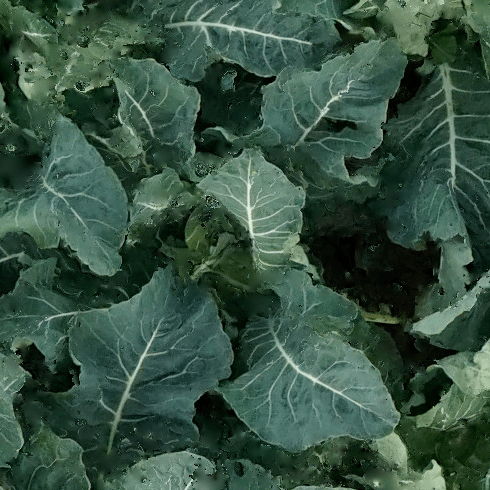}
    \includegraphics[width=0.082\textwidth]{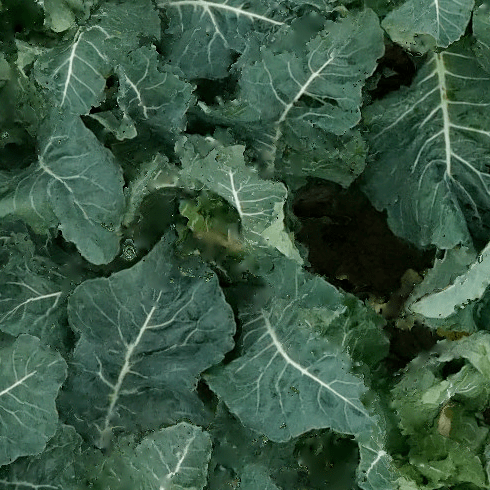}
    \includegraphics[width=0.082\textwidth]{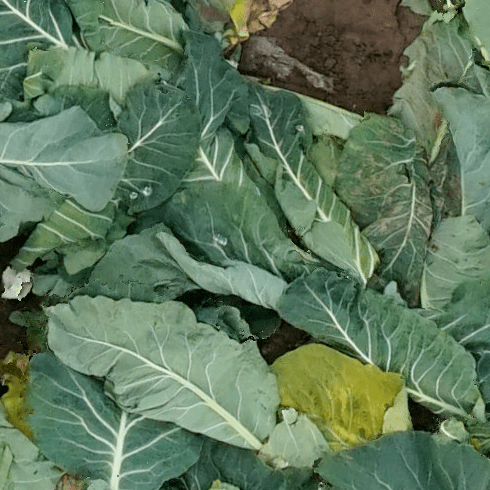}

    \includegraphics[width=0.082\textwidth]{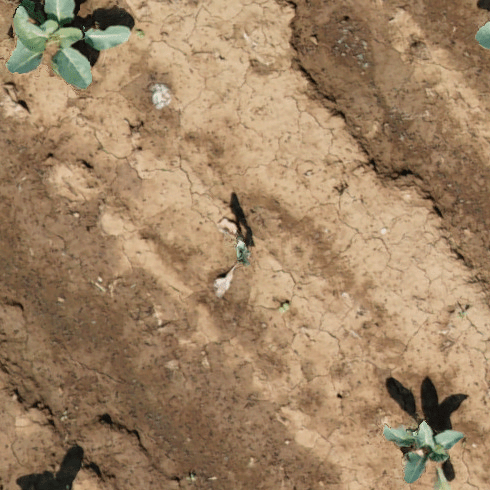}
    \includegraphics[width=0.082\textwidth]{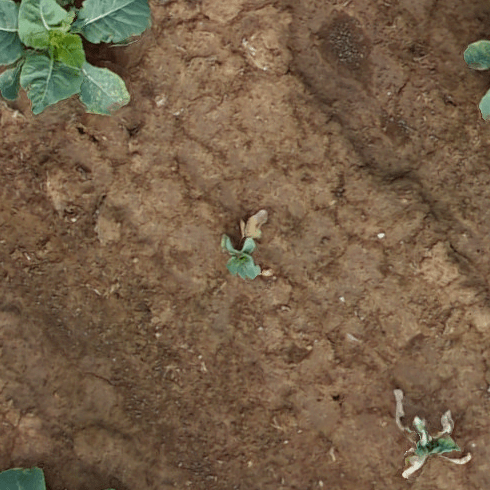}
    \includegraphics[width=0.082\textwidth]{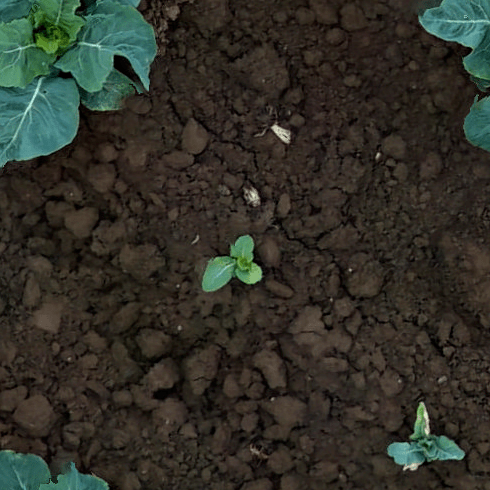}
    \includegraphics[width=0.082\textwidth]{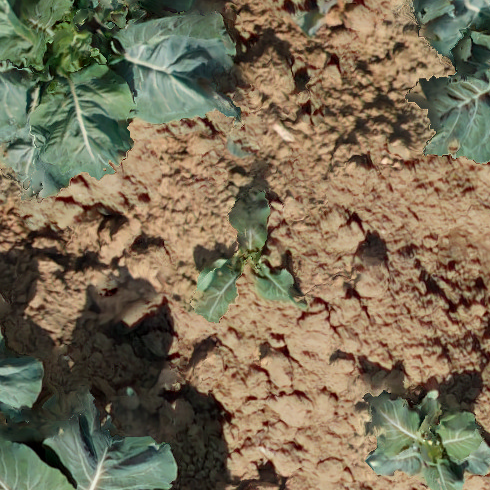}
    \includegraphics[width=0.082\textwidth]{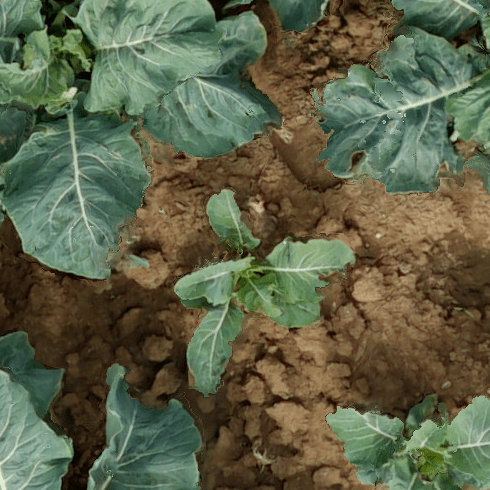}
    \includegraphics[width=0.082\textwidth]{ts_placeholder.png}
    \includegraphics[width=0.082\textwidth]{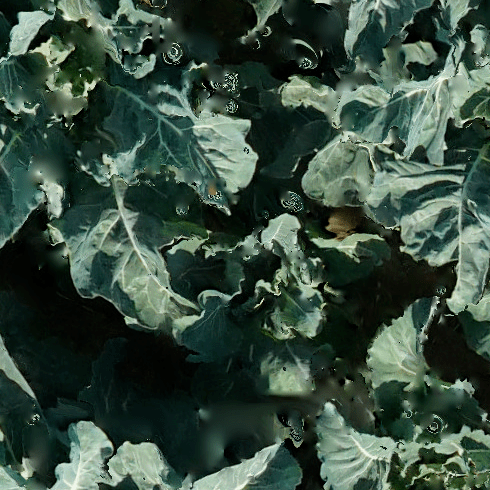}
    \includegraphics[width=0.082\textwidth]{ts_placeholder.png}
    \includegraphics[width=0.082\textwidth]{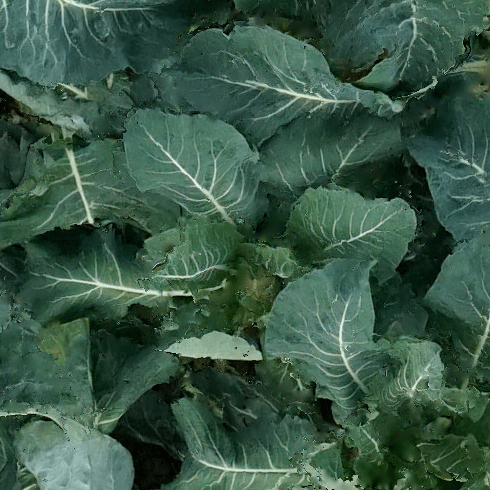}
    \includegraphics[width=0.082\textwidth]{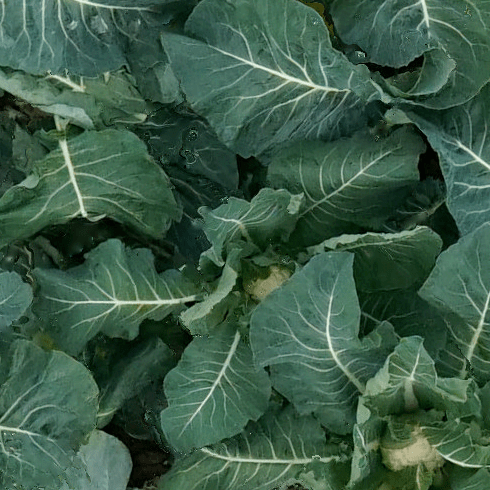}
    \includegraphics[width=0.082\textwidth]{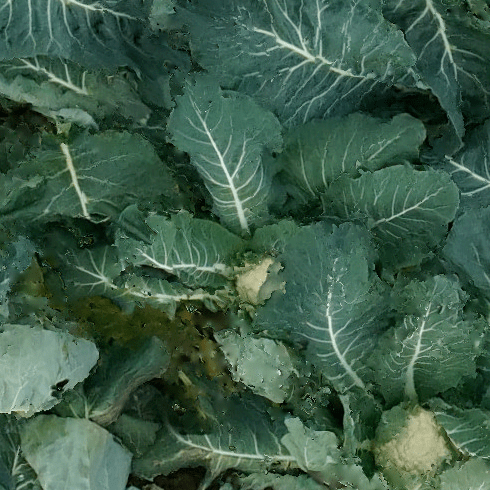}

    \includegraphics[width=0.082\textwidth]{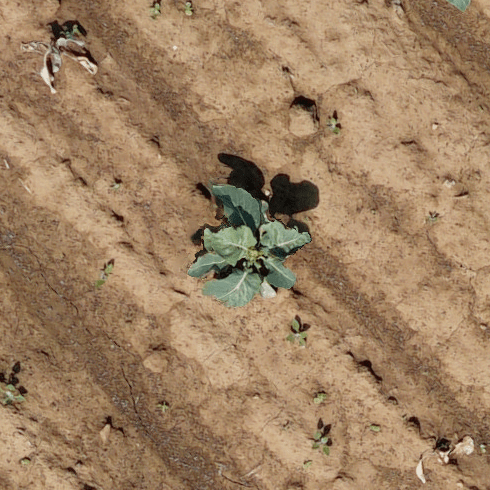}
    \includegraphics[width=0.082\textwidth]{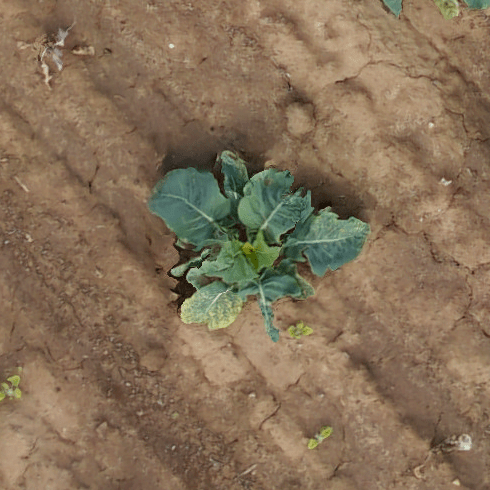}
    \includegraphics[width=0.082\textwidth]{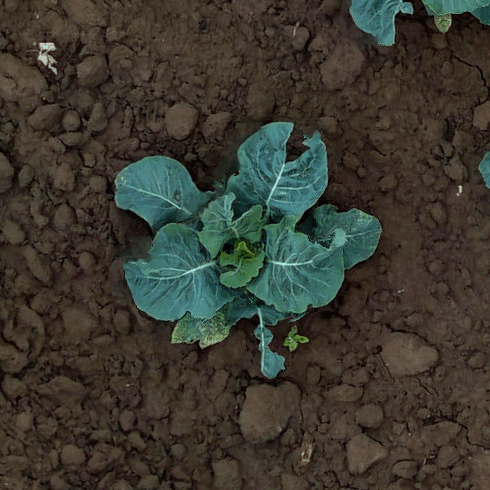}
    \includegraphics[width=0.082\textwidth]{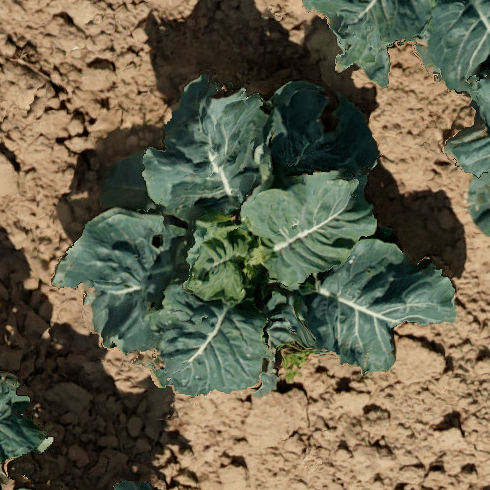}
    \includegraphics[width=0.082\textwidth]{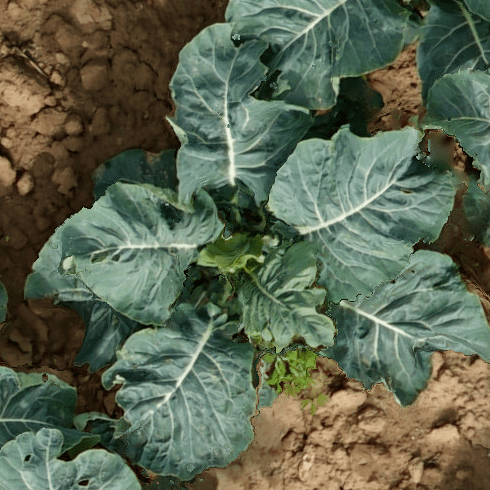}
    \includegraphics[width=0.082\textwidth]{ts_placeholder.png}
    \includegraphics[width=0.082\textwidth]{ts_placeholder.png}
    \includegraphics[width=0.082\textwidth]{ts_placeholder.png}
    \includegraphics[width=0.082\textwidth]{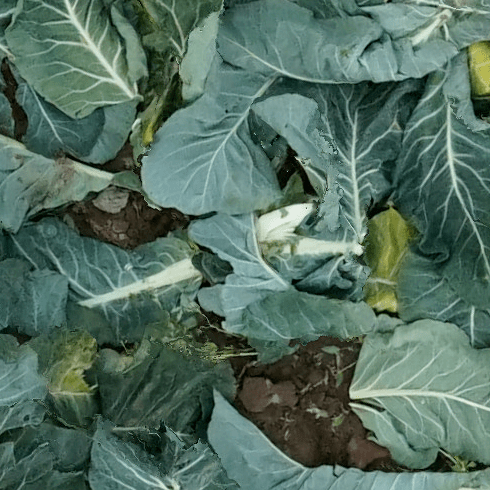}
    \includegraphics[width=0.082\textwidth]{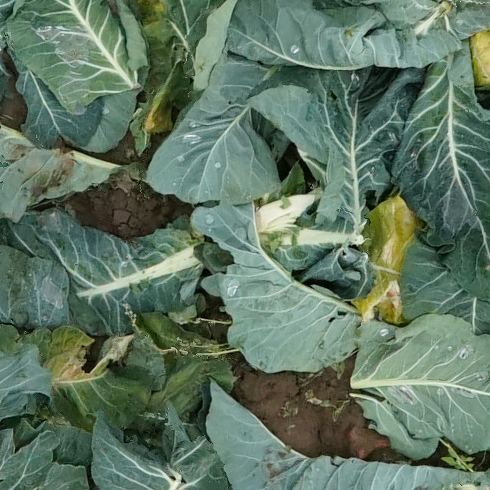}
    \includegraphics[width=0.082\textwidth]{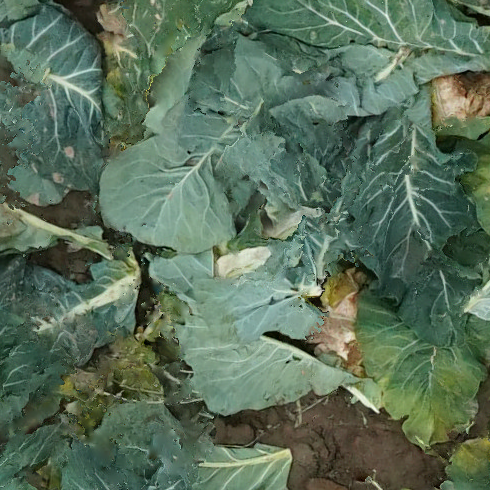}

    \includegraphics[width=0.082\textwidth]{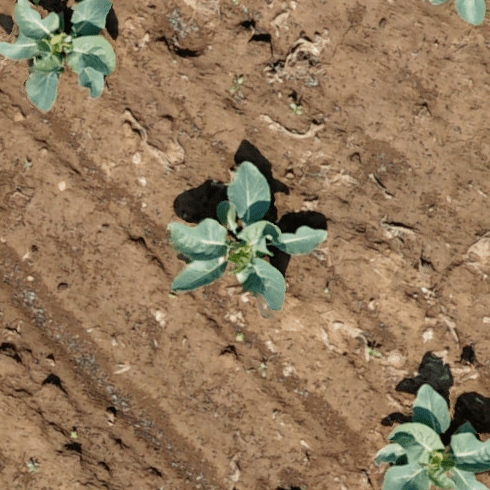}
    \includegraphics[width=0.082\textwidth]{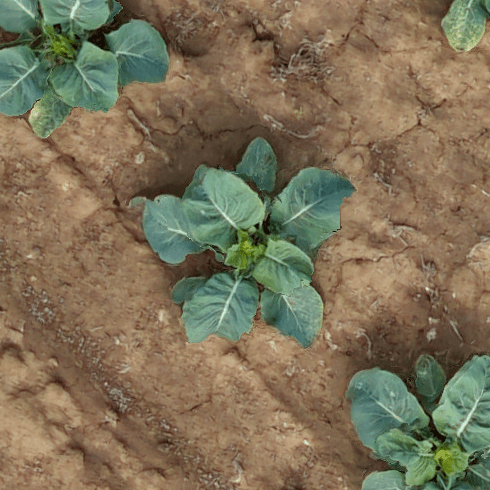}
    \includegraphics[width=0.082\textwidth]{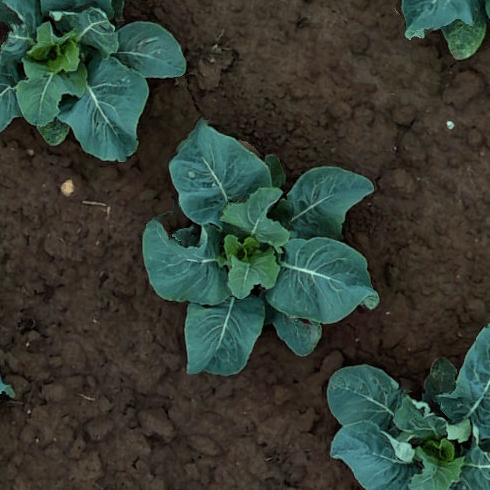}
    \includegraphics[width=0.082\textwidth]{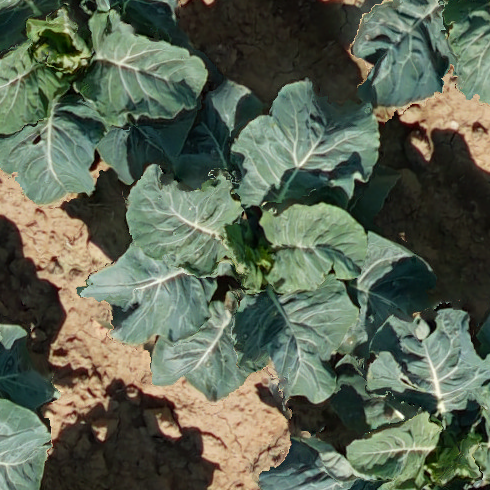}
    \includegraphics[width=0.082\textwidth]{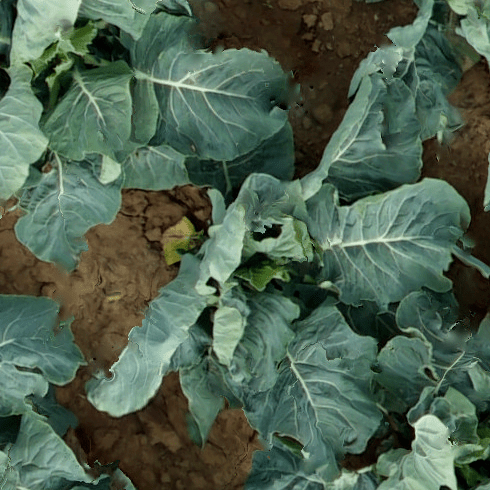}
    \includegraphics[width=0.082\textwidth]{ts_placeholder.png}
    \includegraphics[width=0.082\textwidth]{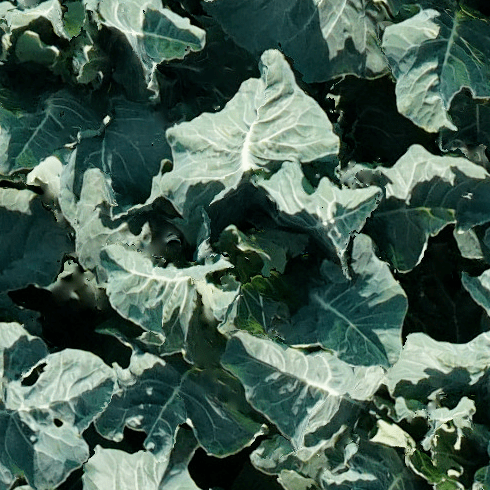}
    \includegraphics[width=0.082\textwidth]{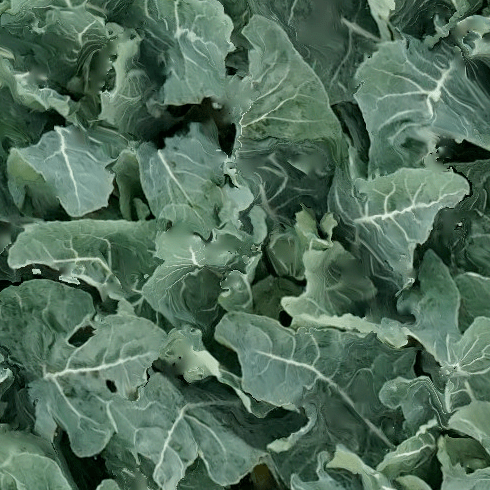}
    \includegraphics[width=0.082\textwidth]{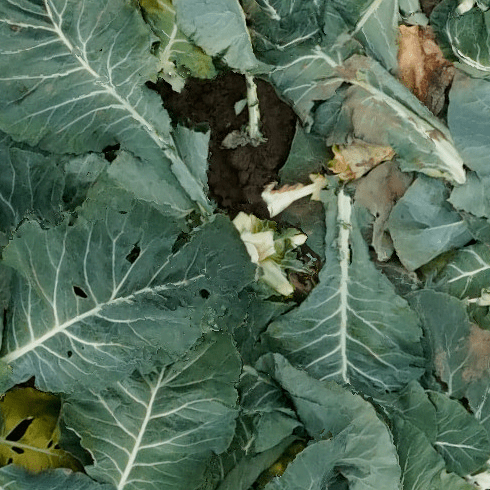}
    \includegraphics[width=0.082\textwidth]{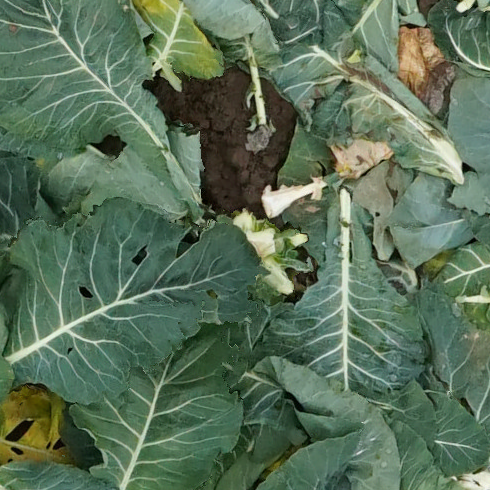}
    \includegraphics[width=0.082\textwidth]{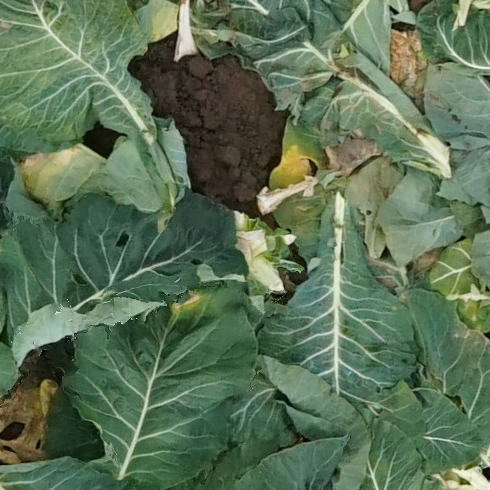}

    \includegraphics[width=0.082\textwidth]{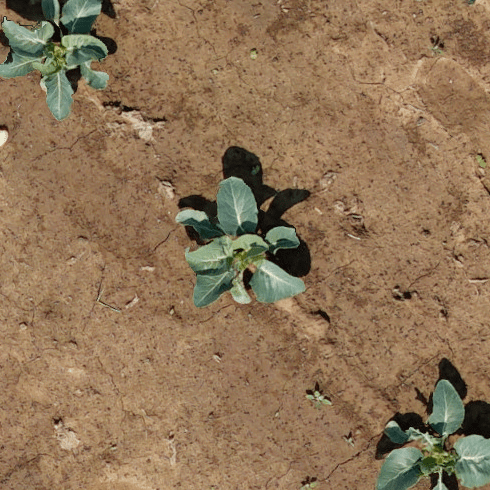}
    \includegraphics[width=0.082\textwidth]{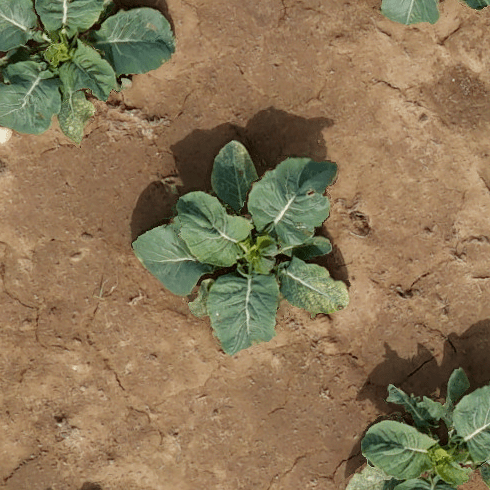}
    \includegraphics[width=0.082\textwidth]{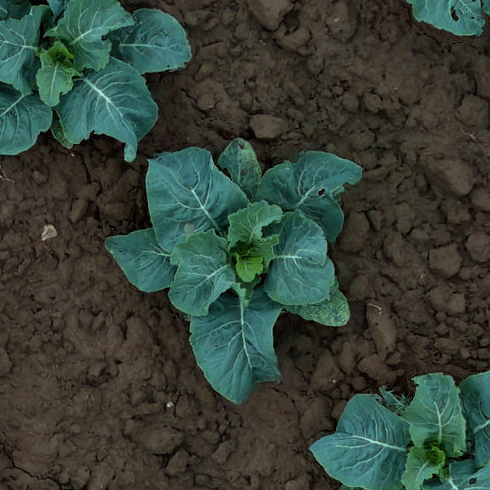}
    \includegraphics[width=0.082\textwidth]{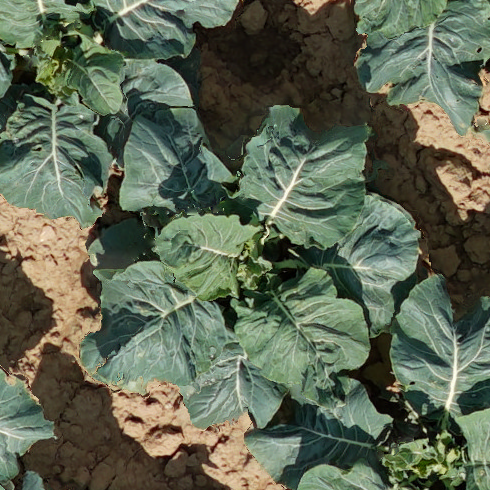}
    \includegraphics[width=0.082\textwidth]{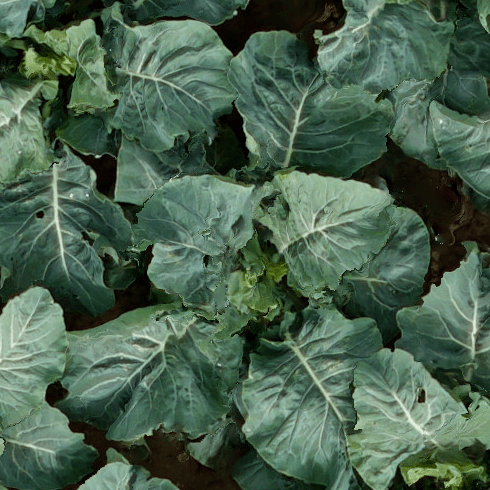}
    \includegraphics[width=0.082\textwidth]{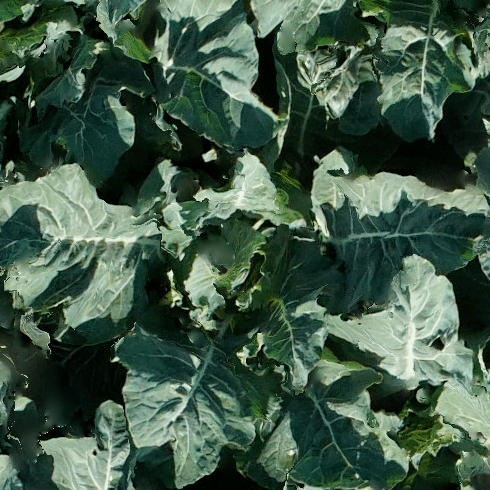}
    \includegraphics[width=0.082\textwidth]{ts_placeholder.png}
    \includegraphics[width=0.082\textwidth]{ts_placeholder.png}
    \includegraphics[width=0.082\textwidth]{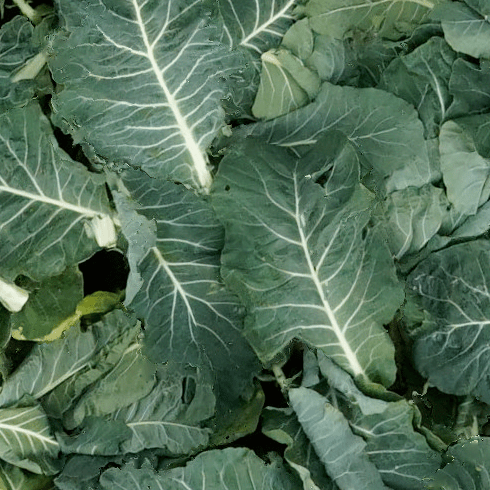}
    \includegraphics[width=0.082\textwidth]{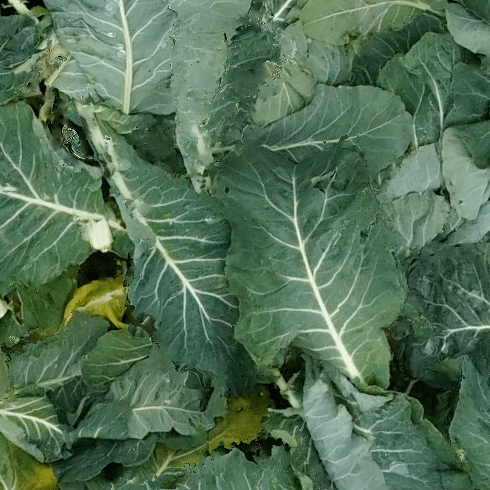}
    \includegraphics[width=0.082\textwidth]{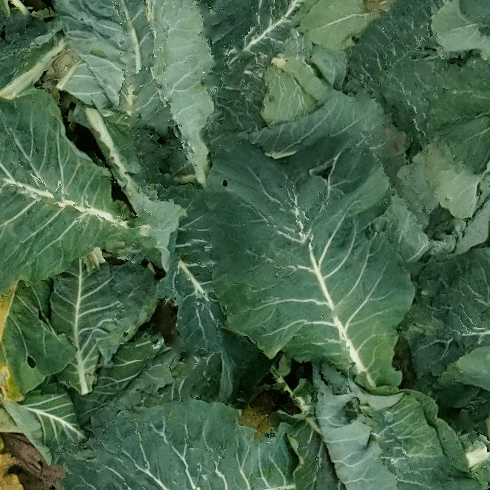}

\caption{Five plant time series of GrowliFlowerT within field 1. A row represents a time series including temporal data gaps. The different columns represent the acquisition dates.} 
	\label{fig:timeSeries_others_field1}
\end{figure}

\begin{figure}[t]
	\centering
	 \begin{minipage}{0.06\textwidth}
	 \centering
	 \textbf{Jun 16}
	 \end{minipage}
	 \begin{minipage}{0.06\textwidth}
	 	 \centering
	 \textbf{Jun 23}
	 \end{minipage}
	 \begin{minipage}{0.06\textwidth}
	 	 \centering
	 \textbf{Jul 01}
	 \end{minipage}
	 \begin{minipage}{0.06\textwidth}
	 	 \centering
	 \textbf{Jul 07}
	 \end{minipage}
	 \begin{minipage}{0.06\textwidth}
	 	 \centering
	 \textbf{Jul 12}
	 \end{minipage}
	 \begin{minipage}{0.06\textwidth}
	 	 \centering
	 \textbf{Jul 20}
	 \end{minipage}
	 \begin{minipage}{0.06\textwidth}
	 	 \centering
	 \textbf{Jul 29}
	 \end{minipage}
	 \begin{minipage}{0.06\textwidth}
	 	 \centering
	 \textbf{Aug 04}
	 \end{minipage}
	 \begin{minipage}{0.06\textwidth}
	 \centering
	 \textbf{Aug 11}
	 \end{minipage}
	 \begin{minipage}{0.06\textwidth}
	 \centering
	 \textbf{Aug 19}
	 \end{minipage}
	 \begin{minipage}{0.06\textwidth}
	 \centering
	 \textbf{Aug 23}
	 \end{minipage}
	 \begin{minipage}{0.06\textwidth}
	 \centering
	 \textbf{Aug 25}
	 \end{minipage}
	 \begin{minipage}{0.06\textwidth}
	 \centering
	 \textbf{Aug 30}
	 \end{minipage}
	 \begin{minipage}{0.06\textwidth}
	 \centering
	 \textbf{Sept 03}
	 \end{minipage}
	 \begin{minipage}{0.06\textwidth}
	 \centering
	 \textbf{Sept 08}
	 \end{minipage}

    \includegraphics[width=0.06\textwidth]{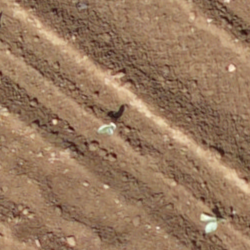}
    \includegraphics[width=0.06\textwidth]{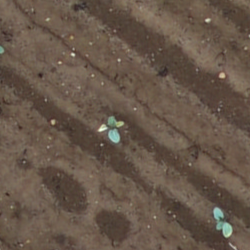}
    \includegraphics[width=0.06\textwidth]{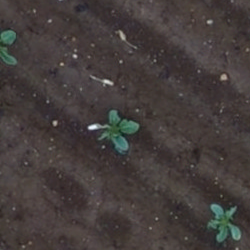}
    \includegraphics[width=0.06\textwidth]{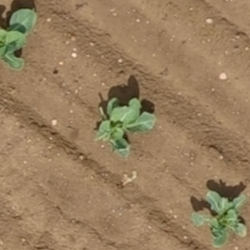}
    \includegraphics[width=0.06\textwidth]{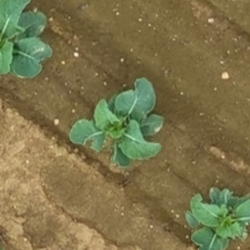}
    \includegraphics[width=0.06\textwidth]{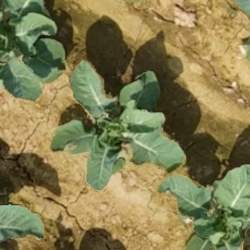}
    \includegraphics[width=0.06\textwidth]{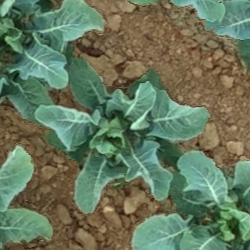}
    \includegraphics[width=0.06\textwidth]{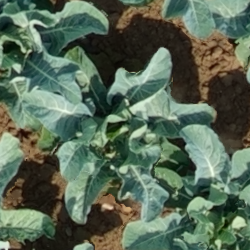}
     \includegraphics[width=0.06\textwidth]{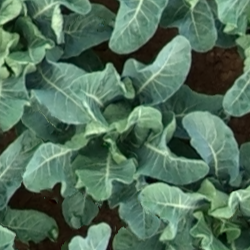}
     \includegraphics[width=0.06\textwidth]{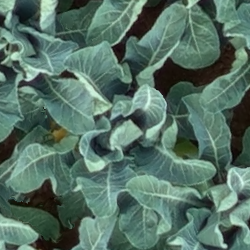}
    \includegraphics[width=0.06\textwidth]{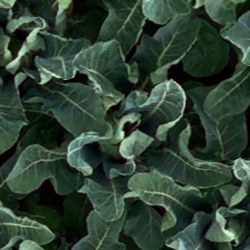}
    \includegraphics[width=0.06\textwidth]{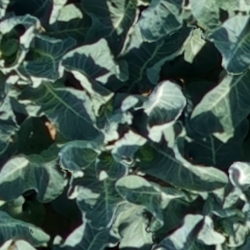}
     \includegraphics[width=0.06\textwidth]{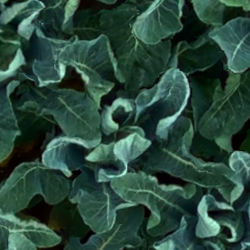}
     \includegraphics[width=0.06\textwidth]{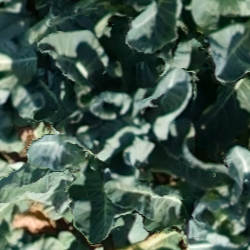}
     \includegraphics[width=0.06\textwidth]{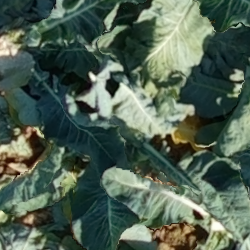}

    \includegraphics[width=0.06\textwidth]{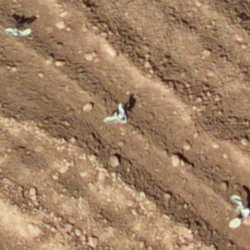}
    \includegraphics[width=0.06\textwidth]{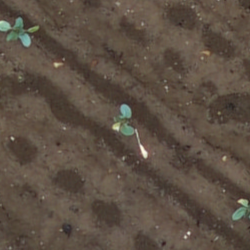}
    \includegraphics[width=0.06\textwidth]{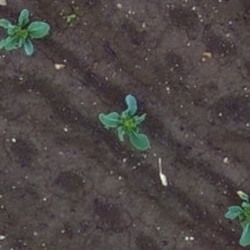}
    \includegraphics[width=0.06\textwidth]{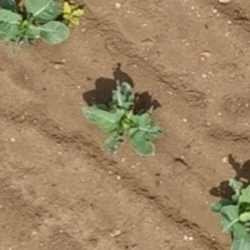}
    \includegraphics[width=0.06\textwidth]{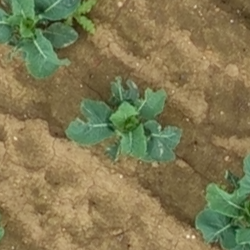}
    \includegraphics[width=0.06\textwidth]{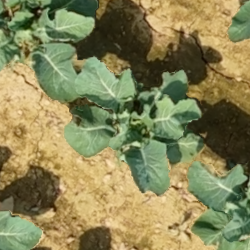}
    \includegraphics[width=0.06\textwidth]{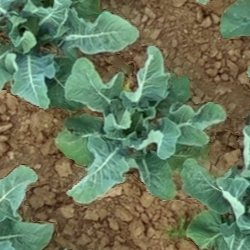}
    \includegraphics[width=0.06\textwidth]{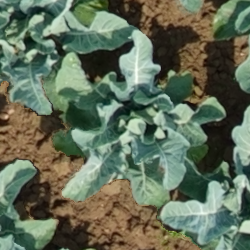}
     \includegraphics[width=0.06\textwidth]{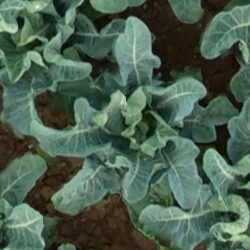}
     \includegraphics[width=0.06\textwidth]{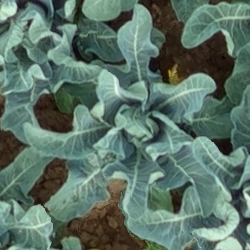}
    \includegraphics[width=0.06\textwidth]{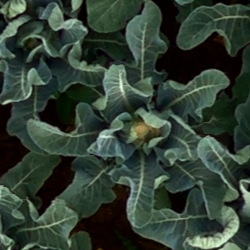}
    \includegraphics[width=0.06\textwidth]{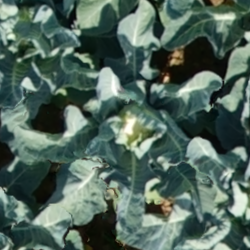}
     \includegraphics[width=0.06\textwidth]{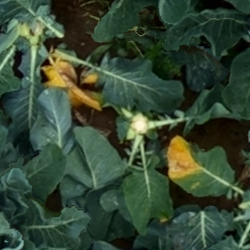}
     \includegraphics[width=0.06\textwidth]{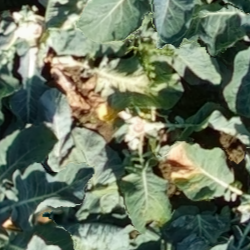}
     \includegraphics[width=0.06\textwidth]{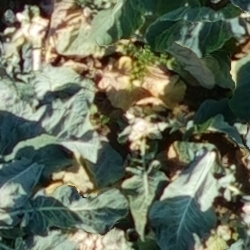}

    \includegraphics[width=0.06\textwidth]{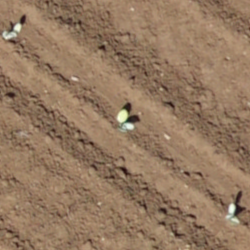}
    \includegraphics[width=0.06\textwidth]{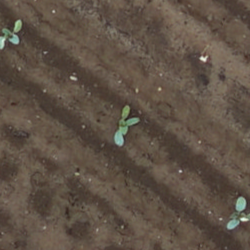}
    \includegraphics[width=0.06\textwidth]{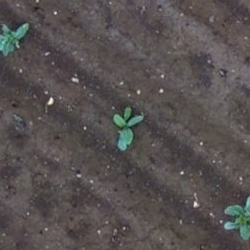}
    \includegraphics[width=0.06\textwidth]{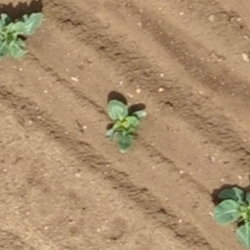}
    \includegraphics[width=0.06\textwidth]{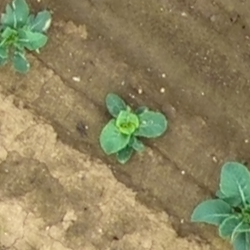}
    \includegraphics[width=0.06\textwidth]{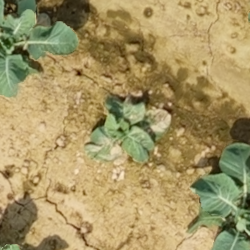}
    \includegraphics[width=0.06\textwidth]{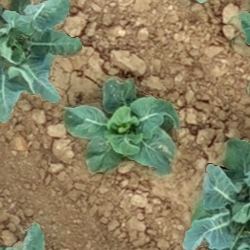}
    \includegraphics[width=0.06\textwidth]{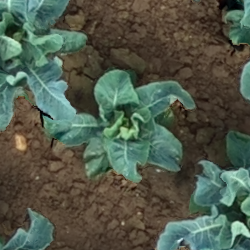}
     \includegraphics[width=0.06\textwidth]{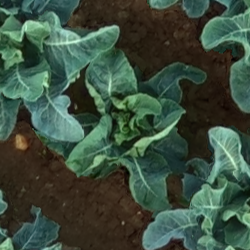}
     \includegraphics[width=0.06\textwidth]{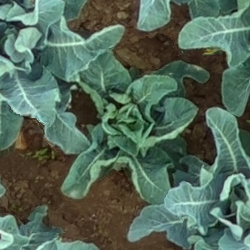}
    \includegraphics[width=0.06\textwidth]{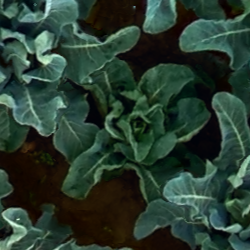}
    \includegraphics[width=0.06\textwidth]{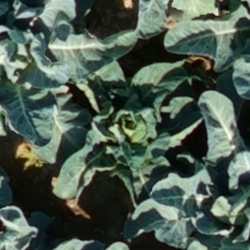}
     \includegraphics[width=0.06\textwidth]{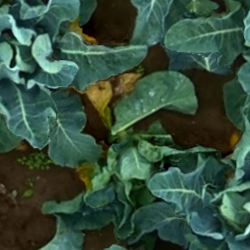}
     \includegraphics[width=0.06\textwidth]{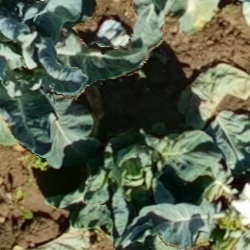}
     \includegraphics[width=0.06\textwidth]{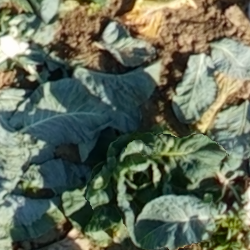}

     \includegraphics[width=0.06\textwidth]{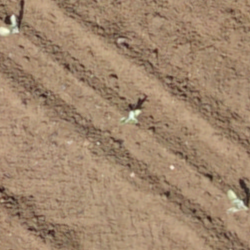}
    \includegraphics[width=0.06\textwidth]{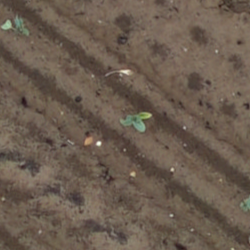}
    \includegraphics[width=0.06\textwidth]{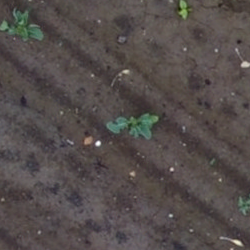}
    \includegraphics[width=0.06\textwidth]{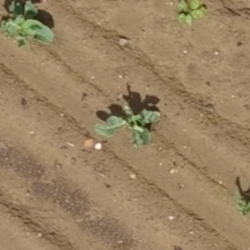}
    \includegraphics[width=0.06\textwidth]{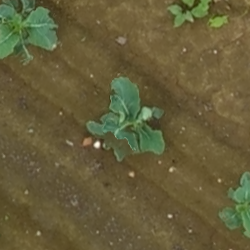}
    \includegraphics[width=0.06\textwidth]{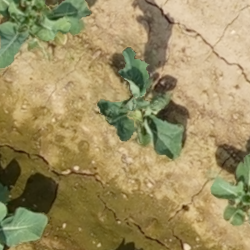}
    \includegraphics[width=0.06\textwidth]{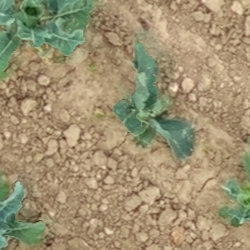}
    \includegraphics[width=0.06\textwidth]{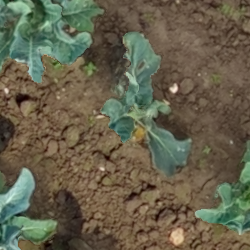}
     \includegraphics[width=0.06\textwidth]{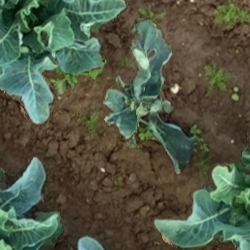}
     \includegraphics[width=0.06\textwidth]{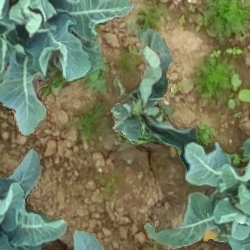}
    \includegraphics[width=0.06\textwidth]{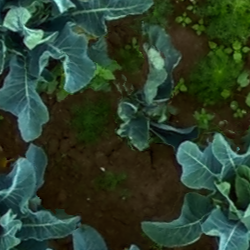}
    \includegraphics[width=0.06\textwidth]{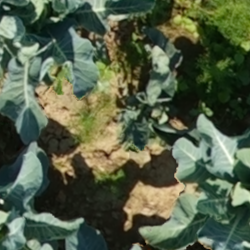}
     \includegraphics[width=0.06\textwidth]{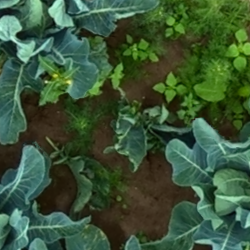}
     \includegraphics[width=0.06\textwidth]{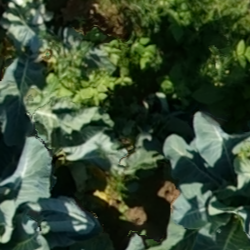}
     \includegraphics[width=0.06\textwidth]{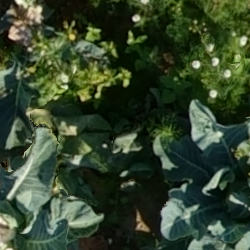}
\caption{Four plant time series within field 2. A row represents a time series. The different columns represent the acquisition dates.} 
	\label{fig:timeSeries_others_field2}
\end{figure}

\begin{figure}[t]
	\centering
	\subfloat[Different amount of weed occurrence for acquisition date August,~\nth{11}.]{
	\begin{minipage}{0.45\textwidth}
	\includegraphics[width=0.24\textwidth]{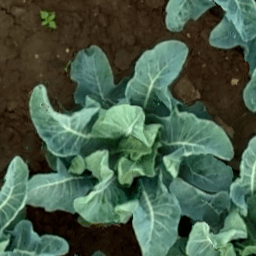}
    \includegraphics[width=0.24\textwidth]{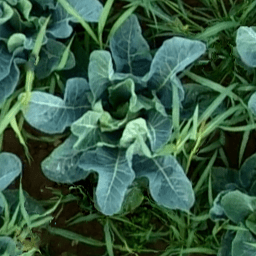}
    \includegraphics[width=0.24\textwidth]{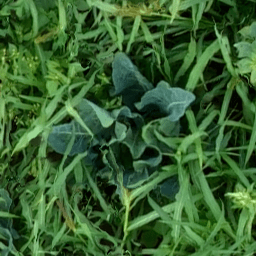}
    \includegraphics[width=0.24\textwidth]{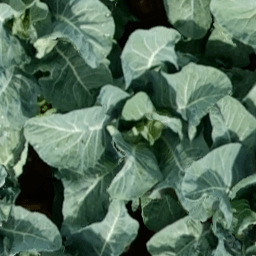}
    \end{minipage}
    \label{fig:weed_in_data}}
    \hspace{20pt}
    \subfloat[Data gap occurrence.]{
	\begin{minipage}{0.45\textwidth}
	\includegraphics[width=0.24\textwidth]{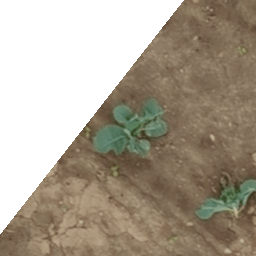}
    \includegraphics[width=0.24\textwidth]{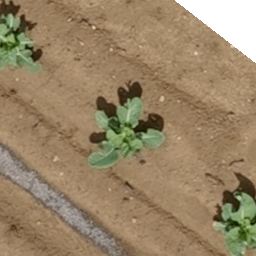}
    \includegraphics[width=0.24\textwidth]{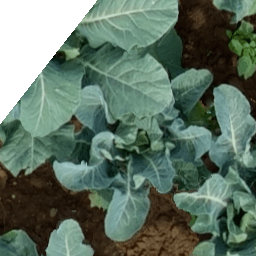}
    \includegraphics[width=0.24\textwidth]{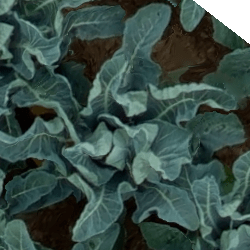}
	\end{minipage}    
	\label{fig:bordering_patches}}
\caption{Data gaps and different amount of weed occurrences within image data at different growing stages.}
	\label{fig:dataGaps_and_weed_in_data}
\end{figure}

For the use of individual plant images, patches have to be cropped by the dataset user around the local plant coordinates determined in the dataset.
The dataset contains, in addition to all global plant coordinates, the local coordinates of the patches for each acquisition date, which at a size of $\SI{490}{\px} \times \SI{490}{\px}$ for field 1 and $\SI{256}{\px} \times \SI{256}{\px}$ for field 2 lie completely within the orthophoto and are not showing spatial data gaps, as patches shown in \figref{fig:bordering_patches}.
Five examples of time series are shown in \figref{fig:timeSeries_others_field1} for field 1 and four in \figref{fig:timeSeries_others_field2} for field 2. 
Due to spatial data gaps, the amount of coordinates per date for field 1 varies, which leads to temporal data gaps within the time series. The largest set of time series that includes equal time steps consists of 3611 time series based on eight time points, including the five time points up to day after planting 42 (Sept, \nth{8}), and all three time points from day after planting 91 (Oct, \nth{27}). In addition to the file that contains all UTM coordinates, a txt-file containing UTM coordinates for this set is also provided, so that time series can be extracted for these selected plant IDs by the dataset user.
After removing patches with spatial data gaps, there are 8402 complete image time series for field 2. Due to the heterogeneous weed occurrence in field 2, the patches contain different amounts of weed, as can be seen in \figref{fig:weed_in_data}.
Due to the given UTM coordinates, there is the possibility to extract the complete time series set of local coordinates for both fields if needed.

\subsubsection{Time series for reference plant data (GrowliFlowerR)}\label{sec:GrowliFlowerR}

For each field, the dataset includes plant IDs and coordinates to enable the dataset user the extraction of an image time series set of monitored reference plants. Those time series look similar to those described in \secref{sec:GrowliFlowerT}. Time series of field 1 comprise the early plant developmental stages as well as the harvest dates, but lack dates, when the canopy was closed. Time series of field 2 comprise all growth stages (see \figref{fig:timeSeries_others_field2}).
\tabref{table:imageRefPlants_field1} gives the distribution of available plant IDs and thus, the number of images of plants per time point for field 1. 
The orthophotos pre defoliation of October, \nth{27} and October, \nth{29} do not overlap with the reference plots due to the low quality of underlying UAV images. Since the reference plants were not defoliated, the orthophotos of the defoliation flights are used to extract images of these days for reference time series. 
For field 2, all local coordinates are given for every acquisition date which enables the user to extract complete image time series. The data is divided into a training, validation, and test set for both fields. Plants of each plot are presented in each set. The visual distribution for both fields is shown in the appendix in \figref{fig:reference_map_trainValTest}.

\begin{table}[t]
    \begin{minipage}{\textwidth}
        \begin{minipage}{\textwidth}
            \begin{flushleft}
                \begin{tabular}{c|c|c|c|c|c|c|c|c|c }
                    \textbf{Date} & Aug 12 & Aug 19  & Aug 25 & Sept 2 & Sept 8 & Sept 17 & Sept 22 & Oct 06 & Oct 19 \\
                    \hline
                    \textbf{\#images} & 239  & 239 & 239 & 239 & 239 & 239 & -- & -- & 193 \\ 
                    
                \end{tabular}
            \end{flushleft}
        \end{minipage}
        \vspace{8pt}
    
        \begin{minipage}{\textwidth}
            \begin{flushleft}
                \begin{tabular}{c|c|c|c}
                    \textbf{Date} & Oct 27 (Post) & Oct 29 (Post) &  Nov 2 \\
                    \hline
                    \textbf{\#images} & 119 & 119 & 12\\
                \end{tabular}
            \end{flushleft}
        \end{minipage}
    \end{minipage}
    \caption{Amount of reference plant image patches per acquisition date for field 1 (2020).}
    \label{table:imageRefPlants_field1}
\end{table}

\subsubsection{Time series for defoliated plant data (GrowliFlowerD)}\label{sec:GrowliFlowerD}

For field 1, the dataset contains in total 130 plant IDs and coordinates of defoliated plants, 30 for October, \nth{27} and 100 for October, \nth{29}. For field 2, it contains in total 722 plant IDs and coordinates of defoliated plants. 
The coordinates enable the dataset user to extract time series of defoliated plants. \tabref{table:imageDefPlants} gives an overview of how many plants were defoliated on which acquisition day. Besides the time series, pairs of pre and post defoliation images are provided.
The data is divided into a training, validation, and test set for both fields. Each day of defoliation is presented in each set. The visual distribution for both fields is shown in the appendix in \figref{fig:defoliation_map_trainValTest}. 

\begin{table}[t]
    \centering
    \begin{tabular}{c|c|c|c|c|c|c}
        \textbf{Date} & Aug 19 & Aug 23  & Aug 25 & Aug 30 & Sept 3 & Sept 8 \\
        \hline
        \textbf{\#images} & 109  & 115 & 193 & 116 & 70 & 54 \\
    \end{tabular}
    \caption{Amount of defoliated plants per acquisition date for field 2 (2021).}
    \label{table:imageDefPlants}
\end{table}

\subsection{In-situ data}

\begin{figure}[t]
	\centering
    \subfloat[Field 1]{
    \includegraphics[width=\textwidth]{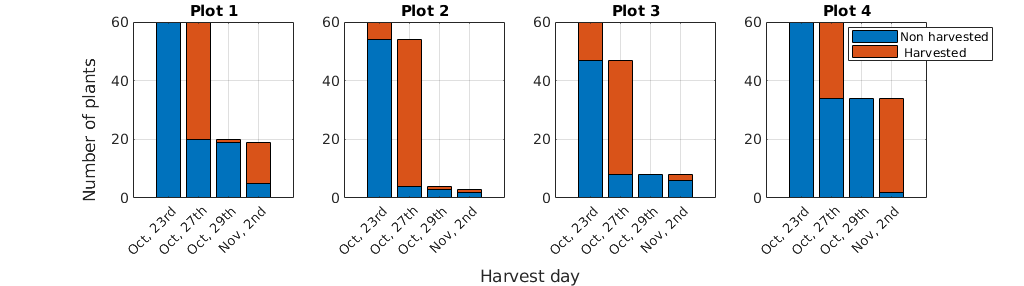}
    \label{fig:harvestDays_field1}}
    
    \subfloat[Field 2]{
    \includegraphics[width=\textwidth]{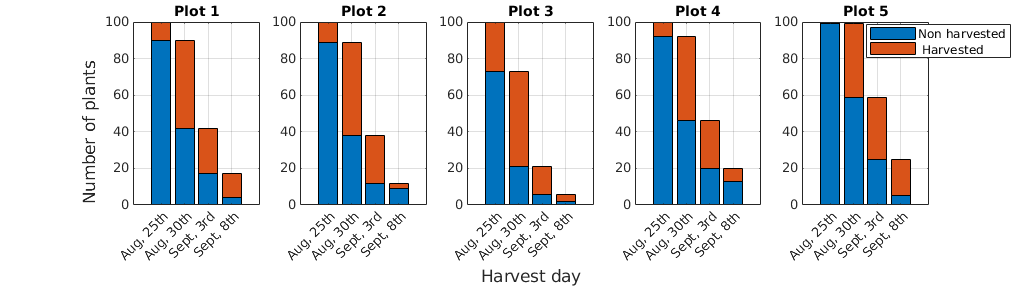}
    \label{fig:harvestDays_field2}}
    \caption{Overview about harvested and non-harvested plants per reference plot per day.}
	\label{fig:harvestDays}
\end{figure}

Two csv-files are provided, one for each field, which contain the plant ID in addition to the described measurements of \secref{sec:manualMeasurements} for each data acquisition day. The measured values correlate with the images of GrowliFlowerR.
\figref{fig:harvestDays} shows the distribution of the number of harvested plants within the reference plots per acquisition date for both fields.

\section{Baseline for instance segmentation application} \label{sec:results}

\subsection{Experimental Setup}
We show two possible applications of the presented data by creating baselines using the labeled dataset \text{GrowliFlowerL} and Mask R-CNN \citep{he2017mask}, a state-of-the-art method for instance segmentation. We address the tasks of segmenting plant instances and segmenting leaf instances, hence, we use the mask and bounding boxes derived from the mask of plant instances as target for one baseline and the mask and bounding box derived from the mask including leaf instances as target for a second baseline. For leaf instance segmentation baseline the given void instances are used as background since only leaves that do not belong to the void plants have been labeled.
The estimation of the semantic masks for the individual instances enables a derivation of phenotypic traits. As data augmentation, we use random horizontal flipping with a probability of 0.5.

We train on a single GPU machine with Intel Core i7-6850K 3.60\,GHz processor and a Geforce GTX 1080Ti with 11\,GB RAM. The network is pretrained on COCO dataset \citep{lin2014microsoft}. The training is done over 100 epochs with a learning rate of 0.001 and batchsize of 2. We use an SGD optimizer and for the backbone network a ResNet-50.

\subsection{Evaluation metrics}

The Intersection over Union (\text{IoU}) was calculated as: 

\begin{equation}
    \text{IoU} = \frac{\text{TP}}{\text{TP} + \text{FP} + \text{FN}}
\end{equation}

following the evaluation metrics of COCO dataset \citep{lin2014microsoft},
where \text{TP} are true positives, \text{FP} false positives and \text{FN} false negatives. Two additional scores used are precision $p$ and recall $r$. They are defined as:

\begin{align}
    p = \frac{\text{TP}}{\text{TP} + \text{FP}} && r = \frac{\text{TP}}{\text{TP} + \text{FN}}
\end{align}

The F1 score summarizes the precision and recall scores 
and is defined as
\begin{equation}
    F1 = \frac{2 \cdot p \cdot r}{p + r}
\end{equation}

We compute precision, recall and F1 score in respect to the single object class cauliflower plant and calculate the scores for IoU thresholds $t_{\text{IoU}} = 0.50$ and $t_{\text{IoU}} = 0.75$. In addition, we determine the average precision (AP), average recall (AR) and average F1 (AF1) score over all IoU in the interval $0.50-0.95$ with step size $0.05$ as for the COCO benchmark. This is indicated by $(\cdot)@0.5-0.95$.
For the leaf instance segmentation baseline, we reduce the evaluation on recall, as we do not want to penalize predictions on void pixels.

\subsection{Results}

We perform the metric calculations with respect to the detected bounding boxes and the segmented masks of the respective objects. In the case of segmented masks, we consider the accumulated number of correctly classified pixels and thus, the more precise shape of the object. With the bounding box, it is more generally about the accuracy of the detection and thus, the localization of the object.

\begin{figure}[t]
	\centering
	 \subfloat[Precision]{
    \includegraphics[width=0.32\textwidth]{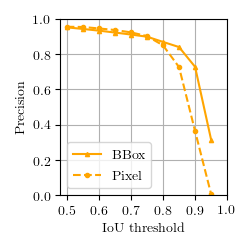}
    \label{fig:precision_plant}}
    \subfloat[Recall]{
    \includegraphics[width=0.32\textwidth]{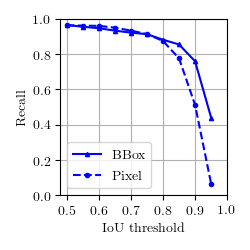}
    \label{fig:recall_plant}}
     \subfloat[F1]{
    \includegraphics[width=0.32\textwidth]{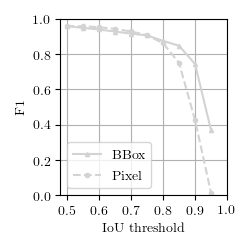}
     \label{fig:f1_plant}}
\caption{Representation of precision, recall and F1 score for class cauliflower \texttt{plant}. 
The graphs show the evaluation at different IoU thresholds on bounding box (BBox), thus object, (solid-line) and pixel (dashed-line) level.}
	\label{fig:results_scores_plant}
\end{figure}

\tabref{table:evalMetric_plant} summarizes the results for plant instance segmentation for the baseline method. The table shows that 95\% at IoU $\geq 0.5$ are predicted correctly. The precision on bounding box level and pixel level is above 80\% for all IoU thresholds $\leq 0.8$ (see \figref{fig:precision_plant}). At an IoU $\geq 0.85$ it decreases quickly. This also applies for recall (\figref{fig:recall_plant}) and F1 score (\figref{fig:f1_plant}).
For higher IoU values the prediction on pixel level is less accurate as on bounding box level since slight changes in the segmentation generally lead to higher errors in the segmentation mask than in the bounding box.
An overview of our metric results is given in \tabref{table:evalMetric_plant}.

\begin{figure}[t]
	\centering
	 \subfloat[Accurate plant instance segmentation results.]{
    \begin{minipage}{0.45\textwidth}
        \includegraphics[width=0.32\textwidth]{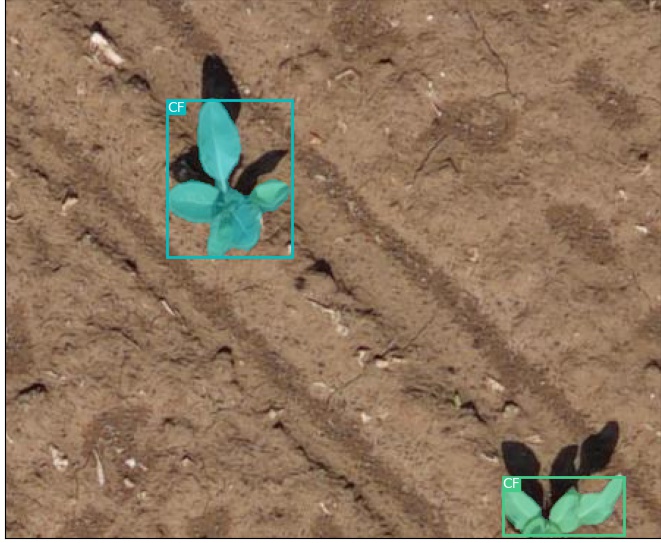}
        \includegraphics[width=0.32\textwidth]{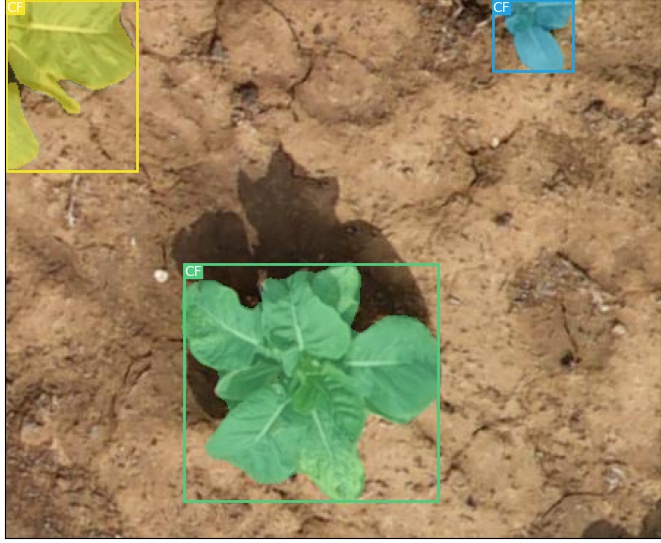}
        \includegraphics[width=0.32\textwidth]{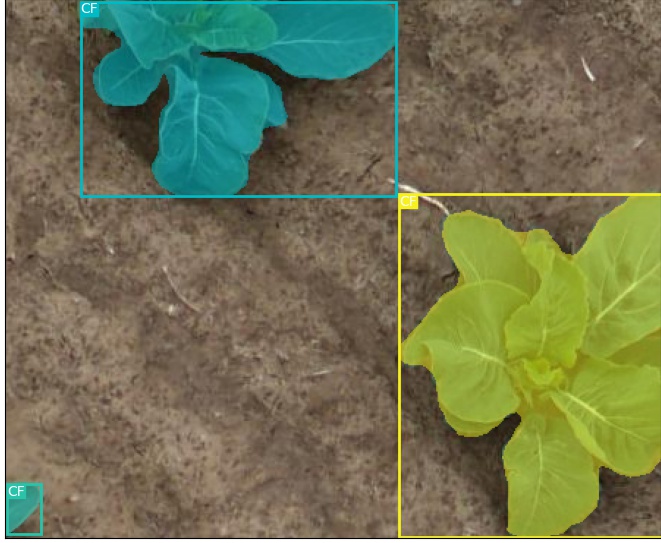}
        
        \includegraphics[width=0.32\textwidth]{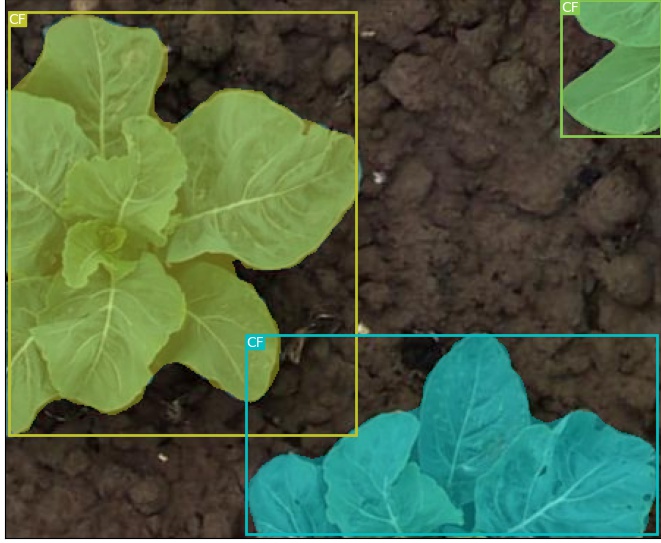}
        \includegraphics[width=0.32\textwidth]{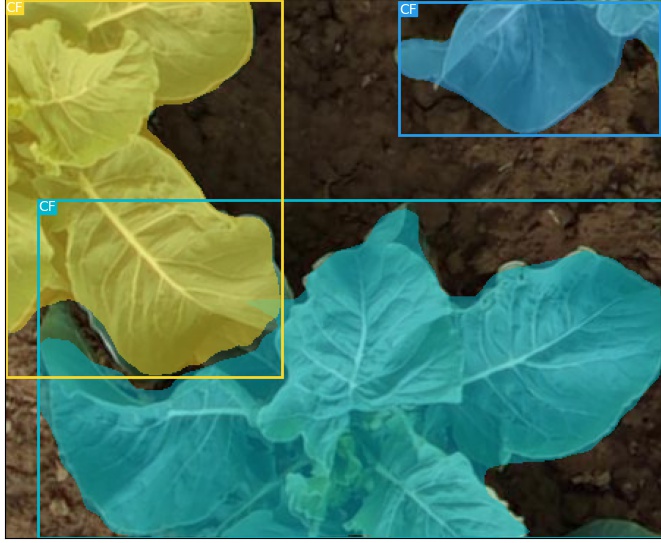}
        \includegraphics[width=0.32\textwidth]{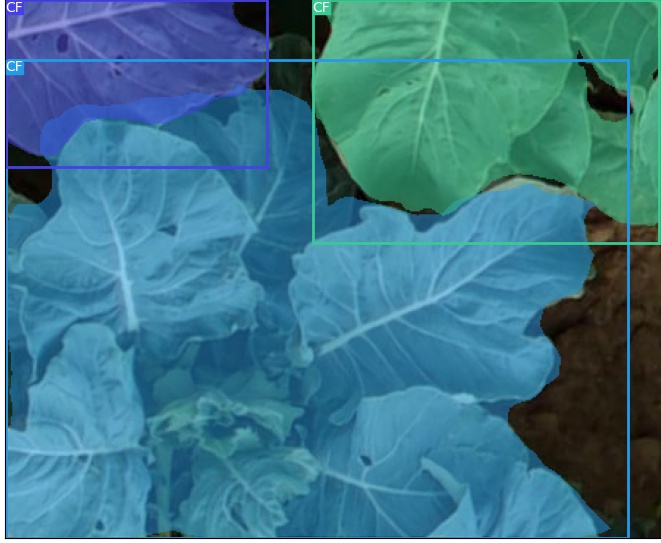}
    \end{minipage} \label{fig:plant_instances_results_acc}}
    \hspace{15pt}
     \subfloat[Improvable plant instance segmentation results.]{
    \begin{minipage}{0.45\textwidth}
        \centering
        \includegraphics[width=0.32\textwidth]{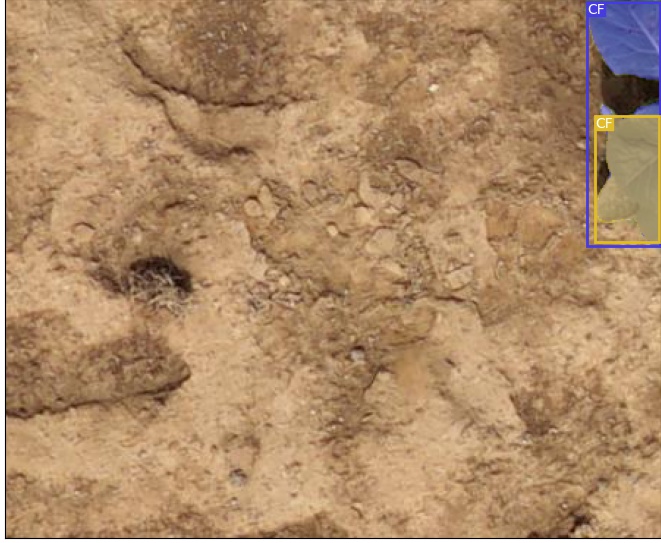}
        \includegraphics[width=0.32\textwidth]{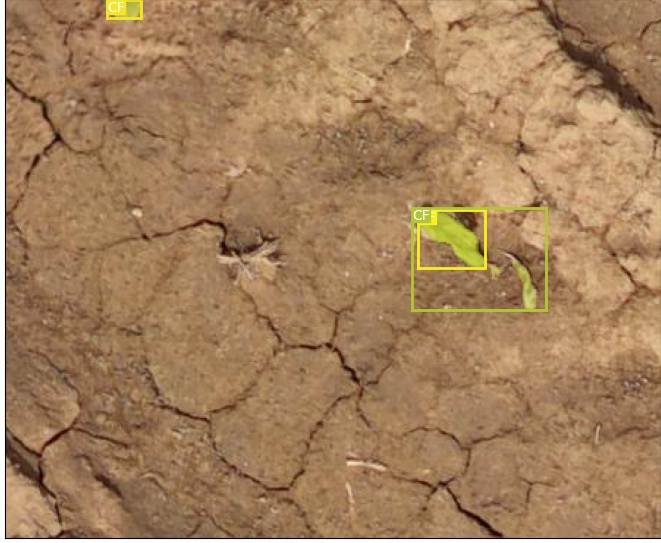}
        
        \includegraphics[width=0.32\textwidth]{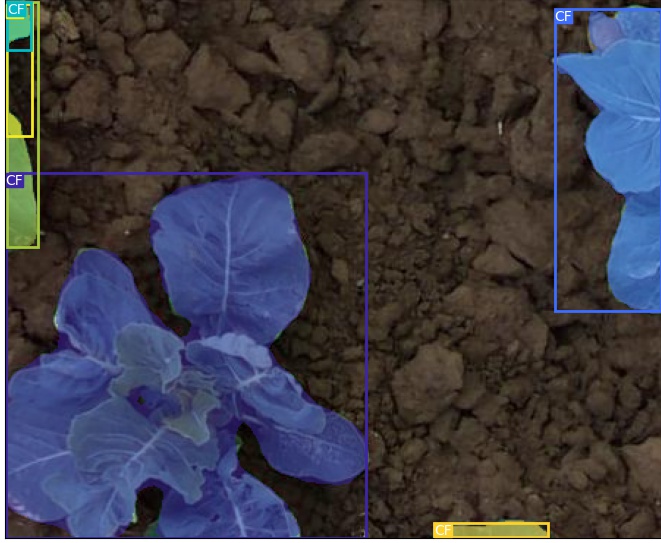}
        \includegraphics[width=0.32\textwidth]{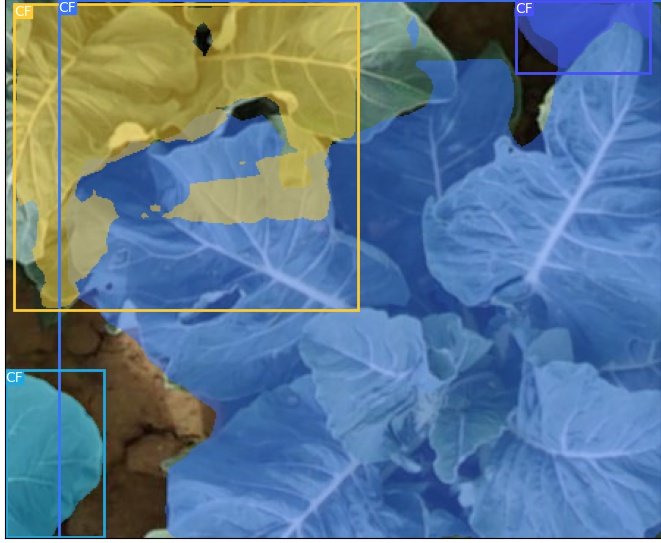} 
    \end{minipage} \label{fig:plant_instances_results_exp}}
    \vspace{15pt}
    \subfloat[Accurate leaf instance segmentation results.]{
    \begin{minipage}{0.45\textwidth}
        \includegraphics[width=0.32\textwidth]{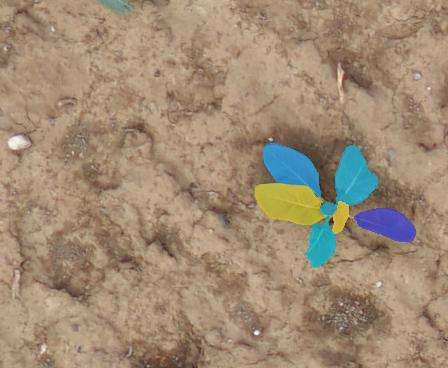}
        \includegraphics[width=0.32\textwidth]{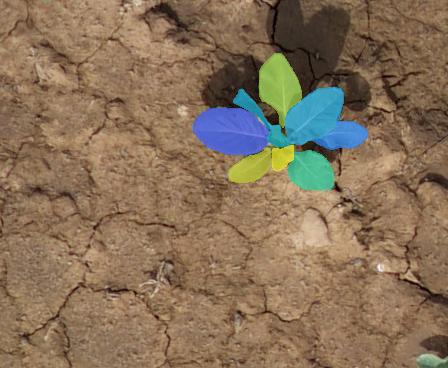}
        \includegraphics[width=0.32\textwidth]{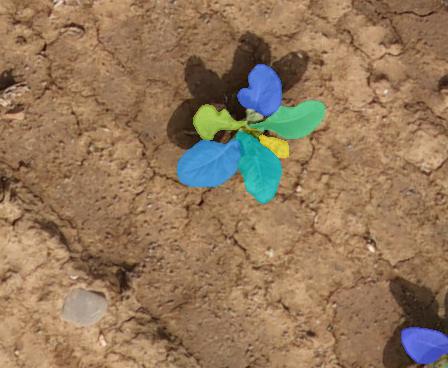}
        
        \includegraphics[width=0.32\textwidth]{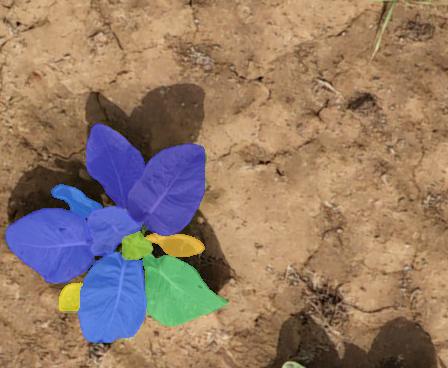}
        \includegraphics[width=0.32\textwidth]{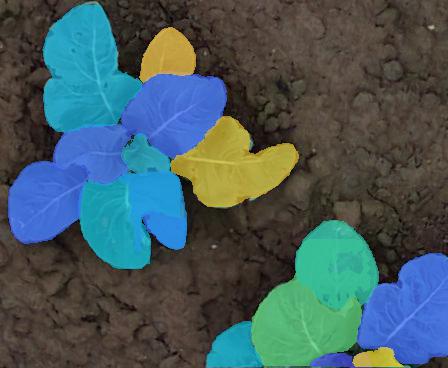}
        \includegraphics[width=0.32\textwidth]{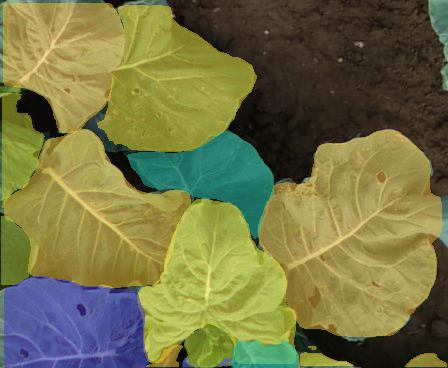}
    \end{minipage}\label{fig:leaf_instances_results_acc}}
    \hspace{15pt}
     \subfloat[Improvable leaf instance segmentation results.]{
    \begin{minipage}{0.45\textwidth}
        \centering
        \includegraphics[width=0.32\textwidth]{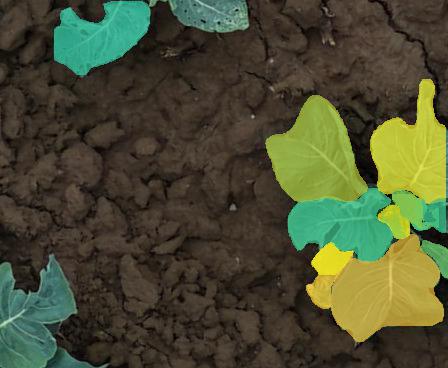}
        \includegraphics[width=0.32\textwidth]{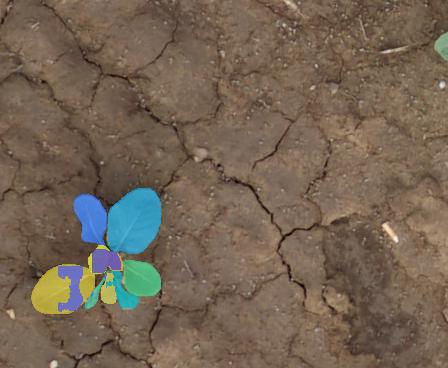}

        \includegraphics[width=0.32\textwidth]{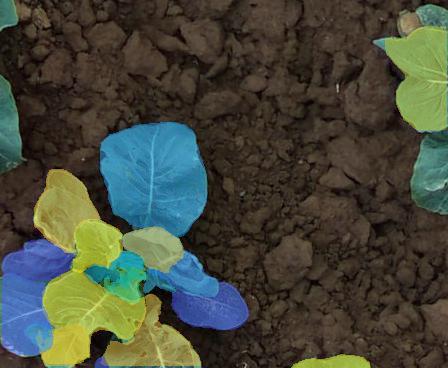}
        \includegraphics[width=0.32\textwidth]{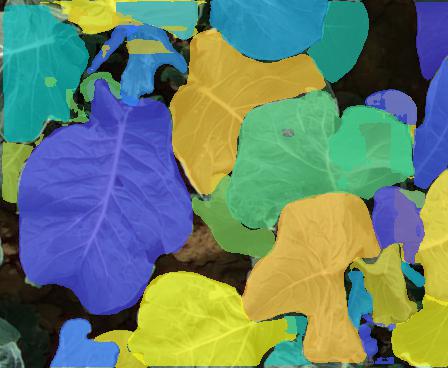}
    \end{minipage}\label{fig:leaf_instances_results_exp}}
    
\caption{Plant and leaf instance segmentation results. The different colors indicate the different instances. In the visualization of the leaf instance segmentation results, we concentrate on the visualization of the masks for the sake of clarity and omit the bounding boxes. Examples in (a) and (c) show accurate results and examples in (b) and (d) show improvable results.}
\label{fig:results_instanceSegmentation}
\end{figure}

Looking at the visual results, it can be seen that many of the objects and masks are estimated accurately (\figref{fig:plant_instances_results_acc}). The results show all prediction with a score higher than a threshold of 50\%. A precise contour is estimated and in the earlier stages of development the instances are well spatially separated. The ground is not considered to be an object in any case, nor are the smaller weeds that can be seen on some patches.
The inaccuracies occur especially with plants that lie at the edge of the image patches. Only small parts of the plant are visible and therefore its leaves are not adjacent to each other, see \figref{fig:plant_instances_results_exp} top left and bottom left. Some errors occur in later developmental stages since the plants overlap (\figref{fig:plant_instances_results_exp} bottom right) which states a more challenging scenario than well-separated plants. In particular for overlapping plants, it is difficult even with the human eye to assign the leaves to the individual instances. Furthermore, there are not as many training images available for the later stage of maturity compared to the earlier stages of development where no overlapping occurs. 
Another distinctive feature involves plant objects that loose leaves or are impaired in their growth and thus decay. Thus, it is difficult for the model to distinguish whether one or more plants are represented (\figref{fig:plant_instances_results_exp} top right).

\begin{table}[t]
        \centering
        \begin{tabular}{c??c|c|c??c|c??c|c??c|c}
        & \multicolumn{3}{c??}{\textbf{Global metrics}} &\multicolumn{2}{c??}{\textbf{Precision}} & \multicolumn{2}{c??}{\textbf{Recall}} & \multicolumn{2}{c}{\textbf{F1}}\\
        \textbf{Evaluation} & \textbf{AP} & \textbf{AR} & \textbf{AF1} & \textbf{P@0.5} & \textbf{P@0.75} & \textbf{R@0.5} & \textbf{R@0.75} & \textbf{F1@0.5} & \textbf{F1@0.75} \\
        \specialrule{.1em}{.05em}{.05em} 
        BBox & 0.917 & 0.933 & 0.843 & 0.952 & 0.899 & 0.965 & 0.913 & 0.958 & 0.906\\
        Pixel & 0.844 & 0.858 & 0.770 & 0.954 & 0.902 & 0.963 & 0.913 & 0.959 & 0.908\\
    \end{tabular}
     \caption{Plant instance segmentation results. Precision, recall and F1 score for predicted bounding boxes (BBox) and segmentation masks (pixel) for class \texttt{plant}.}
    \label{table:evalMetric_plant}
\end{table}

\begin{figure}[t]
\begin{minipage}{\textwidth}
  \begin{minipage}{0.32\textwidth} 
    \centering
   \subfloat[Recall]{
    \includegraphics[width=\textwidth]{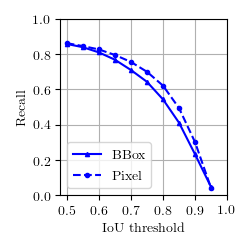}

    \label{fig:recall_leaves}}
  \end{minipage}
    \hspace{15pt}
  \begin{minipage}{0.58\textwidth} 
      \begin{minipage}{\textwidth} 
        \centering
        \end{minipage}
        
        \begin{minipage}{\textwidth} 
        \centering
       \begin{subtable}[h]{1\textwidth}
        \centering
         \begin{tabular}{c??c??c|c}
        &\textbf{Global metrics} & \multicolumn{2}{c}{\textbf{Recall}}\\
        \textbf{Evaluation} & \textbf{AR} & \textbf{R@0.5} & \textbf{R@0.75}  \\
        \specialrule{.1em}{.05em}{.05em} 
        BBox  & 0.735 & 0.857 & 0.642 \\
        Pixel & 0.772 & 0.862 & 0.700 \\
    \end{tabular}
        \caption{}
        \label{table:evalMetric_leaves}
    \end{subtable}
        \end{minipage}
    \end{minipage}
  \end{minipage}
  \caption{Recall results for leaf instance segmentation. The graph (a) shows the evaluation at different IoU thresholds on bounding box (BBox), thus object (solid-line), and pixel (dashed-line) level. Table (b) shows  the respective values for average recall (AR), $R@0.5$ and $R@0.75$.}
	\label{fig:metrics_leaves}
\end{figure}

In the baseline for leaf instance segmentation, the recall results are slightly worse than for plant instance segmentation (see \tabref{table:evalMetric_leaves}), suggesting that the task is more difficult than that of plant instance segmentation.
The distinction between individual leaf instances is more complex than the distinction between plant instances. In addition, we do not assign the void labeled objects to class \texttt{leaf} but class \texttt{BG} for this baseline, since individual void plants can also contain several leaves, which, however, were not individually labeled. The calculated values for recall are similar for pixel level and bounding box level. 

We find explanations in the visual consideration of the results. Even though these results show predictions with a score higher than a threshold of 50\%. By our definition of void instances as background, the model is challenged in non-predicting leaves belonging to void instances, as shown in \figref{fig:leaf_instances_results_exp} top left and bottom left.
The model has difficulty distinguishing whether plants at the edge of the patches are void instances or leaf instances. Therefore, either leaves are predicted that are present in the target (low precision) or no leaves are predicted although they are present in the target (low recall). 
For plants that are completely visible in the patch, the model has more strength in prediction.
Another source of error is the prediction of several instances on one leaf as shown in \figref{fig:leaf_instances_results_exp} top right and bottom right because the model needs to learn features such as leaf structure and size, as these play a crucial role in distinguishing different leaves.

We observe that our instance segmentations, plant instance as well as leaf instance, perform and can be used for different growth stages of the cauliflower plants.

\section{Conclusion and future directions}
\label{sec:conclusion}

This article introduces the GrowliFlower dataset - a georeferenced, image-based UAV time series dataset of two monitored cauliflower fields during their entire growth period.
The paper includes the description of the dataset and provides insights into the data collection process that can be helpful for other data collection activities. 
The dataset consists of weekly RGB and multispectral UAV orthophotos and image time series of individual plants reflecting weekly plant growth. For a subset of the time series, in-situ reference measurements such as plant size are available. For another subset, pre and post images of defoliation are available to provide a relation between the interior and exterior of a cauliflower plant.  
The dataset also contains annotations with segmented plant and leaf instances as well as annotations on stems. The data is available at http://rs.ipb.uni-bonn.de/data/.
The dataset is intended to advance and evaluate machine learning methods and to foster close collaboration between different disciplines such as agricultural sciences, remote sensing, and machine learning.
We present baseline results from two applications that were approached using a Mask R-CNN. One application for plant instance segmentation and one for leaf instance segmentation. 
Furthermore, the findings and descriptions should help to ensure that the conduction of the data collection can be used and transferred to other areas.


\subsubsection*{Acknowledgments} 
This project was funded by the European Agriculture Fund for Rural Development with contribution from North-Rhine Westphalia (17-02.12.01 - 10/16 – EP-0004617925-19-001).
Furthermore, this work was partially funded by the Deutsche Forschungsgemeinschaft (DFG, German Research Foundation) under Germany’s Excellence Strategy – EXC 2070 – 390732324.
The authors like to acknowledge the farmer Markus Schwarz for providing the experimental cauliflower fields and Jonas Westheider for assistance in annotating the instance segmentation dataset.
If data from the GrowliFlower dataset is used or aspects of the data collection are adopted, this paper has to be cited.


\bibliographystyle{apalike}
\bibliography{main}

\newpage
\begin{appendices}

\begin{figure}[h]
	\centering
	\subfloat[Field 1]{
    \includegraphics[trim=750 170 650 170, clip, width=0.45\textwidth]{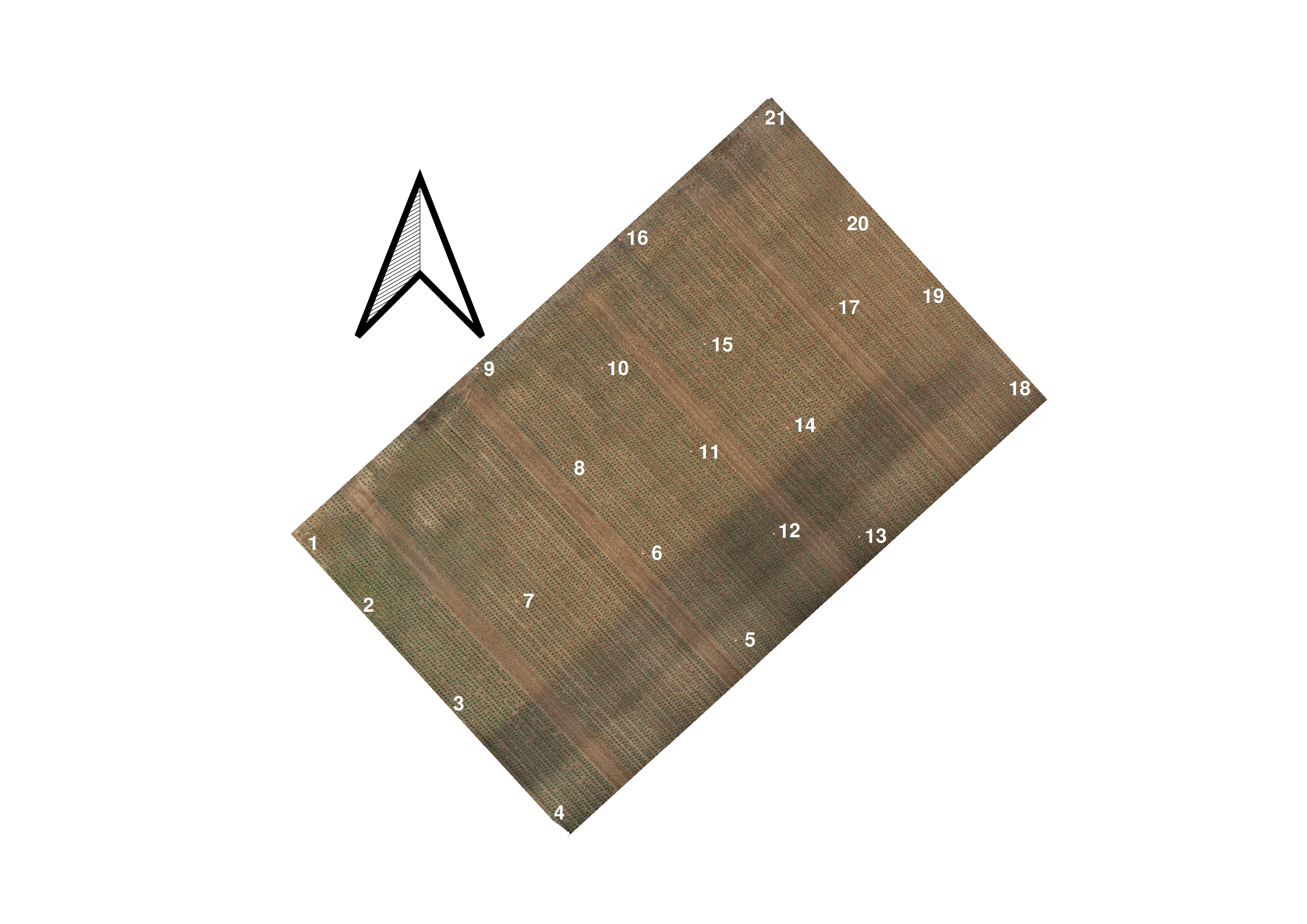}}
    \subfloat[Field 2]{
    \includegraphics[trim=0 0 0 0, clip, width=0.45\textwidth]{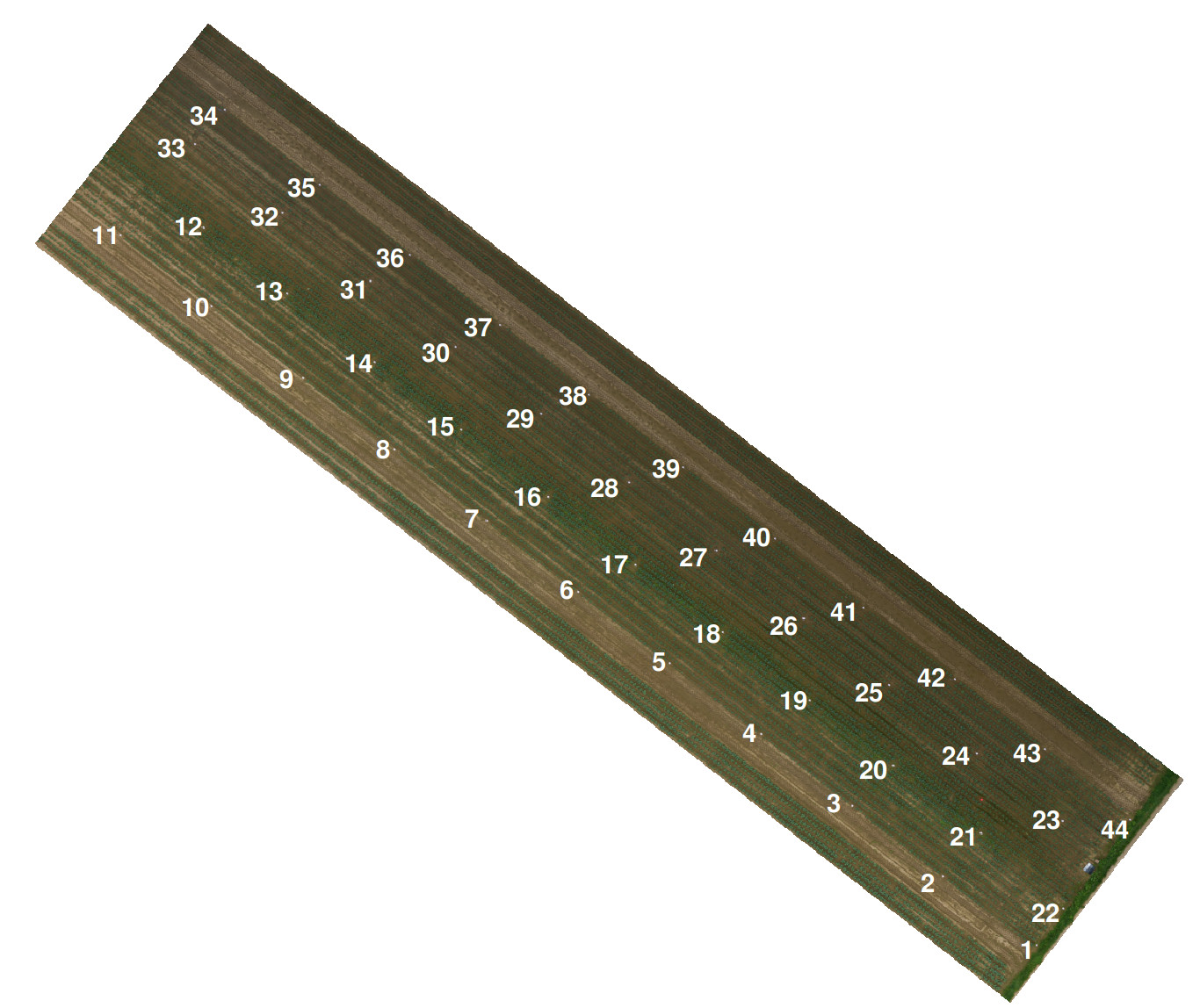}
    }
	\caption{Location of GCPs for both fields.}
	\label{fig:gcp_location}
\end{figure}

\begin{figure}[h]
	\centering
	 \subfloat[]{
    \includegraphics[trim = 100 17 100 0, clip, width=0.5\textwidth]{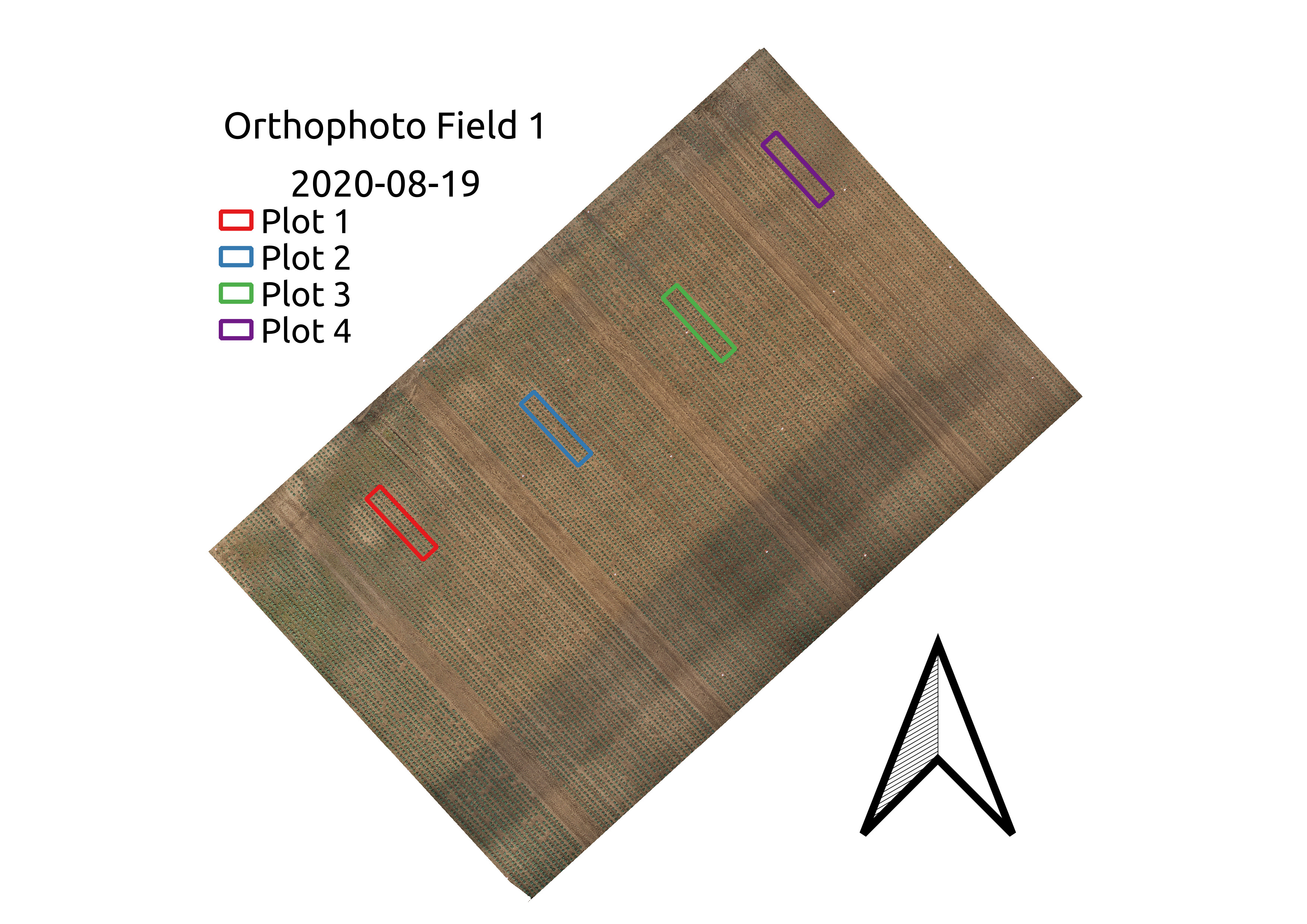}
    \label{fig:refPlots_field1_overview}}
    \hspace{10pt}
     \subfloat[]{
     \includegraphics[width=0.4\textwidth]{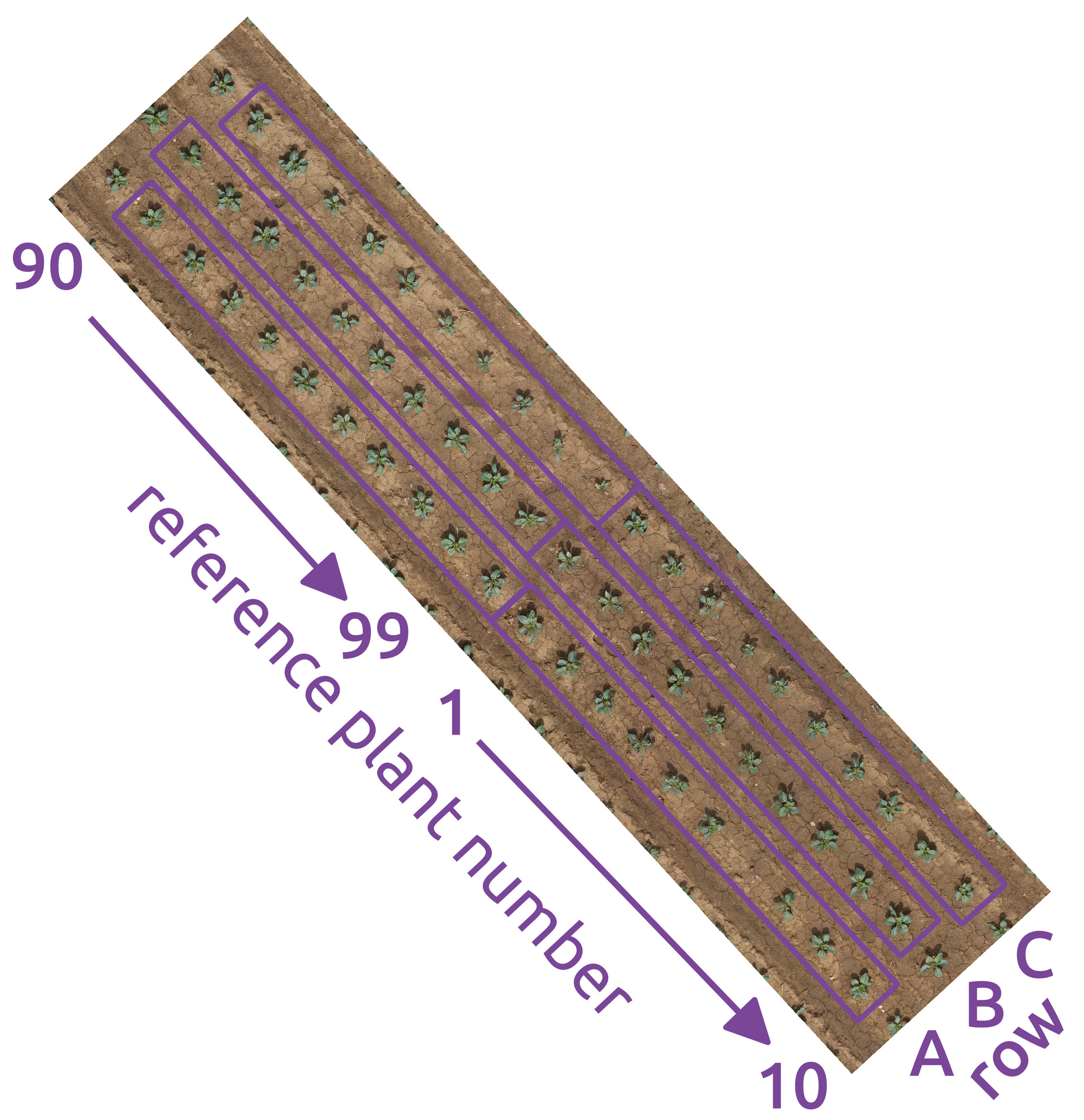}
     \label{fig:refPlot_field1}}
 	\caption{Visual overview of (a) reference plots for in-situ measurements within field 1
and (b) the respective design of reference plot 4 including reference plants and ordering of reference plant numbers. The plot design is valid for all reference plots of field 1.}
	\label{fig:refPlots_field1}
\end{figure}

\begin{table}[h]
\caption{Monitored abiotic and biotic stresses.}\label{table:stresses}
\begin{center}
\begin{tabular}{|c|c||c|c|}
    \hline
    Abbreviation & Meaning & Abbreviation & Meaning \\
    \hline
        P	& plant &  L	& leaf/leaves \\
        nP	& no plant &   wL	& without leaves \\
        Pl	& plant lying down &    oL	& old leaves \\
        wP	& whole plant &   yL	&  yellowish leaves \\
        2P	& 2 plants & 	rL	& reddish leaves \\
        bb	& blind bud &  pgL	& pale green leaves  \\
        pd	& planted too deep &  pygL & pale yellowish green leaves \\
        A & aphids present & sg	& stunted growth with many shoots\\
        C	& coal fleas present & dT	& damage to leaves caused by tractor\\
        F & flies present & & \\
    \hline
\end{tabular}
\end{center}
\end{table}

\begin{figure}[h]
	\centering
    \subfloat[Field 1]{
    \includegraphics[trim= 11 3 11 3, clip, width=0.48\textwidth]{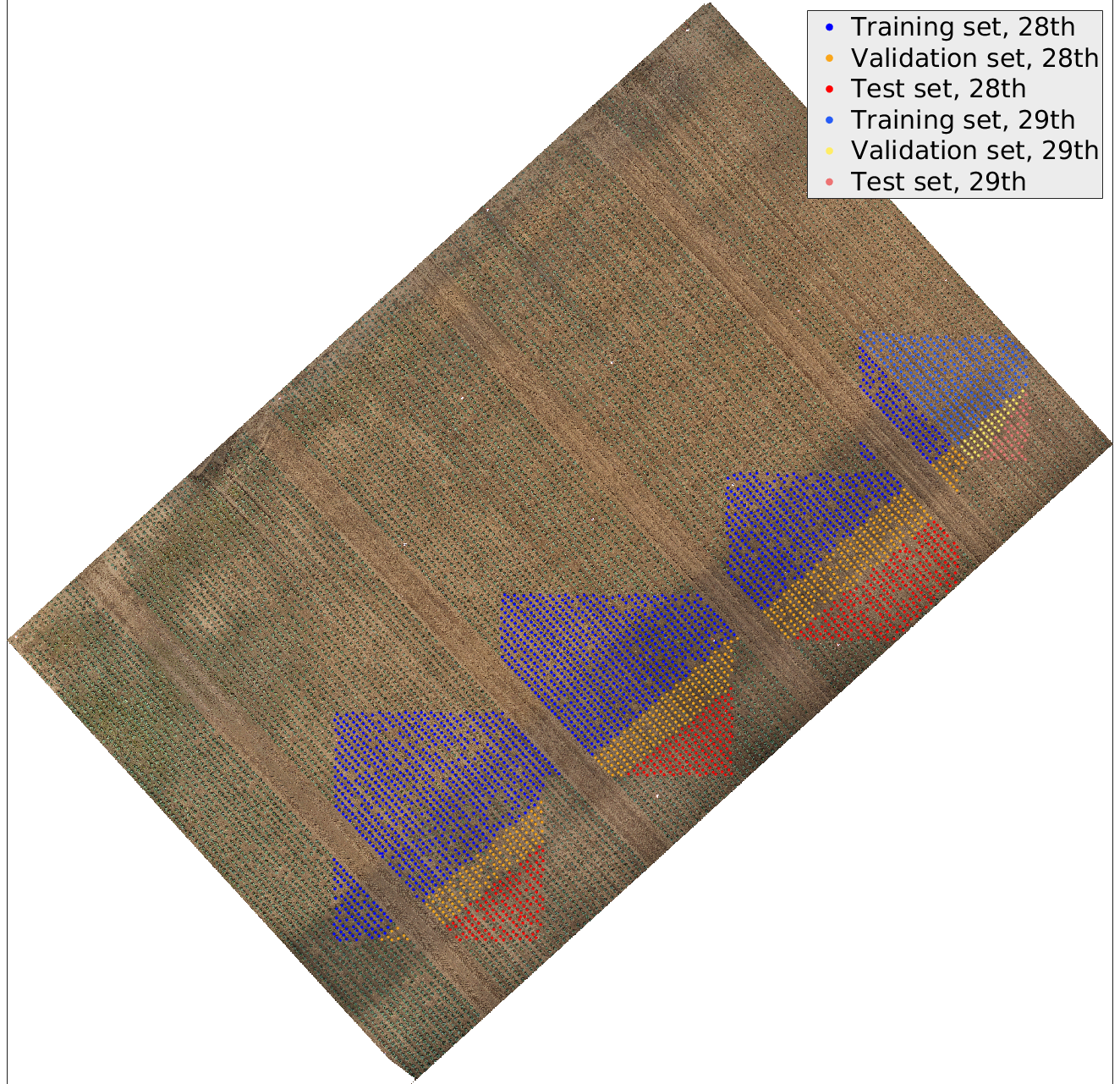}
    \label{fig:random_TrainValTest_field1}}
    \subfloat[Field 2]{
    \includegraphics[width=0.48\textwidth]{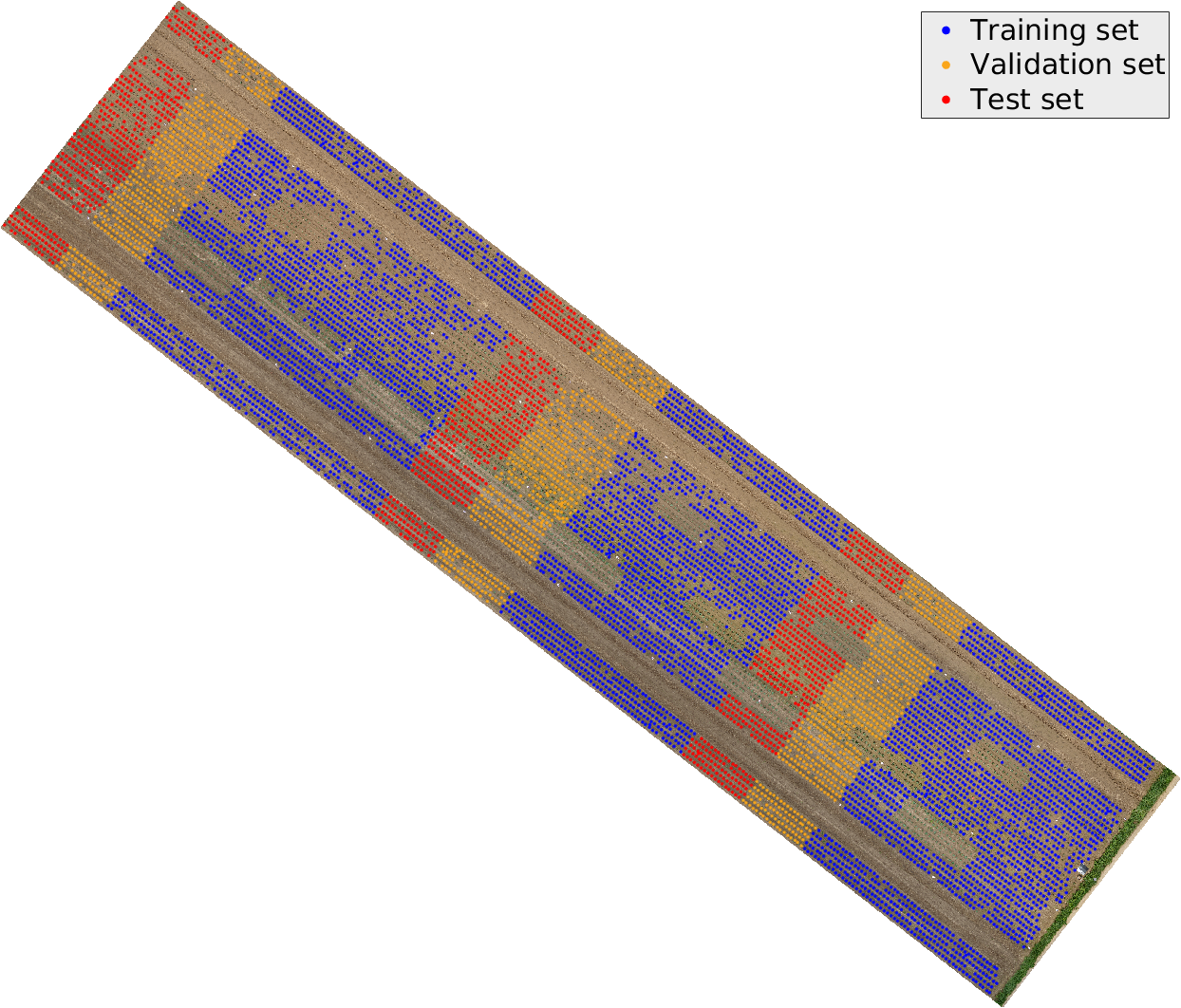}
    \label{fig:random_TrainValTest_field2}}
    \caption{Separation of plants of both fields within GrowliFlowerT in training (blue), validation (yellow) and test set (red).
    For field 1, the two planting days are separated using dark colors for July \nth{28}, 2020 and light colors for July \nth{29}, 2020.
    }
	\label{fig:random_TrainValTest}
\end{figure}

\begin{figure}
    \centering
        \subfloat[Field 1]{
        \includegraphics[trim = 0 3 0 3, clip,width=0.48\textwidth]{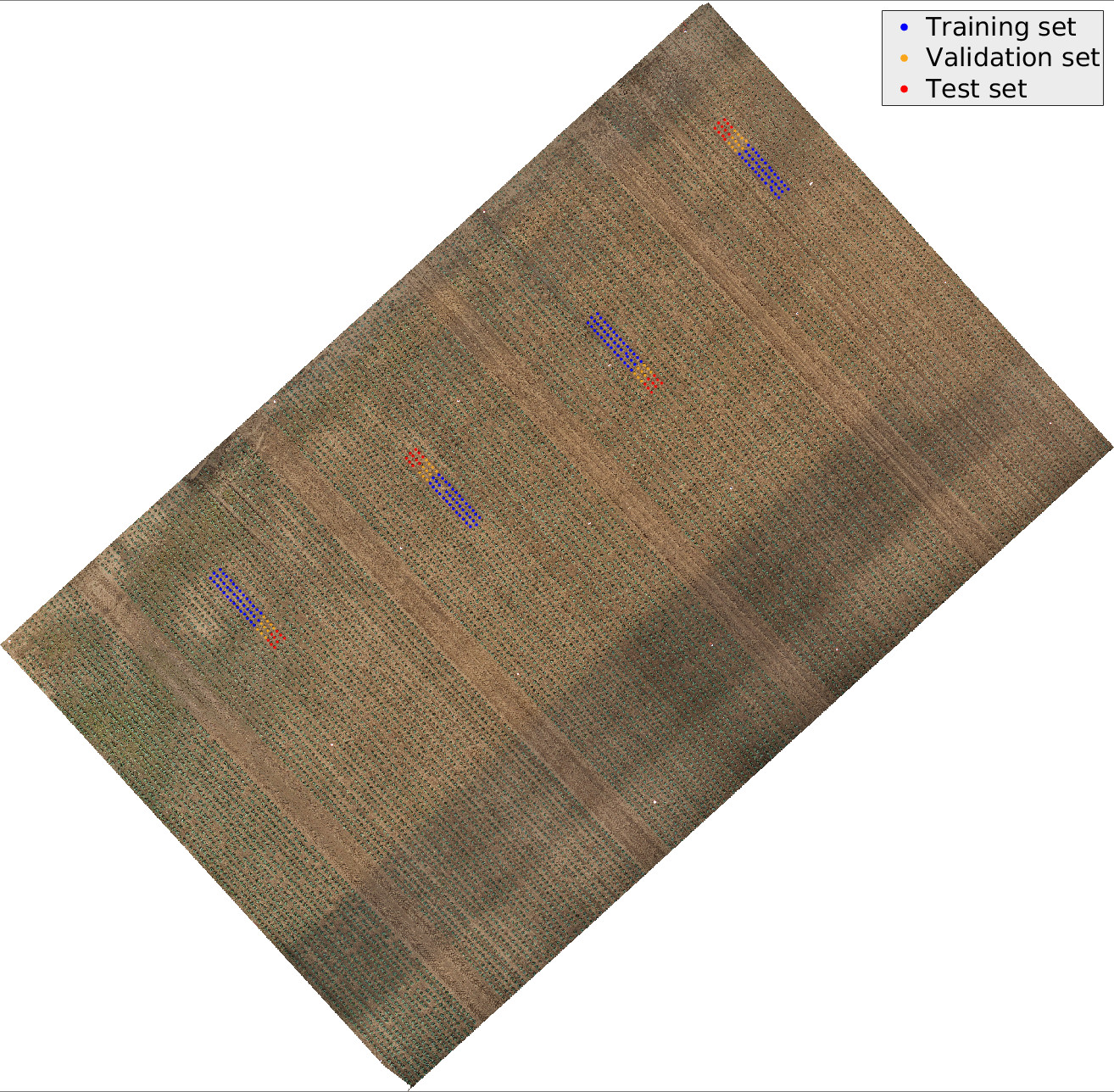}
        \label{fig:reference_map_trainValTest1}}
        \subfloat[Field 2]{
        \includegraphics[width=0.48\textwidth]{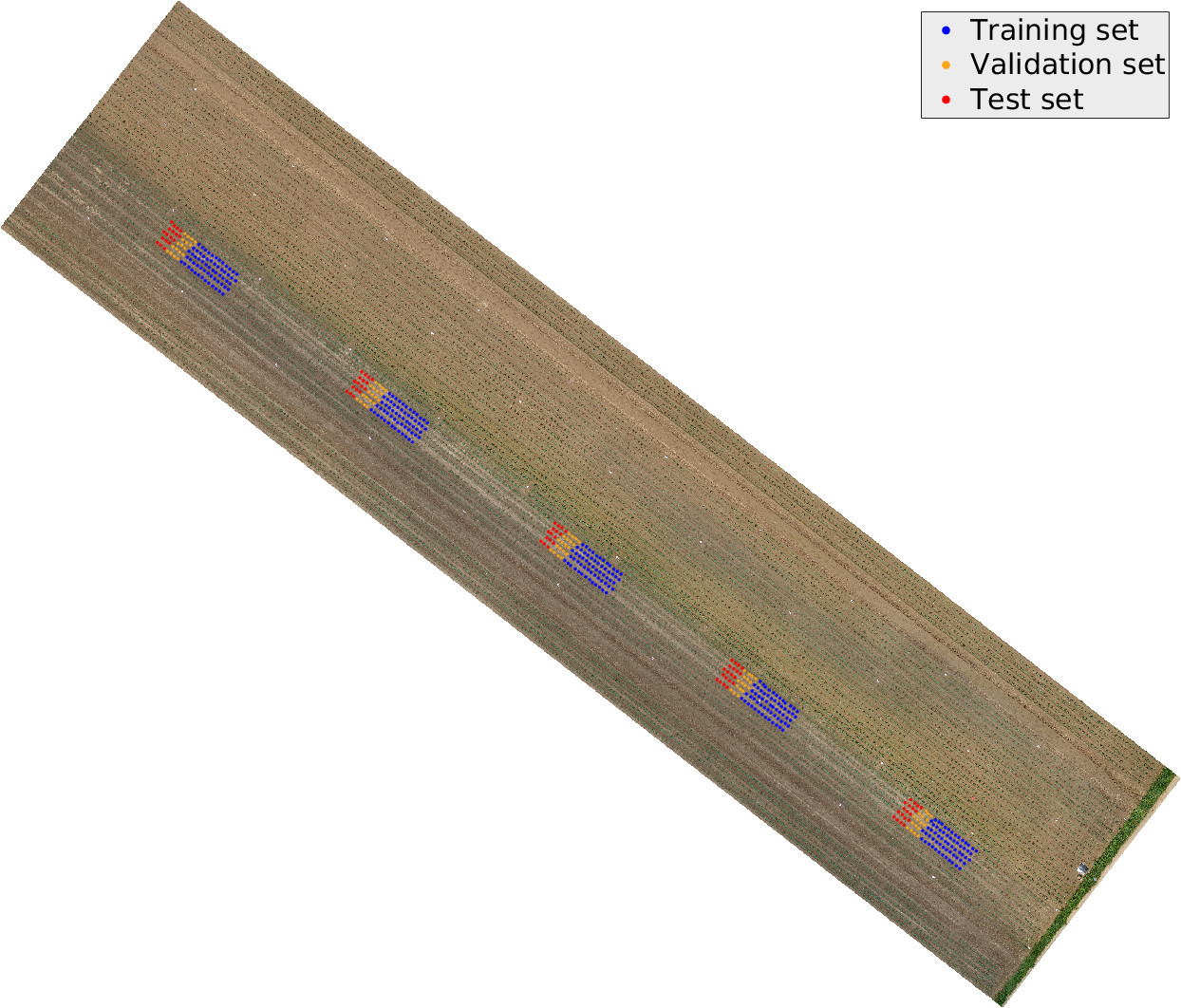}
        \label{fig:reference_map_trainValTest2}}
    \caption{Separation of reference plants of both fields within GrowliFlowerR in training (blue), validation (yellow) and test (red) set.}
    \label{fig:reference_map_trainValTest}
\end{figure}

\begin{figure}
    \centering
        \subfloat[Field 1]{
        \includegraphics[width=0.48\textwidth]{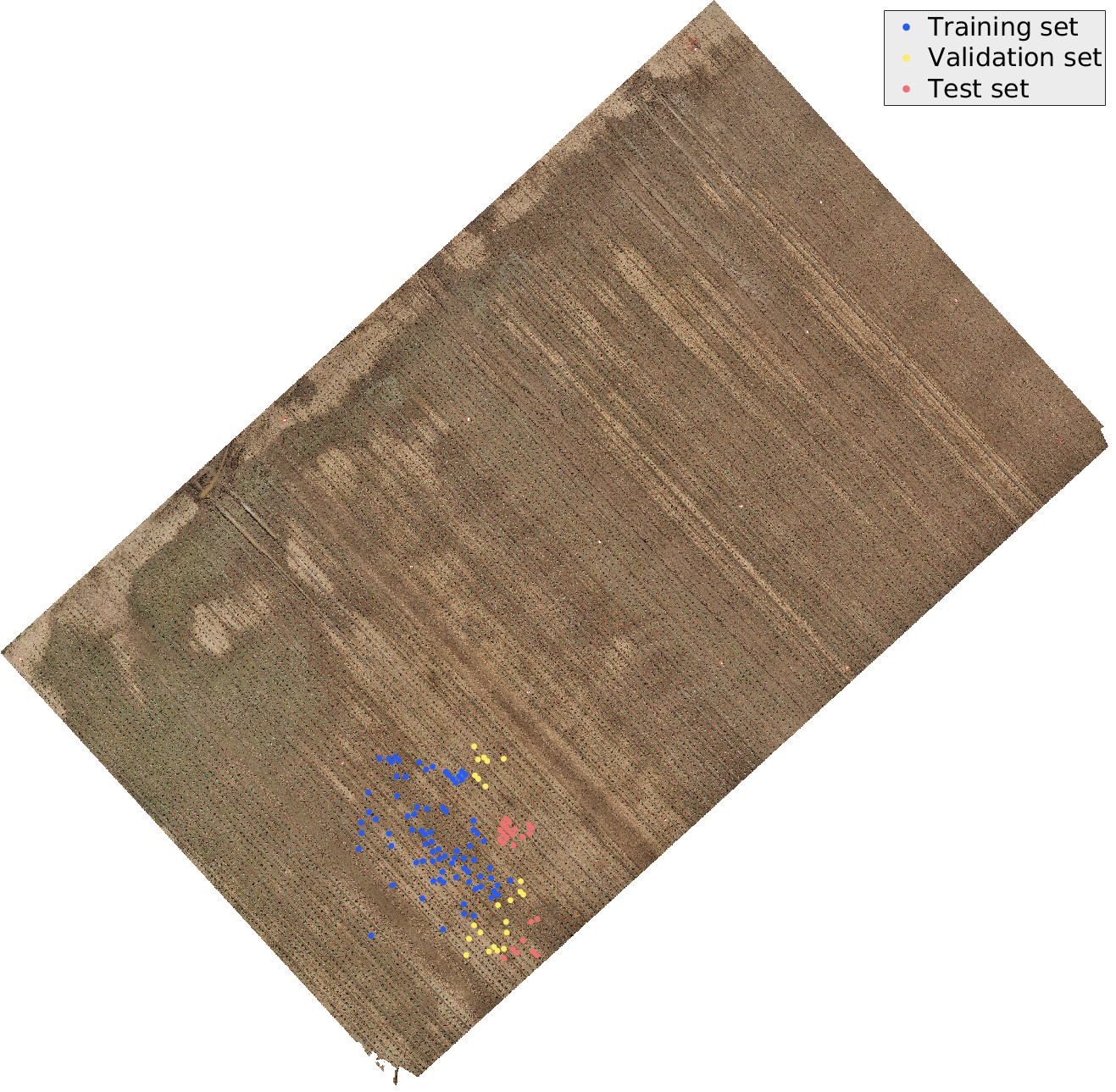}
        \label{fig:defoliation_map_trainValTest1}}
        \subfloat[Field 2]{
        \includegraphics[width=0.48\textwidth]{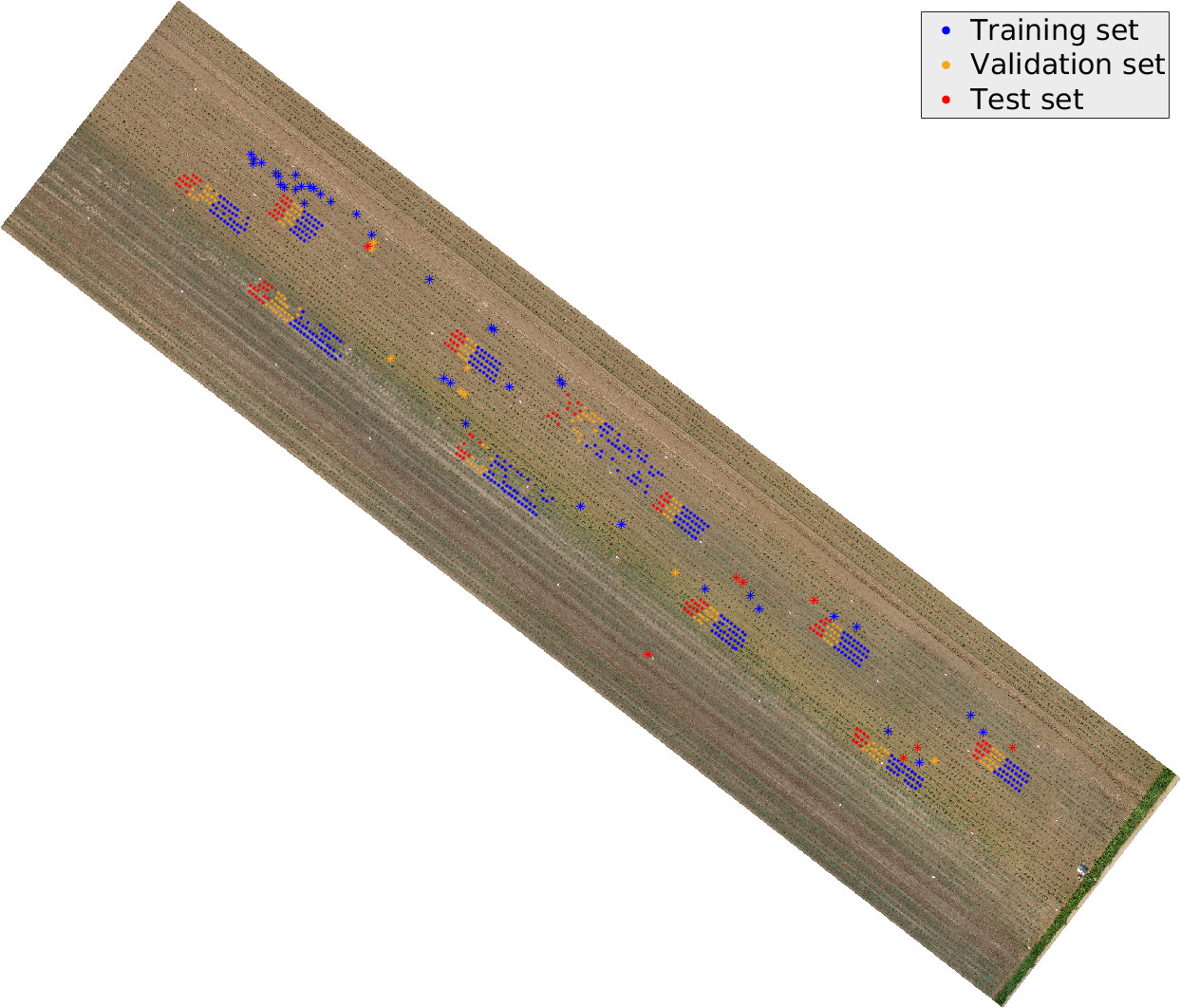}
        \label{fig:defoliation_map_trainValTest2}}
    \caption{Separation of defoliated plants of both fields within GrowliFlowerD in training (blue), validation (yellow) and test (red) set.}
    \label{fig:defoliation_map_trainValTest}
\end{figure}

\end{appendices}

\end{document}